%% file: Thesis.tex
\documentclass[12pt,oneside]{report}

\usepackage{arabtex}
\usepackage{epigraph}
\usepackage[section]{placeins}
\usepackage[dvipdfmx]{graphicx}
\usepackage[utf8]{inputenc}
\usepackage{float}
\usepackage{hyperref}
\usepackage[authoryear]{natbib}
\usepackage{bm}
\usepackage{amsfonts}
\usepackage{amssymb}
\usepackage{amsmath}
\usepackage{amsthm}
\usepackage{epsfig}
\usepackage{setspace} 
\usepackage{geometry}
\usepackage{epigraph}
\usepackage{bmpsize}
\usepackage{rotating}
\usepackage{sidecap}
\usepackage{wrapfig}
\usepackage{latexsym}
\usepackage{enumerate}
\usepackage{multirow}
\usepackage{array}
\usepackage{comment}
\usepackage{url}
\usepackage{latexsym}
\usepackage{listliketab}
\usepackage{mathtools}
\usepackage{esvect}
\usepackage{fmtcount}
\usepackage{booktabs}
\usepackage{lscape}
\usepackage{paralist}
\usepackage[usenames,dvipsnames,svgnames,table]{xcolor}
\usepackage[disable]{todonotes}
\usepackage{subfiles}
\usepackage[russian,english]{babel}
\usepackage{tabu}
\usepackage{caption}
\usepackage{subcaption}
\usepackage[titletoc]{appendix}
\usepackage{longtable}
\usepackage{csquotes}
\usepackage{datatool}
\usepackage{glossaries}
\usepackage{mathtools, cuted}
\usepackage{tikz}
\usetikzlibrary{shapes.geometric, arrows, positioning}
\usepackage{pgfplots, pgfplotstable}
\usepackage{adjustbox}
\usepackage{bibentry}
\usepackage{lipsum}
\usepackage{xr}
\nobibliography*

\definecolor{second_best}{HTML}{FDAE61}
\definecolor{first_best}{HTML}{ABDDA4}
\definecolor{third_best}{HTML}{2B83BA}
\pgfplotsset{compat=1.14}

\DeclareMathOperator*{\argmax}{arg\,max}

\input{lart}

\newcommand\given[1][]{\:#1\vert\:}
\newcommand\chapterbib{}

\hypersetup{colorlinks, citecolor=black, filecolor=black, linkcolor=black, urlcolor=black}

\begin{document}


\input{Cover.tex}


\newpage
\thispagestyle{empty}
\input{Declaration}


\input{Dedication.tex}

\newpage
\doublespacing
\pagenumbering{roman}
\tableofcontents
\listoffigures
\listoftables




\newpage
\phantomsection
\addcontentsline{toc}{chapter}{Abstract}
\input{Abstract.tex}


\newpage
\clearpage 
\addcontentsline{toc}{chapter}{Acknowledgements}
\input{Acknowledgments}

\doublespacing
\newpage
\pagenumbering{arabic} 

\setcounter{page}{1}

\input{Introduction.tex}
\input{Background.tex}
\input{SVAgreement.tex}
\input{Aspect_SMT_NMT.tex}
\input{SuperTags.tex}
\input{GenderNMT.tex}
\input{LexicalLoss.tex}
\input{Conclusion.tex}

\bibliographystyle{apalike}
\bibliography{Thesis}
\clearpage
\addcontentsline{toc}{chapter}{\bibname}
\onehalfspacing



\end{document}

%% file: lart.tex
\newcounter{licntr}			

\newcommand{\stepli}{\refstepcounter{licntr}}	
\newcommand{\lival}{\thelicntr}			
\newenvironment{liout}{
		\stepli			
		\samepage			
		\begin{list}{		
			(\lival)\hfill}{} 	
			\samepage
			\item 			
		}{ \end{list}}

\newenvironment{li}{
		\begin{liout} 			
		\begin{list}{\samepage \noindent}{} 	
		\item}{				
		\end{list} 
		\end{liout}}

\newenvironment{li*}{
	\stepli
	\samepage				
	\begin{trivlist} \item[] (\lival) \end{trivlist}
	\begin{trivlist} 
	{\samepage \item[]}
	}{\end{trivlist}}					

\newcounter{alphcntr}		

\newcommand{\bi}{\begin{itemize}}		 
\newcommand{\ei}{\end{itemize}}		 

\newcommand{\be}{\begin{enumerate}}		 
\newcommand{\ee}{\end{enumerate}}

\newcommand{\bc}{\begin{center}}		
\newcommand{\ec}{\end{center}}

\newlength{\astspace}				

\newcounter{figcntr}			

\newcommand{\stepfig}{\refstepcounter{figcntr}}	
\newcommand{\figval}{\thefigcntr}		

\newenvironment{figout}{
		\stepfig			
		\begin{list}{		
			(\figval)\hfill}{} 	
			\item 			
		}{ \end{list}}

\renewcommand{\thefigcntr}{\Alph{figcntr}}

\newcounter{axcntr}			

\newcommand{\stepax}{\refstepcounter{axcntr}}	
\newcommand{\axval}{\theaxcntr}		

\newenvironment{axout}{
		\stepax			
		\begin{list}{		
			(\axval)\hfill}{} 	
			\item 			
		}{ \end{list}}

\renewcommand{\theaxcntr}{\Roman{axcntr}}

\newcounter{alphacntr}		

%% file: Cover.tex

\begin{center}

\thispagestyle{empty}

{\huge {\bf On the Integration of Linguistic Features into Statistical and Neural Machine Translation}}
\\
[9ex]
{\huge Eva Odette Jef Vanmassenhove}
\\
[1.5ex]
{\large B.A., M.A., M.Sc.}
\\
[9ex]
\singlespacing
A dissertation submitted in fulfillment of the requirements for the
award of\\[3ex]
{\large Doctor of Philosophy (Ph.D.)}\\[3ex]
to the\\[5ex]

\begin{figure}[hp]
	\centering
	\includegraphics[scale=0.2]{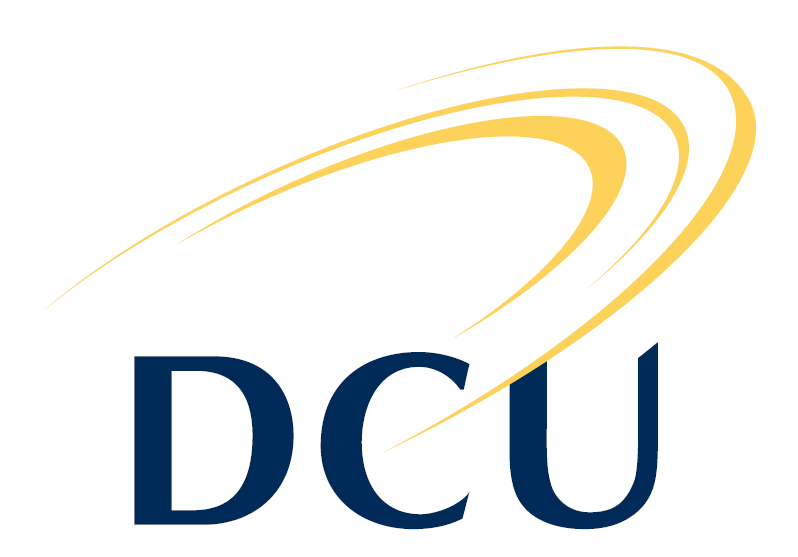}
\end{figure}

{\large Dublin City University}\\[1ex]
{\large School of Computing}\\[8ex]
{\large Supervisor: \\ Prof. Andy Way \\[8ex]}
{\large 2019}

\end{center}


%% file: Declaration.tex


\singlespace

\paragraph{}
I hereby certify that this material, which I now submit for assessment on the program of study leading to the award of Ph.D. is entirely my own work, that I have exercised reasonable care to ensure that the work is original, and does not to the best of my knowledge breach any law of copyright, and has not been taken from the work of others save and to the extent that such work has been cited and acknowledged within the text of my work.

\subparagraph{}
Signed:

\subparagraph{}
ID No.: 15211377

\subparagraph{}
Date: 09/09/2019


%% file: Dedication.tex

\clearpage
\begin{center}
    \thispagestyle{empty}
    \vspace*{\fill}
Aan Dimitar, Mama en Papa.
    \vspace*{\fill}
\end{center}
\clearpage

%% file: Abstract.tex
\singlespacing

\begin{center}
{\Large {\bf On the Integration of Linguistic Features into Statistical and Neural Machine Translation\\}}
\vspace{3ex}
{\large Eva Odette Jef Vanmassenhove}
\end{center}
\vspace{-7ex}

\section*{\center{Abstract}}\label{abstract}

Recent years have seen an increased interest in machine translation technologies and applications due to an increasing need to overcome language barriers in many sectors.\footnote{According to a report by Global Market Insights, Inc., the machine translation market will have a growth rate of more than 19\% over the period of 2016-2024. According to another study by Grand View Research, Inc., the expected market size will reach USD 983.3 million by 2022.} New machine translation technologies are emerging rapidly and with them, bold claims of achieving human parity such as: (i) the results produced approach ``accuracy achieved by average bilingual human translators [on some test sets]''~\citep{Wu2016} or (ii) the ``translation quality is at human parity when compared to professional human translators''~\citep{Hassan2018} have seen the light of day~\citep{Laubli2018}. Aside from the fact that many of these papers craft their own definition of human parity, these sensational claims are often not supported by a complete analysis of all aspects involved in translation.\footnote{Admittedly, analyzing and evaluating all aspects involved in translation systematically would be an extraordinary task, but so is the claim of achieving human parity. The work of~\cite{Toral2018} provides recommendations for future human evaluations of machine translation.}

Establishing the discrepancies between the strengths of statistical approaches to machine translation and the way humans translate has been the starting point of our research. 
By looking at machine translation output and linguistic theory, we were able to identify some remaining issues. The problems range from simple number and gender agreement errors to more complex phenomena such as the correct translation of aspectual values and tenses. Our experiments confirm, along with other studies~\citep{Bentivogli2016}, that neural machine translation has surpassed statistical machine translation in many aspects. However, some problems remain and others have emerged. We cover a series of problems related to the integration of specific linguistic features into statistical and neural machine translation, aiming to analyse and provide a solution to some of them.

Our work focuses on addressing three main research questions that revolve around the complex relationship between linguistics and machine translation in general. By taking linguistic theory as a starting point we examine to what extent theory is reflected in the current systems. We identify linguistic information that is lacking in order for automatic translation systems to produce more accurate translations and integrate additional features into the existing pipelines. We identify overgeneralization or `algorithmic bias'  as a potential drawback of neural machine translation and link it to many of the remaining linguistic issues.  
\\
\\
\emph{Keywords}: Statistical Machine Translation, Neural Machine Translation, Linguistics, Tense, Aspect, Subject-verb Agreement, Gender Bias, Gender Agreement, Lexical Diversity, Lexical Loss, Linguistic Loss, Algorithmic Bias.




%% file: Acknowledgments.tex
\singlespacing
\newpage
\epigraph{Knowledge is in the end based on acknowledgement.}{\textit{Ludwig Wittgenstein}}
~\newpage
\section*{\center{Acknowledgments}}\label{acknowledgments}

First and foremost, I would like to thank my supervisor, Andy Way, for his guidance and support throughout these four years of research. Andy not only encouraged and motivated me by highlighting and believing in the necessity of the research directions explored, he also regularly gave me the opportunity to step outside my comfort zone. By doing so, he allowed me to grow further as a researcher and as a person. Thank you, Andy.

Aside from Andy, I had the opportunity to work with Christian Hardmeier during a research visit in Uppsala University, Sweden. Christian's PhD thesis was the first thesis on Machine Translation I read and it had an influence on the overall direction of my research. As such, having been able to work with Christian in person was a privilege and I believe my work benefited tremendously from his input and knowledge. 

I would like to express my gratitude to Johanna Monti and Joss Moorkens, my examiners, for asking me challenging but interesting questions during the viva and for having shared their knowledge and comments with me. Similarly, I would like to thank Cathal Gurrin for having chaired the viva. 

During my education, I had the privilege to encounter many inspiring teachers and mentors that I believe all played a part in my personal and academic development. Educators whose knowledge and passion inspired me include: Juf Gerda Casteels, Juf Paula and Juf Lea Verhoeven and Meester Luc Michiels from the Vrije Basisschool Haacht-Station; Meneer Kurt Maes, Meneer Willy Wuyts, Meneer Herman Cauwenberghs, Meneer Hugo Godts, Mevrouw Liselot Wolfs, Meneer Guido Locus, Meneer Roel Van De Poel and Mevrouw Erna Vanderhoeven from Don Bosco Haacht; and Professor Nicole Delbecque, Professor Jan Herman, Professor Bert Cornillie, Professor Vincent Vandeghinste and Professor Frank Van Eynde from KULeuven. All these people are fantastic educators with a true passion for their job and field.

~\newline
On a more personal note, I would like to thank, Mama and Papa, who aside from being great teachers themselves, have also been the most supportive parents one could ask for. They are truly inspiring people that have always been supportive in any way they could, providing me with all the necessary tools and guidelines. Thank you, Mama and Papa, this is as much your accomplishment as it is mine. Similarly, I would like to thank the family I gained during this PhD: Shtelian, Elinka, Sasho and Baba Zora.

My life would not have been the same without my amazing grandparents. Moeke, P\'{e}p\'{e}, M\'{e}m\'{e} and Papo have always been there for me. Moeke taught me to pay attention to detail and to do things as precisely as I could. P\'{e}p\'{e} balanced that out a little bit by teaching me how to be efficient and effective. M\'{e}m\'{e} is the most modern and open-minded grandmother you can imagine. She definitely gave me the travel-bug and taught me to be an empath(et)ic human being. Papo is incredibly eager to learn, he is a real bookworm who taught me to be curious and made me see that one should never stop learning.

The four years in DCU would have not been the same without the support of my colleagues. Alberto, we started and finished this journey together. I am very glad to have shared this experience with you. Meghan, I think very highly of you as a researcher and a friend. Thank you for always having my back and for lending a listening ear whenever I needed it. Aliz\'{e}e, thank you for the many chats we had during coffee breaks.

The DCU campus choir offered me a nice break from my research every Wednesday, allowing me to sing and socialize with some people who I now consider my friends. Chrissie, thank you for being the kindest choir director. Lisa, thank you for the many walks and the mental (and physical) support.

My friends in Belgium, in particular ``de Menne'' and ``de Amigas'' for their continuous encouragements and for making me associate education with fun.

~\newline
Last but not least, I would like to thank Dimitar. Dimi, if it weren't for you, I do not think this journey would have started or ended the way it did. First of all, you did not hesitate a single second to follow me to Ireland, a country even less sunny and more rainy than Belgium, when I was offered the possibility to start a PhD here. You had just obtained your PhD and moving to Ireland meant you probably had to start working in a completely different field. This field happened to be Machine Translation, a field in which you, as with all things you do, soon thrived. You were the one who encouraged me to continue when I wanted to give up, the one who cheered me up when I was down but also the one who really taught me what it is to be a good researcher. You are so dedicated, smart, passionate and kind, something I admire as your colleague and as your wife.

~\newline
This research would not have been possible without the financial support received from Dublin City University under the Daniel O'Hare Scholarship Scheme; Science Foundation Ireland (Grant 13/RC/2106); COST Action IS1312 which funded my one-month research stay at Uppsala universitet and Roam Analytics who provided me with a travel grant to attend the ACL conference in August, 2018.


%% file: Introduction.tex

\epigraph{The whole problem with Artificial Intelligence is that bad models are so certain of themselves, and good models are so full of doubt.}{\textit{Bertrand Russell}}

\chapter{Introduction}\label{ch:Introduction}

Machine Translation (MT) is the automatic translation of text (or speech) from one natural language into another by means of a computer system. Uncertainty, creativity and common-sense reasoning are just a few elements that come into play when dealing with natural languages and they pose great difficulties for computer systems. As translations deal with at least two natural languages,\footnote{In some cases, the term monolingual MT is used for tasks such as automatic post-editing. However, essentially, there are still two languages at play in the final translation process.} they involve skills that go beyond mere competence in a single language, making it a complex task for both humans and machines. For machines, it requires a thorough `understanding' and `formalization' of the source language as a whole, as well as formalizing a process that allows it to transfer that understanding into a target language. Current approaches to MT --Statistical MT (SMT) and Neural MT (NMT)-- address this task by leveraging statistical information extracted from large datasets of translated texts. Many of the SMT approaches eventually became hybrid approaches. They leveraged information extracted from the statistical patterns as well as additional linguistic information. The NMT paradigm extended the relatively impoverished context of SMT models to the sentence level. With its arrival, technical constraints and advances started shaping the field more so than any linguistic concerns~\citep{Hardmeier2014}. Did the extension of context available for the NMT models make linguistic features superfluous? Can technological advances in combination with larger datasets solve the remaining issues?

In this thesis, we initially focus on a range of linguistic phenomena, comparing both phrase-based SMT (PB-SMT) and NMT paradigms. Early on, and supported by other research in this area, the superiority of NMT compared to PB-SMT when it comes to tackling sentence-level syntactic and semantic problems became clear. From then on, our focus shifted towards linguistic features and NMT. NMT's superiority initially questioned the need for linguistic features at all. However, we set out to identify whether NMT systems can indeed handle both simple and more complex linguistic phenomena in a systematic way. After having identified some of the remaining issues in NMT, we explore ways of exploiting linguistic features in order to resolve them.

\section{Motivation and Research Questions}\label{sec:motivation}
Establishing the discrepancies between the strengths of statistical approaches to MT and the way humans translate has been the starting point of our research. At a very early stage, we observed that the mistakes PB-SMT systems make are very different from the type of mistakes made by humans. For example, despite being exposed to millions of parallel texts, PB-SMT systems are still not able to produce sentences with correct agreement among its arguments consistently. This is somewhat surprising given that agreement rules are one of the most systematic elements of language and so learning them should not be too hard for an MT system. For PB-SMT systems, many problems could be traced back to their technical constraints that sacrificed more complex linguistic relations for computational efficiency~\citep{Hardmeier2014}. Resolving some of the grammatical issues would simply involve injecting the appropriate linguistic information in the appropriate place by using, for example, reordering techniques. When the field of MT shifted towards neural approaches, the picture became less clear. NMT systems encode the entire sentence at once, which from a theoretical point of view should give it a clear advantage over PB-SMT. Indeed, in practice, NMT resolves many of the most obvious issues of SMT, however not consistently. These inconsistencies in its performance give us an idea of the underlying competence of NMT and it is only by further looking into the output of the systems that we can identify remaining problems.  

The main questions addressed in this thesis revolve around the complex relationship between linguistics and MT in general. Is it at all necessary to still consider linguistic theories? Do we need linguistic features? Are the underlying algorithms and models equipped with the right tools to deal with something as complex as language? In the following section, we formulate the main research questions we aim to address throughout this thesis. 

\subsection{Research Questions}~\label{subsec:rqs}
Our work focuses around three central research questions.

\subsubsection{Research Question 1}
Existing linguistic and translation theories can help us obtain a better understanding of intricate translation problems. However, there are very few contrastive linguistic studies, and the monolingual grammars of languages differ in terms of focus and terminology. Furthermore, monolingual grammars often focus on exceptions or rare cases that are illustrated with sentences that are collected or simply created by grammarians and thus impede necessary generalisations for a field such as MT where the frequency and systematicity of linguistic phenomena are important. Therefore, our first research question is: 
\\
\textbf{RQ1: Is linguistic theory reflected in practice in the knowledge sources of data-driven Machine Translation systems?}
\\
This question is mainly dealt with in Chapter~\ref{ch:Aspect}, where we focused on one particular grammatical category, `aspect', related to the verb tense systems. Although tense and aspect have received a lot of attention in linguistic fields such as formal semantics and logic (Mc Cawley, 1971; Richards, 1982), few translation studies compare one specific linguistic aspect across parallel corpora. Even fewer do so with a computational linguistic application in mind. Our goal is to see whether linguistic theory is reflected in the PB-SMT phrase tables and in the learned NMT sentence-encoding vectors. Although the majority of work related to our first research question is addressed in Chapter~\ref{ch:Aspect}, a large part of our research motivation can be found in linguistic theory itself. Accordingly, we often refer back to relevant linguistic research for issues related to gender agreement or differences in language usage between male and female speakers (see Chapter~\ref{ch:Gender}). A large body of research related to gender and language can be found in the field of sociolinguistics, Lakoff being one of the pioneers~\citep{lakoff1973language}. Similarly, Chapter~\ref{ch:Loss} on lexical richness relies on techniques and work used in (human) translation studies which we applied to the field of MT. 

To some extent, linguistic theory is relevant to all chapters in this thesis as the motivations behind our work are largely based on linguistic errors and issues found in the output of current MT systems. The question whether it is indeed reflected and encoded in MT systems is rather broad, but it is a question that needs to be asked as we still too often see a gap between theoretical linguistic studies and its practical applications in Natural Language Processing (NLP) in general.

\subsubsection{Research Question 2}
Although there has been a lot of research on integrating linguistics in PB-SMT, its integration into NMT has only just begun. Apart from merely adding linguistic knowledge and reporting the BLEU~\citep{Papineni2002} scores, we aim to explore in more depth what specific linguistic information is lacking in order for MT systems to overcome linguistic problems, or which information is too `difficult' for data-driven MT systems to extract by themselves, e.g. linguistic problems that require some deeper understanding and (meta-)context. In order to do so, we will focus on the output of MT systems in order to identify recurring problems. As \cite{Sennrich2016} have shown, integrating `more' linguistic information does not necessarily lead to better translations. Therefore, our second research question is formulated as follows: 
\\
\textbf{RQ2: What type of (necessary) linguistic knowledge is lacking, and how can this be integrated in data-driven MT systems?}
\\
This question is not so hard to answer for PB-SMT systems since we know such systems rely on \emph{n}-grams, which have very obvious shortcomings, e.g. any dependency or construction that requires information further than \emph{n} will be `unsolvable' for a baseline PB-SMT system.  For the recently developed NMT systems, many of their weaknesses remain unclear. Apart from knowing the type of information that is needed and currently lacking, we would like to gain more insights into why this is the case and how this can be resolved. Often linguistic information is added to existing systems without further analysis on how this affects the actual translation output or without mentioning what the potential drawbacks might be. We assume that gaining knowledge about the problem by looking at the actual outputs of the black box that NMT currently is, is the first step towards finding a solution. We analyze linguistic issues and provide feature integration in Chapter~\ref{ch:Agreement} for PB-SMT related to number agreement issues. In Chapter~\ref{ch:Supertag} we integrate sentence-level semantic and syntactic features into NMT systems and observe how, unlike in PB-SMT systems, they can be a useful combination. Chapter~\ref{ch:Gender} deals with the integration of sentence-level features providing the NMT system with information of the gender of the speaker of particular utterances.

\subsubsection{Research Question 3}
Once we have studied the related linguistic theories and identified remaining linguistic problems as well as having incorporated potentially useful linguistic features, we observed general tendencies of MT systems that we believed could be traced back to a common issue: the inability of algorithms to deal with the richness and \emph{many-to-many} relationships that exist in natural languages. Therefore, our third question is formulated as:
\\
\textbf{RQ3: Can we identify and quantify the underlying cause of many of the linguistic issues remaining in current MT systems?}
\\
Throughout the research conducted to answer RQ1 and RQ2, we analyzed and assessed the performance and output of MT systems, identifying issues and aiming to provide solutions to them. While addressing the aforementioned questions, we came to the conclusion that the individual problems we have addressed\footnote{Number agreement, gender agreement, aspectual information and tenses.} might occur due to a more systematic problem at the core of technologies used in MT:
the loss of linguistic richness caused by the learning mechanisms of the currently employed algorithms. Generalizations are crucial to the learning process of Artificial Intelligence (AI) algorithms. However, overgeneralization can be detrimental not only to semantic richness (in terms of synonyms) but also to grammatical issues (as our systems do not distinguish between syntax and semantics) related to `minority' word forms. This is to be understood in the broad sense, for example:
\begin{itemize}
\item \textbf{number agreement}, as 3$^{rd}$ person verb forms are often more frequent than others such as the 1$^{st}$ or 2$^{nd}$, leading to frequent agreement issues in MT output when less frequent verb forms need to be generated (Chapter~\ref{ch:Agreement}),
\item \textbf{gender agreement}, where the MT system determines the gender of nouns based on previously seen examples even when provided with contradictory evidence in the sentence it is presented with (Chapter~\ref{ch:Gender}), and
\item \textbf{aspect}, where the aspectual value of a verb is determined based on its most likely aspect, disregarding the aspectual clues provided in the sentential context (Chapter~\ref{ch:Aspect}).
\end{itemize}

\noindent This research question is addressed in Chapter~\ref{ch:Loss}.



\section{Publications}\label{sec:intro:publications}

A considerable amount of the work discussed in this thesis is based on research that has been published previously in peer-reviewed conference papers, journals or in the form of abstracts. Many of the experiments conducted and described in the individual chapters are based upon these publications but have been updated and extended. We first describe how the individual chapters are related to prior publications in Section~\ref{subsec:relPub}. Aside from presenting our work at conferences, three additional invited talks were given on topics related to our research. They are listed in Section~\ref{subsec:talks}. Finally, we list other publications that were published but that were not directly integrated into the thesis in Section~\ref{subsec:other}.

\subsection{Relation to Published work}\label{subsec:relPub}

We briefly describe how each of the content chapters of this thesis relate to previously published work. Chapter~\ref{ch:Background} provides general background information as well as a discussion of some of the related work relevant to the topics covered.
\\
\noindent{\textbf{Chapter~\ref{ch:Agreement}}} \\
An earlier version of the PB-SMT experiments described in Chapter~\ref{ch:Agreement} was published in paper format and presented at the the European Association for Machine Translation (EAMT) workshop on Hybrid System for Machine Translation (HyTra) in Riga, Latvia, 2017~\citep{vanmassenhove2016improving}.

\begin{itemize}
  \item \textbf{Vanmassenhove, E.}, Du, J. and A. Way (2016). Improving Subject-Verb Agreement in SMT. In Proceedings of HyTra (EAMT), May, Riga, Latvia.
\end{itemize}

\noindent{\textbf{Chapter~\ref{ch:Aspect}}} \\
A first draft of the contrastive linguistic work on the translation of tenses described in Chapter~\ref{ch:Aspect} has been published as a one-page abstract and presented at the Computational Linguistics in The Netherlands Conference (CLIN27) in Leuven, Belgium, 2017~\citep{Vanmassenhove2017b}.

\begin{itemize}
 \item \textbf{Vanmassenhove, E.}, Du, J. and A. Way (2017). Extracting Contrastive Linguistic Information from Statistical Machine Translation Phrase-Tables. In \emph{Book of Abstracts of the 27th Conference on Computational Linguistics in The Netherlands (CLIN27)}, page 77. February, Leuven, Belgium.
\end{itemize}

An extension of this work was later published and presented at the 8th International Conference of Contrastive Linguistics (ICLC8), Athens, Greece, 2017~\citep{Vanmassenhove2017c}.

\begin{itemize}
  \item \textbf{Vanmassenhove, E.}, Du, J. and A. Way (2017) Phrase-Tables as a Resource for Cross-Linguistic Studies: On the Role of Lexical Aspect for English-French Past Tense Translation. In \emph{Proceedings of the 8th International Conference of Contrastive Linguistics (ICLC8)}, pages 21--23. May, Athens, Greece.
\end{itemize}

The final experiments were published and described in more detail in a journal article in the Journal of Computational Linguistics in The Netherlands~\citep{vanmassenhove2017investigating}.

\begin{itemize}
  \item \textbf{Vanmassenhove, E.}, Du, J. and A. Way (2017). `Aspect' in SMT and NMT. \emph{Journal of Computational Linguistics in The Netherlands (CLIN)}, pages 109--128. December, Leuven, Belgium.
\end{itemize}

\noindent{\textbf{Chapter~\ref{ch:Supertag}}}
\\
The initial experiments on integrating supersenses and supertags that served as a basis for Chapter~\ref{ch:Supertag} have been presented and published as an abstract in the Book of Abstracts of the Computational Linguistics in The Netherlands Conference in Nijmegen, The Netherlands, 2018~\citep{Vanmassenhove2018clin}.

\begin{itemize}
  \item \textbf{Vanmassenhove, E.} and A. Way (2018). SuperNMT: Integrating Supersense and Supertag Features into Neural Machine Translation. In \emph{Book of Abstracts of the 28th conference on Computational Linguistics in The Netherlands (CLIN28)}, pages 71. December, Nijmegen, The Netherlands.  
\end{itemize}

An extension of the work has been published in the conference proceedings of the 56th Annual Meeting of the Association for Computational Linguistics SRW in Melbourne, Australia, 2018~\citep{vanmassenhove2018supernmt}.
\begin{itemize}
     \item \textbf{Vanmassenhove, E.} and A. Way (2018). SuperNMT: Neural Machine Translation with Semantic Supersenses and Syntactic Supertags. In \emph{ Proceedings of the 56th Annual Meeting of the Association for Computational Linguistics SRW (ACL-SRW)}, pages 67--73. July, Melbourne, Australia.
\end{itemize}

\noindent{\textbf{Chapter~\ref{ch:Gender}}}
\\
This chapter on the integration of gender features and the compilation of multiple corpora is based on work that was previously published as two short papers. The Europarl dataset we compiled was published and presented in the Proceedings of the 2018 Conference of the European Association for Machine Translation, in Alicante, Spain, 2018~\citep{vanmassenhove2018europarl}. A large part of this research was conducted during a research stay at the University of Uppsala under the supervision of Christian Hardmeier.

\begin{itemize}
  \item \textbf{Vanmassenhove, E.} and Hardmeier, C. (2018). Europarl Datasets with Demographic Speaker Information. In \emph{Proceedings of the 2018 Conference on European Association for Machine Translation (EAMT)}, pages 15--20. May, Alicante, Spain.
\end{itemize}

The experiments conducted on the integration of gender in NMT have been published and presented at the 2018 Conference on Empirical Methods in Natural Language Processing, Brussels, Belgium, 2018~\citep{vanmassenhove2019getting}.

\begin{itemize}
  \item \textbf{Vanmassenhove, E.}, Hardmeier, C. and A. Way (2018). Getting Gender Right in Neural MT. In \emph{Proceedings of the 2018 Conference on Empirical Methods in Natural Language Processing (EMNLP)}, pages 3003--3008. November--October, Brussels, Belgium. 
\end{itemize}

\noindent{\textbf{Chapter~\ref{ch:Loss}}}
\\
Our final experiments on the loss of lexical richness in PB-SMT and NMT briefly refer to our winning model of the CLIN29 shared task for cross-genre gender prediction presented and later on published in the Proceedings of the Shared Task of the Conference of Computational Linguistics in The Netherlands Conference (CLIN29) taking place in Groningen, The Netherlands, 2019~\citep{vanmassenhove2019abi}.

\begin{itemize}
  \item \textbf{Vanmassenhove, E.}, Moryossef, A., Poncelas, A., Way, A. and Shterionov, D. (2019). ABI Neural Ensemble Model for Gender Prediction: Adapt Bar-Ilan Submission for the CLIN29 Shared Task on Gender Prediction. In \emph{Proceedings of the 2019 Computational Linguistics in The Netherlands (CLIN) Shared Task}, pages 16--20. January, Nijmegen, The Netherlands.
\end{itemize}

Chapter~\ref{ch:Loss} furthermore draws on a recent paper that has been presented at the 17$^{th}$ Machine Translation Summit (MT Summit XVII) which took place in August 2019, Dublin, Ireland~\citep{vanmassenhove2019lost}. 

\begin{itemize}
  \item \textbf{Vanmassenhove, E.}, Shterionov, D. and A. Way (2019). Lost in Translation: Loss and Decay of Linguistic Richness in Neural and Statistical Machine Translation. \emph{Proceedings of the 17$^{th}$ Machine Translation Summit (MT Summit XVII)}, pages 222-232. August, Dublin, Ireland. 
\end{itemize}


\subsection{Invited talks}\label{subsec:talks}
The following invited talks relate to Chapter~\ref{ch:Aspect} and Chapter~\ref{ch:Gender}.

\begin{itemize}
    \item Vanmassenhove, E. What do NMT and SMT Know about `Aspect' and How Does this Translate? The Time in Translation Kick-off Workshop. 23 June 2017. University of Utrecht. Utrecht, The Netherlands.  
\end{itemize}

\begin{itemize}
    \item Vanmassenhove, E. Getting Gender Right in Neural MT. Women in Research. 18 March 2019. ADAPT, Trinity, Dublin, Ireland. 
\end{itemize}

\begin{itemize}
    \item Vanmassenhove, E. Gender and Machine Translation. 3 April 2019. Google. Mountain View, San Francisco, USA.  
\end{itemize}

\begin{itemize}
    \item Vanmassenhove, E. On the Integration of (Extra-) Linguistic Information in Neural Machine Translation: A Case Study of Gender. 19 August 2019. MomenT Workshop, The Second Workshop on Multilingualism at the intersection of Knowledge Bases and Machine Translation, co-located with the Machine Translation Summit (MT Summit XVII). Dublin, Ireland.  
\end{itemize}

To additional invited talks related to Chapter~\ref{ch:Gender} and Chapter~\ref{ch:Loss} have been scheduled and will take place in November 2019.  
\begin{itemize}
    \item Vanmassenhove, E. Lexical Loss, Gender and Machine Translation. 13 November 2019. Amazon. Berlin, Germany.
\end{itemize}

\begin{itemize}
    \item Vanmassenhove, E. Gender and Machine Translation. 19 November 2019. CrossLang. Luxembourg, Luxembourg.
\end{itemize}

\subsection{Additional Publications}\label{subsec:other}

Other publications~\citep{cabral2016adapt,moorkens2016translation,reijers2016need, vanmassenhove2016prediction} that were co-authored during this PhD but are not directly related to the work conducted in this thesis are listed below:
\begin{itemize}

    \item Reijers, W., \textbf{Vanmassenhove, E.}, Lewis, D. and J. Moorkens (2016). On the Need for a Global Declaration of Ethical Principles for Experimentation with Personal Data. In \emph{Proceedings of ETHI-CA2 (LREC): Ethics In Corpus collection, Annotation and Application}, pages 18--22 , May, Portoroz,  Slovenia.

    \item Moorkens, J., Lewis, D., Reijers, W., \textbf{Vanmassenhove, E.} and A. Way (2016). Language Resources and Translator Disempowerment. In \emph{Proceedings of ETHI-CA2 (LREC): Ethics In Corpus collection, Annotation and Application}, pages 49--53. May, Portoroz, Slovenia.
    
    \item \textbf{Vanmassenhove, E.}, Cabral, J. P. and F. Haider (2016). Prediction of Emotions from Text Using Sentiment Analysis for Expressive Speech Synthesis. In \emph{Proceedings of the 9th ISCA Speech Synthesis Workshop (SSW9)}, pages 22--27. September, Sunnyvale, CA, USA, September. 
    
    \item  Cabral, J. P., Saam, C., \textbf{Vanmassenhove, E.}, Bradley, S. and F. Haider (2016). The ADAPT Entry to the Blizzard Challenge. In \emph{Proceedings of the 9th ISCA Speech Synthesis Workshop (SSW9)}. September, Sunnyvale, CA, USA. 
    
\end{itemize}
\chapterbib


%% file: Background.tex


\newpage
\epigraph{Que otros se jacten de las p\'aginas que han escrito; a m\'i me enorgullecen las que he le\'ido.\footnotemark}{\textit{Jorge Luis Borges}}

\footnotetext{[May others be proud of the pages they have written, I am proud of the ones I have read.] Translated by Eva Vanmassenhove.}
\chapter{Background and Related Work}\label{ch:Background}

In this initial chapter, we provide a detailed description of the different MT paradigms covered and used for experimentation throughout the chapters of this thesis (Section~\ref{sec:related:machine_translation}). Additionally, we discuss previous work on the integration of linguistics in the field of MT. As integrating linguistic knowledge is a recurrent theme in our work, we elaborate on the perception and the integration of linguistic features through the different MT paradigms (Section~\ref{sec:linguisticsMT}). We targeted several translational difficulties related to differences in terms of explicitation and morphology. While covering issues such as gender agreement, the link between overcoming simple morphological problems and broader ethical ones related to gender bias became apparent. As questions on diversity and ethics in the field of AI have seen a surge recently and have become apparent as well in the field of MT, we include a section on bias in AI, focusing specifically on MT (Section~\ref{sec:diversityMT}).

The objective of this chapter is to give a broad overview of the state-of-the art in MT. A complete and detailed overview of research pertinent to the topics covered can be found in the separate related work sections throughout the content chapters of our thesis.

\section{Machine Translation}\label{sec:related:machine_translation}


Until the end of the 80s, linguistic Rule-Based Machine Translation (RBMT) methods governed the field. The first statistical models appeared when \cite{Brown1990} introduced Word-Based SMT (WB-SMT). Several short-comings of WB-SMT were improved upon by PB-SMT~\citep{Koehn2003}. Soon after PB-SMT was first suggested, it became the dominant paradigm. In 2015, when we started our research, PB-SMT was still the dominant paradigm in the field. 
More recently, however, NMT, a statistical method based on deep learning techniques, has taken over the field, beating previous PB-SMT state-of-the-art results on multiple levels for many language pairs.

Following this chronological order, we start by introducing WB-SMT models and PB-SMT in Section~\ref{sec:related:statistical_machine_translation}.
An overview of the different NMT models is provided in Section~\ref{sec:related:neural_machine_translation}. Finally, when carrying out MT experiments, the topic of evaluation cannot be avoided. As such, we dedicate Section~\ref{sec:evaluationMT} to automatic evaluation metrics.






\subsection{Statistical Machine Translation}\label{sec:related:statistical_machine_translation}
SMT formalizes the idea of producing a translation that is both faithful to the original source text and fluent in the target language. This goal is achieved in SMT by combining probabilistic models that maximize faithfulness (or accuracy) and fluency to select the most probable translation candidate, as in Equation~(\ref{eq:intuitive}):   

\begin{equation}\label{eq:intuitive}
  \textrm{best-translation}~\hat{E}=\argmax_{E}\textrm{faithfulness(E,F) fluency(E)}
\end{equation}

\begin{figure}[!ht]
  \centering
  \includegraphics[scale=0.75]{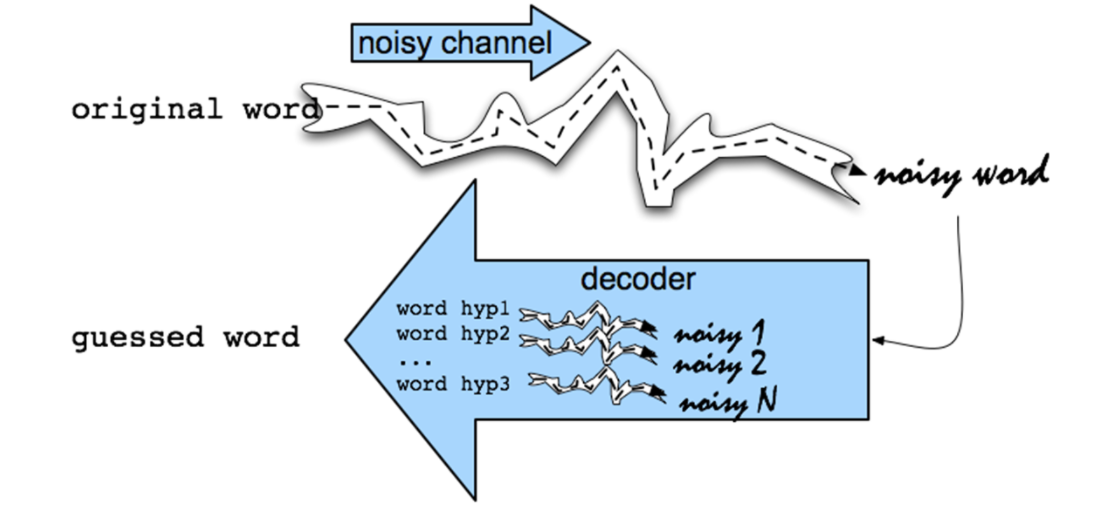}
  \caption{The noisy channel model of SMT~\citep{Jurafsky2014}.}
  \label{fig:noisy_channel}
\end{figure}

To achieve this, SMT uses the noisy channel model. The intuition behind a noisy channel model is that the original source (F) sentence is a distortion of the target sentence (E) as it has been passed through a noisy communication model. The goal, is to model this `noise' in such a way that we can pass the observed `distorted' source sentence through our model and discover the hidden target language sentence ($\hat{E}$). More concretely, say we have a French source sentence for which we want to produce an English translation. The noisy channel model assumes the French sentence is simply a distortion of the English one. The task is to build a model that allows you to generate from an English `source' sentence the French `target' sentence by discovering the underlying noisy channel model that distorted the `original' English sentence. Once this has been modeled, we take the French sentence, pretend it is the output of an English sentence that has been passed through our model and we generate the most likely English sentence~\citep{Jurafsky2014}. An illustration of the noisy channel model can be found in Figure~(\ref{fig:noisy_channel}).

More formally, we want to translate a French sentence $F$ into an English sentence $E$. To do so, we traverse the search space and find the English sentence $\hat{E}$ that maximizes the probability $P(E \given F)$, as in Equation~(\ref{eq:noisy-channel}):

\begin{equation}\label{eq:noisy-channel}
  \hat{E} = \argmax_{E}{P(E \given F)}
\end{equation}

Rewriting Equation~(\ref{eq:noisy-channel}) with Bayes' rule results in Equation~(\ref{eq:noisy-channel-bayes-theorem}). The resulting noisy channel equation consists of two components: a translation model $P(F \given E)$ and a language model $P(E)$~\citep{Brown1990}.

\begin{equation}\label{eq:noisy-channel-bayes-theorem}
  \hat{E}=\argmax_{E \in English}{ P(F \given E) P(E)}
\end{equation}

Aside from the language model taking care of the fluency of the output, the translation model makes sure the translation is adequate with respect to the source. A decoder is needed in order to compute the most likely English sentence $\hat{E}$ given the French sentence $F$.

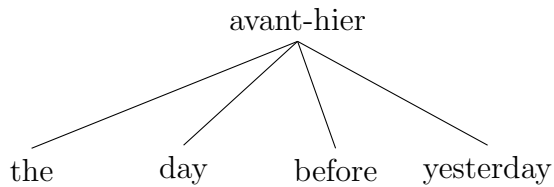
\begin{figure}[!ht]
\centering
\begin{tikzpicture}[node distance=2cm]
\node (sw1) [xshift=0cm] {the};
\node (sw2) [xshift=2cm] {day};
\node (sw3) [xshift=4cm] {before};
\node (sw4) [xshift=6cm] {yesterday};
\node (tw1) [xshift=3.5cm, yshift=2cm] {avant-hier};
\draw (sw1.north) -- (tw1.south);
\draw (sw2.north) -- (tw1.south);
\draw (sw3.north) -- (tw1.south);
\draw (sw4.north) -- (tw1.south);
\end{tikzpicture}
\caption{\emph{One-to-many} relation between the French word `avant-hier' and its English translation that consists of multiple words `the day before yesterday'.} \label{fig:onetomany}
\end{figure}

Initially, WB-SMT used words~\citep{Brown1990} as fundamental units in order to compute the equations described, but it soon became clear that working with phrases~\citep{Zens2002,Koehn2003} as well as single words could lead to considerably better translations. One of the major issues with WB-SMT models is the fact that such models do not allow multiple words to be mapped or moved as one unit. In reality, we know that so-called \emph{one-to-many} and \emph{many-to-one} mappings are in no way exceptional when dealing with translations (see Figure~\ref{fig:onetomany}). Note that, in PB-SMT, the term \emph{phrases} is not to be confused with what is called a phrase in linguistics. A phrase in linguistics refers to a group of words that form a unit within the grammatical hierarchy, while the term \emph{phrase} in PB-SMT refers to consecutive words in a sentence (commonly referred to as $n$-grams). 

Using phrases instead of words did not change the fundamental components of the SMT pipeline (language model, translation model and decoder). However, the decoding process became a more complex task consisting not only of words (or unigrams) as features but unigrams in combination with bigrams, trigrams, etc.\footnote{Typically an PB-SMT system would not use more than $5$-grams.} As such, \cite{Och2001} propose a more general framework, the log-linear model, to replace the noisy-channel model (described in Equation~(\ref{eq:noisy-channel})) that allows for the integration of an arbitrary number of features. The most likely translation can now be found by computing Equation~(\ref{eq:logLin}).\footnote{Note that the noisy-channel model can be considered a special case of the log-linear model when $M$=$2$, $\lambda_{1}$=$\lambda_{2}$=$1$, $h_{1}$=$log P(F\given E)$ and $h_2$=$logP(T)$~\citep{Hearne2011}} As in the previous equations, F represents the French source sentence, E the English target sentence and $\hat{E}$ the most likely English translation. Additionally, $h_{i}(F,E)$ defines the feature functions, $M$ the number of feature functions and $\lambda_{i}$ their weights.

\begin{equation}\label{eq:logLin}
  \hat{E} = \argmax_t \lambda_i h_{i}(F,E)
\end{equation}


As our work did not involve changing any of the underlying components of SMT systems, we have only touched upon the technicalities and computations involved in SMT. For a more complete and technical overview of all the components involved in language modeling, translation modeling and decoding, we refer the reader to: ``Statistical Machine Translation'' by \cite{Koehn2010}. 

By 2000, PB-SMT had become the state-of-the-art in MT~\citep{Zens2002,Koehn2003}. Although the PB-SMT approach provides a better way of dealing with the \emph{many-to-one} and \emph{one-to-many} mappings that occur in translations, it still has multiple drawbacks. Reordering within phrases, discontinuous phrases, the ability to learn across phrases (i.e. long-distance dependencies) or across sentences are just a few of them. Over the years, researchers worked on integrating additional knowledge and features into the existing framework. The integration of specific linguistic information in SMT will be further discussed in Section~\ref{sec:linguisticsMT}.

\subsection{Neural Machine Translation}\label{sec:related:neural_machine_translation}

More recently, NMT approaches have started to dominate the field of MT. Although the idea of using neural networks (NNs) for MT had already been explored in the 1990s~\citep{Castano1997,Forcada1997}, aside from the lack of sufficiently large parallel datasets, the computational resources were not powerful enough to deal with the complexity of the neural algorithms. The idea was abandoned and only resurged when~\cite{Schwenk2007} successfully applied a neural network language model to large vocabulary continuous speech recognition. The first `pure' NMT systems arrived with the convolutional~\citep{Kalchbrenner2013,Kalchbrenner2014} and sequence-to-sequence NMT models~\citep{Cho2014,Sutskever2014} which showed promising results but only for short sentences. By adding the attention mechanism~\citep{Bahdanau2014}, back-translated monolingual data~\citep{Sennrich2015b} and byte-pair-encoding (BPE)~\citep{Sennrich2016b}, NMT systems improved and quickly became state-of-the-art. \cite{Vaswani2017} present a model based on self-attention, revoking the complexity of Recurrent Neural Networks (RNNs) which further pushed the boundaries of the state-of-the-art in NMT.

The success of Neural Networks (NNs) and their popularity becomes clear when comparing the 2015 submissions for the WMT shared task (on MT), where one neural system was submitted but was still outperformed by the PB-SMT ones, while in 2017 the majority of the systems submitted were neural and most outperformed the more traditional PB-SMT models~\citep{Koehn2017neural}. 

Some of the main architectures and concepts relevant to the work we conducted in this thesis will be covered in the following paragraphs. These include: RNNs, Long Short-Term Memories (LSTMs)~\citep{Hochreiter1997}, attention and the Transformer architecture along with the concept of self-attention. By covering these concepts and architectures, we aim to provide information relevant to Chapter~\ref{ch:Aspect} and Chapter~\ref{ch:Loss} which contain experiments conducted on encoding vectors of RNNs and comparisons between different RNN and Transformer architectures in terms of lexical richness.

\paragraph{Encoder--Decoder Model}
One of the main bottlenecks with the application of NNs to MT (and other tasks related to NLP) had to do with a recurrent issue that language poses went it comes to computational models: its degree of randomness and the inability of computational models to account for it. In particular, input and output sequences\footnote{Sentences are often referred to as sequences in the field of NMT.} can be of variable lengths, might contain long-distance dependencies and exhibit complex alignments with each other. 

While the simple multilayer perceptron models could be used for MT, they cannot handle variable-length input and output sequences. 
The encoder--decoder model, however, uses two NNs. As such, it provides an architecture that can handle variable-length input and output sequences. It consists of two NNs
, the \textit{encoder} and the \textit{decoder}, where:

\begin{itemize}
 \item The \textit{encoder} NN encodes a variable--length input sequence
 $\bm{X} = \{ \bm{x}_1, \dots, \bm{x}_{n} \}$, $\bm{x}_t \in \mathbb{R}^{d_x}$, into
 a fixed-length vector representation $\bm{v} \in \mathbb{R}^{d_h}$; and
 \item The \textit{decoder} NN decodes the fixed-length encoded vector representation $\bm{v}$ into a variable-length output sequence
 $\bm{Y} = \{ \bm{y}_1, \dots, \bm{y}_{m} \}$, $\bm{y}_t \in \mathbb{R}^{d_y}$.
\end{itemize}

Note that the input sequence $\bm{X}$ and the output sequence $\bm{Y}$ of size \emph{n} and \emph{m}, respectively, allow for \emph{n} and \emph{m} to differ, while the internal representation $\bm{v}$ is fixed. 

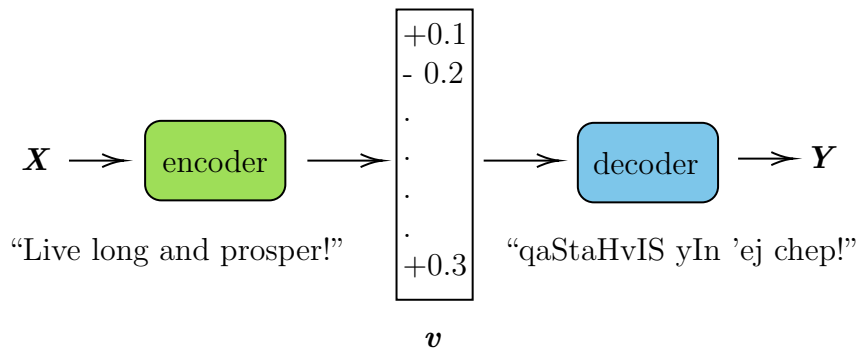
\begin{figure}[!ht]
\centering

\tikzset{every picture/.style={line width=0.75pt}} 

\begin{tikzpicture}[x=0.75pt,y=0.75pt,yscale=-1,xscale=1]

\draw  [fill={rgb, 255:red, 126; green, 211; blue, 33 }  ,fill opacity=0.75 ] (98,133) .. controls (98,128.58) and (101.58,125) .. (106,125) -- (160,125) .. controls (164.42,125) and (168,128.58) .. (168,133) -- (168,157) .. controls (168,161.42) and (164.42,165) .. (160,165) -- (106,165) .. controls (101.58,165) and (98,161.42) .. (98,157) -- cycle ;
\draw    (179,145) -- (214.5,145) ;
\draw [shift={(216.5,145)}, rotate = 180] [color={rgb, 255:red, 0; green, 0; blue, 0 }  ][line width=0.75]    (10.93,-3.29) .. controls (6.95,-1.4) and (3.31,-0.3) .. (0,0) .. controls (3.31,0.3) and (6.95,1.4) .. (10.93,3.29)   ;

\draw  [fill={rgb, 255:red, 80; green, 178; blue, 227 }  ,fill opacity=0.74 ] (314,134) .. controls (314,129.58) and (317.58,126) .. (322,126) -- (376,126) .. controls (380.42,126) and (384,129.58) .. (384,134) -- (384,158) .. controls (384,162.42) and (380.42,166) .. (376,166) -- (322,166) .. controls (317.58,166) and (314,162.42) .. (314,158) -- cycle ;
\draw    (267,145) -- (302.5,145) ;
\draw [shift={(304.5,145)}, rotate = 180] [color={rgb, 255:red, 0; green, 0; blue, 0 }  ][line width=0.75]    (10.93,-3.29) .. controls (6.95,-1.4) and (3.31,-0.3) .. (0,0) .. controls (3.31,0.3) and (6.95,1.4) .. (10.93,3.29)   ;

\draw   (223.5,69) -- (261.5,69) -- (261.5,215) -- (223.5,215) -- cycle ;
\draw    (394,144) -- (419.5,144) ;
\draw [shift={(421.5,144)}, rotate = 180] [color={rgb, 255:red, 0; green, 0; blue, 0 }  ][line width=0.75]    (10.93,-3.29) .. controls (6.95,-1.4) and (3.31,-0.3) .. (0,0) .. controls (3.31,0.3) and (6.95,1.4) .. (10.93,3.29)   ;

\draw    (60,145) -- (85.5,145) ;
\draw [shift={(87.5,145)}, rotate = 180] [color={rgb, 255:red, 0; green, 0; blue, 0 }  ][line width=0.75]    (10.93,-3.29) .. controls (6.95,-1.4) and (3.31,-0.3) .. (0,0) .. controls (3.31,0.3) and (6.95,1.4) .. (10.93,3.29)   ;

\draw (133,145) node  [align=left] {encoder};
\draw (348,146) node  [align=left] {decoder};
\draw (114,191) node  [align=left] {``Live long and prosper!''};
\draw (366,191) node  [align=left] {``qaStaHvIS yIn 'ej chep!''};
\draw (242.75,140) node  [align=left] {+0.1\\\mbox{-} 0.2\\.\\.\\.\\.\\+0.3};
\draw (44,144) node  [align=left] {\textbf{\textit{X}}};
\draw (436,143) node  [align=left] {\textbf{\textit{Y}}};
\draw (242,235) node  [align=left] {\textbf{\textit{v}}};

\end{tikzpicture}

\caption{An \emph{encoder--decoder} architecture consisting of three parts: the \emph{encoder} encoding the English input sequence $\bm{X}$ (``Live long and prosper!''), the fixed-length \emph{encoded vector} $\bm{v}$ generated by the \emph{encoder} and the \emph{decoder} generating the Klingon output sequence $\bm{Y}$ (``qaStaHvIS yIn 'ej chep!'') from $\bm{v}$.} \label{fig:encoder-decoder}
\end{figure}

A simplified visual representation of the \emph{encoder--decoder} architecture is given in Figure~\ref{fig:encoder-decoder}.

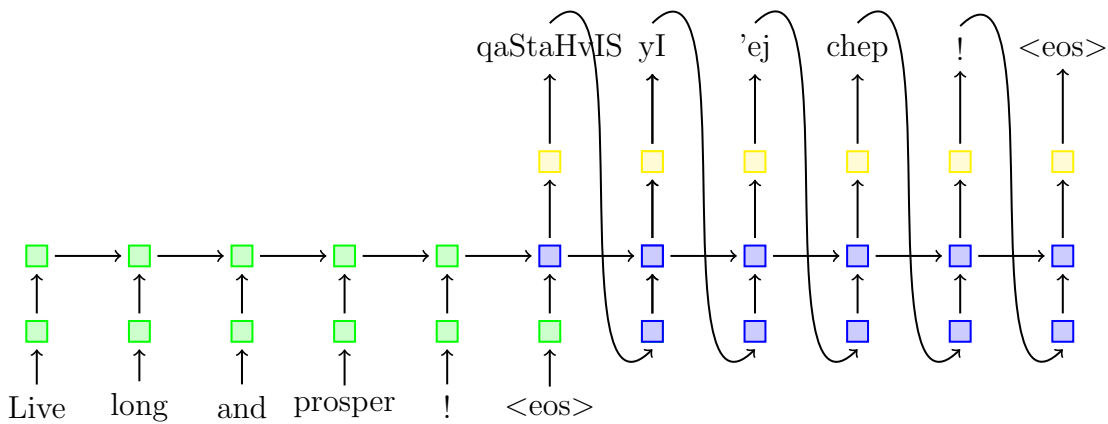
\begin{figure}[!ht]
\centering
\tikzset{every picture/.style={line width=0.75pt}} 

\begin{tikzpicture}[
    node distance=1cm,
  hid/.style 2 args={
    draw=#2,
    fill=#2!20,
    outer sep=1mm}]
  \foreach \i [count=\step from 1] in {Live, long, and, prosper, !, {{$<$eos$>$}}}
    \node (i\step) at (1.35*\step, -2) {\i};
  \foreach \t [count=\step from 6] in {qaStaHvIS, yI, 'ej, chep, !,{{$<$eos$>$}}} {
    \node[align=center] (o\step) at (1.35*\step, +2.75) {\t};
  }
  \foreach \step in {1,...,5} {
    \node[hid={3}{green}] (h\step) at (1.35*\step, 0) {};
    \node[hid={3}{green}] (e\step) at (1.35*\step, -1) {};    
    \draw[->] (i\step.north) -> (e\step.south);
    \draw[->] (e\step.north) -> (h\step.south);
  }
  \foreach \step in {6,...,7} {
    \node[hid={3}{yellow}] (s\step) at (1.35*\step, 1.25) {};
    \node[hid={3}{blue}] (h\step) at (1.35*\step, 0) {};
    \node[hid={3}{green}] (e\step) at (1.35*\step, -1) {};    
    \draw[->] (e\step.north) -> (h\step.south);
    \draw[->] (h\step.north) -> (s\step.south);
    \draw[->] (s\step.north) -> (o\step.south);
  }  
  
  \foreach \step in {7,...,11} {
    \node[hid={3}{yellow}] (s\step) at (1.35*\step, 1.25) {};
    \node[hid={3}{blue}] (h\step) at (1.35*\step, 0) {};
    \node[hid={3}{blue}] (e\step) at (1.35*\step, -1) {};    
    \draw[->] (e\step.north) -> (h\step.south);
    \draw[->] (h\step.north) -> (s\step.south);
    \draw[->] (s\step.north) -> (o\step.south);
  }  
  \draw[->] (i6.north) -> (e6.south);
  \foreach \step in {1,...,10} {
    \pgfmathtruncatemacro{\next}{add(\step,1)}
    \draw[->] (h\step.east) -> (h\next.west);
  }
  \foreach \step in {6,...,10} {
    \pgfmathtruncatemacro{\next}{add(\step,1)}
    \path (o\step.north) edge[->,out=45,in=225] (e\next.south);
  }
\end{tikzpicture}
\caption{The \emph{encoder--decoder} architecture with RNNs. The \emph{encoder} is shown in green and the \emph{decoder} in blue} \label{fig:rnnE}
\end{figure}

In NMT, the \emph{encoder} and \emph{decoder} are usually implemented with RNNs, most frequently using LSTM cells. An RNN can be viewed as stacked copies of identical networks. The input sequence is fed one token at a time, through one instance of the network. Its output is used alongside the following token and fed to the next instance of the network. Figure~\ref{fig:rnnE} reveals the chain-like structure of the RNN. When the special end-of-sentence symbol ($<$eos$>$) is reached, the decoding process is triggered. Taking Figure~\ref{fig:rnnE}, the English word `Live' is passed through the first instance of the identical networks, its output is combined with the information of the next word `long' and fed through the next identical network. As such, when the last token, here `!', is reached, the values of the nodes of the final hidden layer contain information on all the previous tokens: ``Live long and prosper!''. The $<$eos$>$ symbol will trigger the decoding process. The first Klingon word `qaStaHvIS' is generated by the decoder and used as an input for the next network.



The RNN approach can be applied successfully to MT. However, these models only work well for relatively short sentences and fail for longer ones. During the encoding phase, the hidden state needs to remember all the information of the input sentence. During decoding, it not only needs to encode information in order to correctly predict each word, but it also needs to keep track of what parts have already been covered and what still needs to be translated. As such, the hidden layer has to simultaneously serve as the memory of the network and as a continuous space representation used to predict output words~\citep{Koehn2017neural}. However, not all context words are always equally important when predicting specific words. 

In Example~(\ref{ex:longDist}), the verbs \emph{are} and \emph{make} agree with \emph{my parents}. However, for \emph{are} it is clear that the immediately preceding word is very important, while for \emph{make} we have to be able to look further back. In the network, the hidden state is always updated with the most recent word, so predicting \emph{are} correctly would not be too hard for the network. However, the hidden state's memory of words it has seen multiple steps before that decreases over time. As such, predicting \emph{make} would result to be a more difficult task.

\begin{li}
\label{ex:longDist}
\item ``\textbf{My parents} [\textit{are}] very busy but always [\textit{make}] time for me.''
\end{li}

\paragraph{Long Short-Term Memory}
To address the aforementioned issue related to the memory of the hidden states, LSTM cells were introduced into the RNN~\citep{Hochreiter1997}. The LSTM is a special kind of a cell composed of three gates: the input gate, the forget gate and the output gate. Those gates allow the LSTM to deal with long-term dependencies as in Example~(\ref{ex:longDist}). 
Unlike a simple RNN, it is able to regulate the information flow and has the ability to remove or add certain information to the cell state regulated by its gates~\citep{Koehn2017neural}. Next, we briefly explain the gates of the LSTM and their functions:

\begin{itemize}
    \item Forget gate: The forget gate decides what information to keep and what information to throw away. With respect to our Example~(\ref{ex:longDist}), passing through the sentence, the forget gate might decide to get rid of some of the irrelevant information in between \emph{my parents} and the second verb \emph{make}, but it should retain the information on the number of the subject as this information will be needed to generate a correct subject-verb agreement. 
    \item Input gate: The input gate decides what new information should be stored in the cell. This new information is then merged with the `old' memory that made it through the forget gate, creating a new cell state. 
    \item Output gate: The output gate controls how strongly the memory state is passed to the next layer, i.e. what the next hidden state should be.
\end{itemize}

In short, the LSTM cell considers the current input, previous output and previous memory to then generate a new output and alter the memory. An alternative to LSTMs are Gated Recurrent Units (GRUs)~\citep{Cho2014}, which are also widely used to deal with memory issues in RNNs. They are very similar to LSTMs but unlike the LSTMs, a GRU only uses two gates, a reset and and update gate. Both GRUs and LSTMs are used in NMT, although LSTMs seem to be more common. The NMT systems we trained for our experiments all use LSTM cells. 

\paragraph{Attention}
Incorporating LSTM cells into an RNN NMT model alleviates some of the memory-related issues. However, the fixed-size hidden state(s) need to encode the entire source sentence and retain all the important elements. This observation led to the already famous plain-spoken statement by Ray Mooney during an ACL workshop in 2014:

\begin{displayquote}
``You can’t cram the meaning of a whole \%\&!\$ing sentence into a single \$\&!\*ing vector!''
\end{displayquote}

Furthermore, when generating the translation, not every part of what has been encoded is equally relevant at every step of the translation process. When a translator translates a sentence, they look back and focus on different parts of the source sentence for specific sections of the target sentence. So far, the RNN (with LSTMs) we have described has no way to \textit{attend to} or look back at specific parts of the source sentence while generating its translation. Take the example sentence in~(\ref{ex:longDist2}). We established before that it is important to remember the number of the subject in order to predict the correct verb forms \emph{are} and \emph{make}. However, this information is relevant for those words that agree with the subject, but less so (or even completely irrelevant) when predicting other words such as \emph{very} or \emph{but}.  

\begin{li}
\label{ex:longDist2}
\item ``\textbf{My parents} [\textit{are}] \underline{very} busy \underline{but} always [\textit{make}] time for me.''
\end{li}

To alleviate the aforementioned issue where all the source-side information needs to be compressed into a fixed-sized hidden layer, \cite{Bahdanau2014} introduced an attention mechanism encoder--decoder framework. The attention mechanism allows the decoder to have access to all the hidden states that were generated by the encoder at every time step. Instead of squeezing all the information into the fixed-sized vector, the input sequence is now encoded into multiple vectors. The decoder can then attend to, or choose, a subset of these vectors while decoding specific parts of the translation. This particularly helps the NMT systems deal with longer sentences, which had proven to degrade the quality of NMT systems considerably~\citep{Cho2014}. The attention mechanism is somewhat comparable to the alignments in SMT.

With this new approach presented in~\cite{Bahdanau2014}, their NMT system for English--French obtained results comparable to the state-of-the-art PB-SMT systems.

Most of the NMT systems we trained to conduct experiments consisted of RNNs with LSTMs and an attention mechanism. However, in Chapter~\ref{ch:Loss}, we included some experiments with the Transformer architecture. 

\paragraph{Self-Attention Networks}
Self-attention networks gained popularity in NMT following the ``Attention is all you need'' paper by~\cite{Vaswani2017} from Google Brain. Self-attention networks differ considerably from the previous NMT approaches presented as they do not use RNNs (LSTMs or GRUs) nor Convolutional Neural Networks (CNNs). The best-known self-attention network for NMT is the Transformer architecture which is based solely on attention mechanisms. 

The Transformer architecture extends the idea of attention by using self-attention. The idea behind attention was to consider associations between input and output words. Self attention extends this idea individually to the encoder and the decoder~\citep{Koehn2017neural}. As such, it is related to the associations between the input words themselves. Consider the sentence in Example~(\ref{ex:transf}). A self-attention mechanism would refine the representation of the word `race'. In this particular example, a word such as `human' could receive a high attention score when constructing the representation of the word `race' as it helps to disambiguate the otherwise ambiguous word `race'.

\begin{li}
\label{ex:transf}
\item ``I believe there is only one race -- the human race.''\footnote{Quote from Rosa Parks.}
\end{li}

Similar to the encoder, the decoder will also attend to specific previously generated words in order to make better informed decisions. Furthermore, aside from using what it has translated already, it will attend to what has been encoded.

Every layer in the encoder and the decoder contains a fully connected feed-forward network. A feed-forward NN differs from a RNN as it does not allow information to flow in both directions (or loops). They are bottom-up networks where information from an input is associated with an output and propagated through a network. As feed-forward NNs are less complex than RNNs or CNNs, the Transformer architecture allows for faster training. With their novel approach, \cite{Vaswani2017} achieved new state-of-the-art results for English--French and English--German on the WMT 2014 datasets. 

We have presented the two main NMT architectures used in the experiments presented in this thesis: RNNs with LSTMs and attention and the Transformer architecture consisting of self-attention layers and fully connected feed-forward NNs. We provided a very high-level overview of how these NMT architectures evolved and how different components were added over time to deal with NMTs main shortcomings. For a more complete overview including CNNs, GRUs, Feed-forward NN, neurons, including the internal working and mathematics involved in the computations, we refer to~\cite{Koehn2017neural}. Aside from the state-of-the-art architectures, we employed two techniques commonly used to overcome NMT's limitations: BPE and Back-Translation.

\paragraph{Byte-Pair Encoding}
One of the shortcomings of NMT is its inability to deal with large vocabularies. The vocabulary is typically fixed to 30,000--50,000 unique words as an open vocabulary would be too computationally expensive.\footnote{The internal representation would not fit into memory if the vocabulary size crosses a certain upper-bound which depends on the architecture and the allocated memory.} This limitation is problematic for a translation task, especially for morphologically rich or agglutinative languages. Word-level NMT models would address the issue by backing-off to a dictionary look-up~\citep{Jean2015}, but, such approaches would rely on assumptions (like a \emph{one-to-one} correspondence between source and target words) that do not always hold up (see the \emph{one-to-many} and \emph{many-to-one} alignments discussed in Section~\ref{sec:related:statistical_machine_translation}, Figure~\ref{fig:onetomany}). \cite{Sennrich2016b} propose working with subword units instead of words in order to model out-of-vocabulary (OOV) words. They adapt the BPE algorithm~\citep{Gage1994} for word segmentation and merge frequent pairs of characters or character sequences. It is important to note that this method is purely based on occurrences of characters. Thus, the so-called `subwords' are not linguistically motivated.\footnote{They can overlap with morphemes as they appear frequently throughout training corpora.} An example of BPE operations on a toy dictionary is given in Figure~\ref{fig:BPE}.

\begin{figure}[h]
\centering
\begin{tabular}{lcl}
    l o  & $\rightarrow$  &    lo     \\  
    lo w & $\rightarrow$  &    low    \\
    e r  & $\rightarrow$  &    er     \\
\end{tabular}\caption{BPE operations on a toy dictionary \{`low',`lowest', `newer', `wider'\}~\citep{Sennrich2016b}.}\label{fig:BPE}
\end{figure}
As can be observed in Figure~\ref{fig:BPE}, BPE subwords could overlap with linguistic morphemes as especially derivational or inflectional morphemes are character sequences that tend to appear frequently in datasets. However, there is no guarantee that the BPE subwords will overlap with linguistic units, as the segmentation depends on the character sequences observed in the training data and the amount of BPE operations conducted. A non-linguistically motivated  BPE segmentation is given in Figure~\ref{fig:BPE2} where `stormtroopers' is split into 5 BPE units `stor', `m', `tro', `op' and `ers'.

\begin{figure}[h]
\centering
\begin{tabular}{ll}
    Input   & stormtroopers     \\  
    BPE:    & stor m tro op ers     \\
\end{tabular}\caption{BPE subwords of `stormtroopers'~\citep{vanmassenhove2018supernmt}.}\label{fig:BPE2}
\end{figure}

\paragraph{Back-Translation}
In PB-SMT, monolingual data would be used by the language model in order to improve the fluency of the generated translations by the translation model. The NMT architectures initially did not have any way of integrating additional monolingual data. Back-translation~\citep{Sennrich2015b,Poncelas2018} is not only a popular approach to leveraging monolingual datasets, but also to generate more training data for low-resource languages as it has been identified that NMT only performs well in high-resource settings~\citep{Koehn2017six}. The back--translation~\citep{Sennrich2015b} pipeline involves the following steps:

\begin{itemize}%
\item[] (i) Collect monolingual target-language data.
\item[] (ii) Train a reverse system that translates from the intended target language into the source language with the same setup as the final NMT system.
\item[] (iii) Use the trained system from (ii) to translate the target monolingual data into the source language.
\item[] (iv) Finally, combine the synthetic parallel data generated with the back-translation pipeline with the original parallel data for the final NMT system.
\end{itemize}

\noindent We conduct experiments using back-translation in Chapter~\ref{ch:Loss}.

\subsection{Automatic Evaluation Metrics}\label{sec:evaluationMT}
Both PB-SMT and NMT quality is most frequently measured using automatic evaluation metrics. These metrics compare translations generated by MT systems with reference translations in terms of \emph{n}--gram overlap. Greater overlap is correlated with a higher score and arguably a better translation quality. Such metrics have several shortcomings that are well-known in the community, but few reasonable alternatives are available~\citep{Hardmeier2014}. BLEU can be considered the standard automatic evaluation metric within the field of MT. It is computed by comparing the overlap of \emph{n}--grams (usually of size 1 to 4) between a candidate translation and the reference(s). First, the \emph{n}--gram precision is calculated by comparing the candidate translation with the reference translation(s) in terms of \emph{n}--gram overlap divided by the total number of \emph{n}--grams in the candidate. Second, the recall is measured by incorporating a brevity penalty which punishes candidate translations shorter than the reference(s). Other alternative metrics include METEOR~\citep{Banerjee2005} and TER~\citep{Snover2006}. METEOR computes unigram matches not only based on the words but also on their stems. TER calculates the amount of editing that would be required in order to match a candidate with its reference translation.

These metrics address some of the issues with BLEU, but none of them overtook BLEU as the standard metric. Although we attempted to conduct (semi-) manual evaluations in most of our chapters to corroborate our findings, for practical reasons and comparability, we frequently relied on automatic evaluation metrics, specifically BLEU. Our motivation behind this is that we were often interested in specific endings (e.g. `heureux' vs `heureuse' for gender agreement) and the BLEU metric does look for \emph{exact} matches. As such, one of BLEU's main shortcomings (the fact that it requires exact matches which is often undesirable when evaluation translation output), was for our purposes sometimes more an asset than a shortcoming. Still, BLEU would not be able to deal with synonymy nor would it have a notion of error gravity.\footnote{Some errors are perceived to be more serious to readers than others~\citep{Vann1984,Lommel2014}} Therefore, aside from testing our approaches on general test sets, we furthermore relied on more specific test sets containing the relevant phenomena.

\section{Linguistics in Machine Translation}\label{sec:linguisticsMT}

In this section we will provide a brief overview of the main research on integrating linguistics into SMT (Section~\ref{subsec:smt}) and NMT (Section~\ref{subsec:nmt}). 

In the last 25 years, data-driven approaches to MT have been demonstrated to produce better quality output than RBMT systems for most language pairs. Many PB-SMT systems have evolved to be hybrid and include some linguistic knowledge. While PB-SMT practitioners acknowledge now that integrating linguistic knowledge is useful, with the arrival of NMT the role of linguistic information for MT was initially questioned once again. However, some recent papers (e.g. \cite{Sennrich2016}) have already demonstrated the usefulness of integrating linguistic knowledge in NMT. 

Many MT researchers recognise that all systems have their drawbacks and limitations and that future models should aim to combine their strengths into `hybrid' models \citep{Hutchins2007}. On the one hand, rule-based systems generate more grammatically correct output at the morphological level but make poor semantic choices. On the other hand, PB-SMT performs well with respect to the semantic aspect of translation but due to fact that the basic model exploits only \emph{n}--gram sequences, morphological agreement and word order remain problematic \citep{Costa2012study}. Recent studies have shown that NMT partially overcomes some of these issues but the handling of longer sentences as well as more intricate linguistic phenomena that require a deeper semantic analysis remain problematic \citep{Bentivogli2016}.  One simple, yet classical example sentence of ambiguity and its translations produced by Google Translate's NMT system (GNMT)\footnote{These results were produced by Google Translate (\url{https://translate.google.com/}) in June 2019.} illustrates such a shortcoming: \

\begin{li}
\label{duck1}
\item Source:  Somebody was shooting bullets and we saw {\em her duck}.
\item EN--FR:  Quelqu'un tirait des balles et nous avons vu {\em son canard}.
\end{li}
 
 Although the sentence in (\ref{duck1}) is ambiguous, most (if not all) translators would not hesitate to translate `duck' as a verb instead of a noun (although the noun is, in general, more common)\footnote{According to a frequency search of the Corpus of Contemporary American English (COCA) \citep{davies2008} available online at \url{http://corpus.byu.edu/coca/}, containing 520 million words, `duck' appears 9199 times as a noun while only 3333 times as a verb.} because they process the sentence in a semantico-syntactic way. However, GNMT for English--French, translates `duck' in this particular context as a noun. Although this is technically not incorrect, it is a very unlikely translation given the first part of the sentence. It is hard to give a concrete analysis or explanation on why the GNMT system decided to opt for the semantically least likely option.\footnote{We can however assume it is related to the frequency of `duck' appearing as a noun or a verb in the training data used for the models.} The hidden layers in a neural network represent the learning stages of the system but this knowledge is encoded in such a way that it is currently very difficult for humans to infer anything from them. This makes it hard to identify the exact cause of the problem as well as a remedy for it. We would like to point out that we used the same example sentences back in 2017.\footnote{April 2017.} Back then, the same issue (as in Example~\ref{duck1}) occurred when translating into Spanish and Dutch. However now in 2019,\footnote{June 2019.} the Dutch and Spanish translations opt for the more likely translation of `duck' as a verb. The example, however, illustrates how even GNMT still struggles with encoding the entire meaning of a sentence, although semantics is claimed to be NMT's strongsuit.
 
Another example we encountered that illustrates well how GNMT can be very inconsistent is given in Example~(\ref{ex:rain}):

\begin{li}
\label{ex:rain}
\item Source: I trained.
\item EN--FR: Il pleuvait.
\item EN--ES: Llovi\'o.
\item EN--NL: Het regende.
\end{li}

Although NMT is often able to produce good translations for long and complicated sentences, Example~(\ref{ex:rain}) shows how a very easy and unambiguous sentence can pose difficulties. All translations have translated `I trained.' into a French, Dutch and Spanish sentence that we can translate back to English as `It rained.'. As we have little insights as to how GNMT works exactly, we assume the subword units are the underlying cause for this segmentation mix-up. This is an error that a human translator, rule-based or PB-SMT system would never make.

One last example translation, presented in Example~(\ref{ex:belle}), illustrates how a relatively short and simple sentence `We are very beautiful.' is translated incorrectly into the French sentence `Nous sommes tr\`es belle.'. The English pronoun `We' is plural and not marked for gender. The word `beautiful' is in agreement with the subject and in French this agreement is marked explicitly. The translation fails to make this agreement as the word `belle' is singular instead of plural. Note also how GNMT opts for the female variant of the word `beautiful' in French (`belle'). 

\begin{li}
\label{ex:belle}
\item Source: We are very beautiful.
\item EN--FR: \emph{Nous} sommes tr\`es \textbf{*belle}.
\end{li}


As the example translations have shown, there are many linguistic issues remaining. In the two following sections, we briefly describe the most common techniques used to enhance PB-SMT and NMT systems with linguistic features. 

\subsection{Statistical Machine Translation}\label{subsec:smt}
PB-SMT~\citep{Koehn2007} learns to translate phrases of the source language to target-language phrases based on their co-occurrence frequencies in a parallel corpus. Usually, additional monolingual data is used to improve the fluency of the produced translations. All source-language phrases and their target-language counterparts are stored in phrase-tables together with their probabilities. In a PB-SMT system, every phrase is seen as an atomic unit and thus translated as such. Given a source sentence $F$, the system aims to find a translation $E^*$ in the target language so that:
\begin{equation}
E^* = \argmax_{t} p(E|F)
\\  = \argmax_{t} \frac{p(F|E)p(E)}{p(F)}
\\  = \argmax_{t} p(F|E)p(E)
\end{equation} 
\noindent where $p(F|E)$ (the translation model probability) is estimated using bilingual data, and $p(E)$ (the language model probability) is estimated based on monolingual data.
\label{LingSMT}

Linguistic information has been integrated into SMT systems over the last 20 years in various ways resulting in different types of `hybrid' systems (e.g. \cite{Avramidis2008}, \cite{Toutanova2008}, \cite{Haque2010}, \cite{Marevcek2011}, \cite{El2012}, \cite{Fraser2012}, etc.). Lemmas, stems, part-of-speech (POS) tags, parse trees etc. can be integrated by pre- and/or post-processing the data. Since a substantial part our work focuses on morphology, we will give an overview of the most common techniques used specifically to integrate morphological information.

Morphological agreement rules are language-dependent and become increasingly difficult to `learn' for a PB-SMT system when source and target languages have significantly different morphological structures. Languages that are not morphologically rich such as English where, for example in the present tense, only the third person singular (infinitive +s) can be distinguished from the others by looking at its surface form, are particularly hard since one verb form in English can be matched with several verb forms in (say) French, as Table \ref{verbs} illustrates.

\begin{table}
\begin{center}
\begin{tabular}{ |c|c| } 
 \hline
 see & vois, voyons, voyez, voient, voir\\ 
 \hline
 sees & voit  \\ 
 \hline
\end{tabular}
\end{center}
\caption{Single English surface verb forms mapping to multiple French
  verb forms}\label{verbs}
\end{table}

Ueffing \& Ney (\citeyear{Ueffing2003}) were one of the first to enrich the English source language to improve the  correct selection of a target form when still working with WB-SMT. By using POS-tags, they spliced sequences of words together (e.g. `you go' $\rightarrow$ `yougo') to provide the source form with sufficient information to translate it into the correct target form. By introducing phrase-based models for SMT, this particular problem of WB-SMT \emph{seemed} to be largely solved. However, the language model statistics are sparse and due to an increase in morphological variations they become even sparser which can cause a PB-SMT system to output sentences with incorrect subject-verb agreement even when subject and verb are adjacent to one another. Syntax-based MT models tend to produce translations that are linguistically correct, although the syntactic annotations increase the complexity which leads to slower training and decoding. 

Within the field of PB-SMT, several works have focused on dealing with problems specific to translations into morphologically richer languages. Generally, those works focus on improving PB-SMT by: (i) source-language pre-processing \citep{Avramidis2008,Haque2010}, and (ii) a combination of both pre-processing of the source language and post-processing of the target language \citep{Virpioja2007,Marevcek2011,El2012,Fraser2012}. \cite{Avramidis2008} added per-word linguistic information to the English source language in order to improve case agreement as well as subject-verb agreement when translating to Greek and Czech. To improve subject-verb agreement they identified the person of a verb by using POS-tags and a parser. The information of the person was added to the verb as a tag containing linguistic information. Their initial system suffered from sparsity problems which led to the creation of an alternative path for the decoder with fewer (or no) factors. Although there were no significant improvements in terms of BLEU score, manual evaluation revealed a reduction in errors of verb inflection. \cite{Haque2010} presented two kinds of supertags to model source-language context in hierarchical PB-SMT: those from lexicalized tree-adjoining grammar  and combinatory categorial grammar. With English as a source language and Dutch as the target language, they reported significant improvements in terms of BLEU.  

Other research has focused on both pre- and post-processing the data in a two-step translation system. This implies, in a first step, simplifying the source data and creating a translation model with stems \citep{Toutanova2008}, lemmas \citep{Marevcek2011,Fraser2012} or morphemes \citep{Virpioja2007}. In a second step, an inflection model tries to re-inflect the output data. In \cite{Toutanova2008}, stems are enriched with annotations that capture morphological constraints applicable on the target side to train an English--Russian translation model, with target forms inflected in a {\em post hoc} operation.  Two-step translation systems working with lemmas instead of stems were presented in both Mare{\v{c}}ek et al.~(\citeyear{Marevcek2011}) and \cite{Fraser2012}. While Mare{\v{c}}ek et al.~(\citeyear{Marevcek2011}) perform rule-based corrections on sentences that have been parsed to dependency trees for  English--Czech, \cite{Fraser2012} use linear-chain Conditional Random Fields to predict correct German word forms from the English stems. Opting for a pre- and post-processing step is necessary when language-specific morphological properties that indicate various agreements are missing in the source language \citep{Marevcek2011}. Note that all the methods described above require (a combination of) linguistic resources such as POS-taggers, parsers, morphological analyzers etc., which may not be available for all language pairs.

\subsection{Neural Machine Translation}\label{subsec:nmt}

NMT~\citep{Cho2014,Sutskever2014,Bahdanau2014} encodes the entire source sentence in a single encoding vector. As described in Section~\ref{subsec:nmt}, in an encoder--decoder NMT model, there are two neural networks at work: the first encodes information about the source sentence into a vector of real-valued numbers (the hidden state), and the second neural network decodes the hidden state into a target sentence. Unlike PB-SMT, the neural network responsible for the decoding of the hidden state has access to a vector that contains information about the entire source sentence. This should allow NMT to handle certain linguistic phenomena better than PB-SMT (e.g. long-distance dependencies within sentence boundaries). Indeed, a detailed comparison by \cite{Bentivogli2016} revealed that NMT does not only generate outputs that require for certain tasks less efforts from post-editors when compared to PB-SMT systems, it also outperforms PB-SMT on all sentence lengths and has an advantage when it comes to translating lexically richer texts. Furthermore,  \cite{Bentivogli2016} analyzed the types of errors made by PB-SMT and NMT systems and concluded that NMT made fewer morphological, lexical and word-order errors compared to PB-SMT. However, NMT's performance degrades faster with the input length compared to PB-SMT and, as noted in the introduction, some seemingly `easy' problems as well as some more intricate problems remain.



Given the novelty of the field of NMT, relatively little research has been done to date on the incorporation of linguistic information in NMT.  The fact that NMT systems perform comparably to other systems relying on nothing more than characters \citep{Lee2016} and a study observing that the NMT encoder automatically extracts syntactic categories from the input \citep{Shi2016} question to some extent the benefits of integrating explicit linguistic information. However, a couple of studies show promising results with respect to the use of linguistics in NMT. Linguistic information can be integrated in an NMT system explicitly (e.g. \cite{Eriguchi2016}; \cite{Sennrich2016}) or implicitly \citep{Eriguchi2017} to the source or target side \citep{Garcia2016}.

\cite{Sennrich2016} integrate linguistic features by generalizing the embedding layer of the encoder in such a way that arbitrary features can be included. They add morphological features (POS-tags and dependency labels) to the source side as features for the English--German and English--Romanian language pairs. They show that linguistic input features improve the quality of NMT models in terms of perplexity, BLEU and CHRF3 \citep{popovic2015chrf}. \cite{Eriguchi2016} outperform a sequence-to-sequence attentional English--Japanese NMT model by integrating syntactic information. They extend an attentional NMT system with phrase structure information on the source side. During decoding, an attention mechanism allows the generation of translated words by softly aligning them with words and phrases of the source. \cite{Eriguchi2017} implicitly integrate linguistic information into their NMT system by designing a hybrid decoder. Apart from the usual conditional language model, their decoder relies also on an RNN grammar \citep{Dyer2016}, which is designed to model hierarchical relations between words and/or phrases. They conducted experiments for four language pairs (Czech, German, Japanese and Russian into English) and obtained significant BLEU score improvements for all language pairs except for German--English. \cite{Nadejde2017} added syntactic information to the source or target language in the form of CCG supertags. Their syntactically-enriched NMT models improved the baseline NMT systems for Romanian--English and German--English. They note as well that a tight coupling of words and syntax outperforms multitask training. Unlike the above-mentioned research, \cite{Garcia2016} use linguistic features in the form of factors (e.g. morphological and/or grammatical decomposition of the words) in the output side of their English--French NMT system. In order to do so, they added an attention mechanism so that two outputs can be generated: (1) the lemmas and (2) the remaining factors. Although their experiments show that the factored NMT system does not always outperform the baseline system, the factored NMT system does reduce the number of OOV words and can handle a much larger vocabulary \citep{Garcia2016}.

\cite{Shi2016} show that the sentence vectors of an English--French and English--German NMT systems `encode' (or maybe better `preserve') syntactic information, which might indirectly suggest that linguistic information is superfluous. However, their detailed analysis also reveals that not all the necessary subtleties are encoded in the sentence vectors. Their method consists: (1) of labeling the original source sentences with syntactic labels, (2) learning the sentence encoding vectors with an NMT system, and (3) trying to predict the syntactic labels of the source sentences with a logistic regression model trained on NMT sentence-encoding vectors. 

Other research focused more on understanding the type of linguistic knowledge encoded in sentence embeddings includes the work of~\cite{Belinkov2017,Belinkov2017b}. Similar to \cite{Shi2016}, \cite{Belinkov2017} shed light on NMT's ability to capture morphology by training a classifier on features extracted from the internal representations of English$\Leftrightarrow$German and English$\Leftrightarrow$Czech models. Their work focuses on POS-tags and morphological tagging. They compared the influence of different types of representations and the depth of the layers used to predict certain morphological features or POS-tags. Their main observations were the following: (i) for morphology, character-based representations produced better results than word-based ones, (ii) the lower layers of the NN were better at capturing morphology while deeper layers led to better translations. They hypothesize that the lower layers focus more on surface phenomena while the deeper layers are able to abstract better and can better grasp the overall meaning of what is encoded, (iii) when the target language was morphologically-poor, the source-side representations would be better for predicting POS-tags or morphology and (iv) the representations in the attentional decoder contained little information on morphology. In a follow-up paper ~\citep{Belinkov2017b}, where they train models from English to Arabic, Chinese, French, Spanish and Russian, they conclude that the deeper layers are better at learning semantics, while lower layers tend to be better for POS-tagging. Furthermore, although they observed in~\cite{Belinkov2017} that a morphologically poor target would lead to better source representations, they now observe little effect of the target language on source-side representations (when working with higher quality NMT models). They hypothesized this might be the case because more training data was used in \cite{Belinkov2017b} and confirmed this by repeating their experiments on a smaller dataset.

\section{Bias in Artificial Intelligence}\label{sec:diversityMT}

Recently, bias in AI has rightfully gained a considerable amount of attention, not only in the research community but also in the media as the scope of AI applications has been growing. Machine-learning architectures like SMT and NMT (described in Section~\ref{subsec:smt} and Section~\ref{subsec:nmt}) learn by maximizing overall prediction accuracies. As such, the algorithms learn to optimize over more frequently appearing patterns or observations. If a specific group of individuals appears more frequently than others in the training data, the program will optimize for those individuals because this boosts their overall accuracy~\citep{Zhou2018}. Computer scientists evaluate algorithms on test sets, but typically these are random sub-samples of the original training set and thus likely to contain the same biases as observed during training. As such, biased behaviour is rewarded rather than punished, consequently the outputs we create show a lack of diversity on multiple levels. In Chapter~\ref{ch:Loss} we discuss the loss of linguistic diversity in more detail and relate this to the algorithmic bias observed.

In Chapter~\ref{ch:Gender}, we zoom in on issues related to gender agreement in MT. Our work focuses on morphological agreement by incorporating gender features into an NMT pipeline. The analysis we conducted prior to our experiments related to gender revealed that Europarl~\citep{Koehn2005} has a 2:1 male-female speaker ratio. As some languages express gender agreement with the speaker, this can lead to a higher frequency of male pronouns or male-endings for nouns. This, in turn, can influence the translations and lead to exacerbation of the observed phenomena as statistical approaches learn by generalizing over the seen patterns. Similar to our observation, recent studies~\citep{garg2018,Lu2018,Prates2018} have highlighted issues with biased training and testing data and some already alluded that there might be an exacerbation of the observed biases~\citep{Lu2018,zhao2018learning} by the algorithms themselves. The systematic bias problem extends to a range of AI applications. This is particularly problematic as one of the reasons for employing such applications is the fact that they ought to be more objective than humans. We dedicate a section to bias and related issues as we should strive towards fair algorithms that do not sustain or worsen observed data biases. 


Because of the fact that bias is relatively well hidden in MT, it did not receive a lot of attention until recently. In order to find examples of bias in MT one would need to find a sentence that is ambiguous in the source, but unambiguous in the target. This ambiguity arises when one language makes something explicit which is left implicit in the other language. An example could be the implicit natural gender of a speaker in a language like English compared to the explicit natural gender markers in a language such as French. For example `I am happy' is not marked for gender in English. Its French translation, however, requires the translator to pick between `Je suis heureux' (male) or `Je suis heureuse' (female). Similary, the word `sister' in Basque, would, depending on the gender of the person whose sister is referred to, be translated into `arreba' (male) or `ahizpa' (female). When no gender is explicitly mentioned in the source text, most of these choices could be interpreted as being rather innocent. However, the mere fact that we do not exactly know or control these endings is problematic. For example, when passing a list\footnote{\url{https://www.frenchtoday.com/blog/french-verb-conjugation/french-reflexive-verbs-list-exercises/}. May 2019.} of 111 French reflexive verbs\footnote{We used the reflexive verbs as they often mark for gender.} through Google Translate, none of them received a female ending. This corresponds to what we have observed before, i.e. the male endings can almost be considered the default form in Google Translate. As such, the female endings are somehow already marked because of the fact that they do not appear frequently. However, when translating Example~(\ref{genderRaped}), the verb `viol\'ee' (raped) has a female ending. 

\begin{li}
\label{genderRaped}
\item `I was raped' 
\item `J'ai \'et\'e viol\'e\textbf{e}'
\end{li}

These `uncontrolled' fluctuations between male and female endings depend on the training data reflecting conscious and unconscious biases present in our day-to-day communication, but also on the further generalizations made by the algorithms themselves. Users might not always be aware of this when using MT systems, and currently there is nothing to notify the user about the assumptions the algorithms have made. For users that do grasp the target language well enough, we understand how `marked'\footnote{With the male forms being the default, any female ending becomes marked.} translations such as Example~(\ref{genderRaped}) can be considered inappropriate or even offensive, especially when claims about reaching human parity are becoming commonplace~\citep{Hassan2018,Laubli2018,Toral2018}.




We address the issue of gender bias to some extent with the integration of gender features in NMT. Nevertheless, our approach is still limited, and more research needs to be conducted in this direction. Until an appropriate way of handling gender agreement, controlling it, and presenting it to users has been proposed, it is important that researchers and users are aware of these issues and that the necessary checks are put in place. 

\section{Conclusion}\label{sec:conclusionsMT}
In this chapter, we provided background information on the SMT and NMT models used throughout our thesis. We covered some of the basic concepts and architectures and provided a high-level comparison of their internal architectures to shed light on their shortcomings and strengths. We discussed general work on the integration of linguistic information into SMT and NMT and alluded to some of the problems that remain. We furthermore linked some of the morphological issues to broader ethical ones and briefly addressed bias in AI. In the next chapter, we will elaborate on our first set of experiments related to the integration of linguistic features into SMT and compare SMT's performance to NMT with respect to subject-verb number agreement.

\chapterbib


%% file: SVAgreement.tex
~\newpage
\epigraph{If no agreement, \\ then separation.}{\textit{Arabic Proverb}}

\chapter{Subject-Verb Number Agreement in Statistical and Neural Machine Translation}\label{ch:Agreement}

In this chapter, a simple method for dealing with subject-verb number agreement issues in MT will be discussed. When this work was conducted, the main paradigm in the field of MT was still PB-SMT. An initial exploration of the outputs produced by PB-SMT systems showed that even very basic agreement issues between subject and verb are often still problematic to handle. Subject-verb agreement rules are basic and systematic grammar rules that are relatively easy for humans to learn. PB-SMT's inability to correctly deal with such seemingly easy grammatical rules that harm the overall quality of the translation, was the motivation behind our first set of experiments.

The main part of this chapter will thus focus on dealing with subject-verb number agreement in PB-SMT. However, for completion, we have also included the results produced by NMT systems automatically and manually evaluated on the same test sets as the PB-SMT systems. This chapter is related to RQ2 as we identify gaps in the necessary linguistic knowledge available to the PB-SMT system and integrate additional information in order to improve its performance.

\section{Introduction}

Ensuring correct agreement between a subject and a verb is a relatively straightforward task for humans. The agreement between the subject of a sentence and the main verb is crucial for the correctness of the information that a sentence conveys as failing to do so can alter the meaning of a sentence, as illustrated in~(\ref{adequacy}).  

\begin{li}
\label{adequacy}
\item \hspace{10mm} [The men and the boy who \textbf{are} laughing] are family!
\item \hspace{10mm} [The men and [the boy who \textbf{is} laughing]] are family!
\end{li}

Subject-verb agreement unifies the sentence from a syntactic point of view but also semantically. As illustrated in~(\ref{adequacy}), both sentences are correct, but, they have different meanings. In the first sentence, both `the men and the boy' are laughing, while in the second one, only `the boy' is.

Disagreement between subject and verb may lead to ambiguity which affects the overall adequacy and fluency of a sentence considerably. RBMT produces translations that are syntactically better than those produced by PB-SMT, where very obvious errors such as lack of number and gender agreement can occur, but RBMT systems tend to have problems with lexical selection and fluency in general. Their success, however, heavily relies on the accuracy of advanced linguistic resources such as syntactic parsers that may fail, propagate errors and are unavailable for many languages~\citep{EspanaBonet2011}.

Although generating correct agreements in translations is relatively straightforward in RBMT, the task seems much harder for PB-SMT systems. Indeed, research  on subject-verb agreement of Persian sentences translated from English revealed that, even for Google Translate --at the time the world's most widely used PB-SMT system -- subject-verb agreement remains an issue~\citep{Bozorgian2015}. 

Apart from harming the overall quality of the translation, agreement issues can distract human post-editors from the benefits of using PB-SMT as a tool to increase their productivity; as subject-verb agreement is deemed to be relatively `easy' for both native and non-native speakers, translators rightly expect MT systems to get this right, and when they do not, whatever benefits do accrue from using MT as a productivity enhancer are masked by such obvious, `simple' errors.

The reason why agreement causes difficulties for MT systems is due to the fact that  agreement rules differ between languages and are dependent on the specific morphological structure of the languages involved \citep{Avramidis2008}. It becomes increasingly hard to achieve correct subject-verb agreement for an automatic translation system when translating from a morphologically poor(er) language (e.g. English) into a morphologically richer one (e.g. French). For example, one surface verb form in an English source sentence can correspond to multiple surface verb forms on the French target side. As such, the system will have to pick one of the possible translations by relying on context words, as simply relying on the source word would not be sufficient to pick the correctly inflected form. Depending on the sentence, this can be relatively easy when subject and verb are in close proximity to each other. However, as soon as other words are placed in between the subject and the verb, PB-SMT systems' reliance on \emph{n}-grams limits its ability to deal with long(er) distance dependencies considerably. An example of an ambiguous word and its possible translations in French is given in Figure~\ref{fig:work}.

\tikzstyle{startstop} = [rectangle, rounded corners, minimum width=2cm, minimum height=0.5cm,text centered, draw=black, fill=white!30]
\tikzstyle{transl} = [rectangle, minimum width=2cm, minimum height=0.5cm,text centered, draw=black, fill=red!30]
\tikzstyle{arrow} = [thick,->,>=stealth]

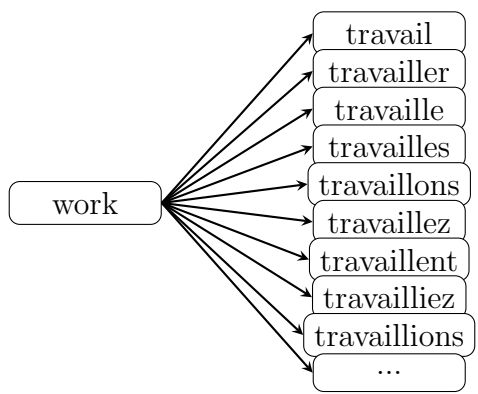
\begin{figure}[h!]
\centering
\begin{tikzpicture}[node distance=2cm]
\node (work) [startstop, yshift=-0.5cm] {work};
\node (travail) [startstop, right of=work, xshift=2cm, yshift=2.25cm] {travail};
\node (travailler) [startstop, right of=work, xshift=2cm, yshift=1.75cm] {travailler};
\node (travaille) [startstop, right of=work, xshift=2cm, yshift=1.25cm] {travaille};
\node (travailles) [startstop, right of=work, xshift=2cm, yshift=0.75cm] {travailles};
\node (travaillons) [startstop, right of=work, xshift=2cm, yshift=0.25cm] {travaillons};
\node (travaillez) [startstop, right of=work, xshift=2cm, yshift=-0.25cm] {travaillez};
\node (travaillent) [startstop, right of=work, xshift=2cm, yshift=-0.75cm] {travaillent};
\node (travaillions) [startstop, right of=work, xshift=2cm, yshift=-1.25cm] {travailliez};
\node (travailliez) [startstop, right of=work, xshift=2cm, yshift=-1.75cm] {travaillions};
\node (etc) [startstop, right of=work, xshift=2cm, yshift=-2.25cm] {...};
\draw [arrow] (work.east) -- (travail.west);
\draw [arrow] (work.east) -- (travailler.west);
\draw [arrow] (work.east) -- (travaille.west);
\draw [arrow] (work.east) -- (travailles.west);
\draw [arrow] (work.east) -- (travaillons.west);
\draw [arrow] (work.east) -- (travaillez.west);
\draw [arrow] (work.east) -- (travaillent.west);
\draw [arrow] (work.east) -- (travailliez.west);
\draw [arrow] (work.east) -- (travaillions.west);
\draw [arrow] (work.east) -- (etc.west);
\end{tikzpicture}
\caption{\textit{One-to-many} relation between English verb `work' and some of its possible translations in French} \label{fig:work}
\end{figure}

The word `work' is first of all ambiguous as, just like in English, it can be either a noun (`travail') or a verb (`travailler',`travaille'...) in French. Additionally, in French, as a verb, it can be present tense (`travaille(s)',`travaillons', `travaillez' and `travaillent'), subjunctive mood (`travaille(s)',`travaillent',`travaillions' and `travailliez') or an infinitive (`travailler'). In the present tense and subjunctive mood, it can be translated into either 1$^{st}$ (`travaille') or 2$^{nd}$ (`travailles') person singular or 1$^{st}$ (`travaillons'), 2$^{nd}$ (`travailliez) or 3$^{rd}$ person plural (`travaillent'). Aside from the many morphological forms the English verb `work' can take in French, from a lexical point of view, `travailler' is not the only possible translation as other synonymous or semi-synonymous French verbs (e.g. `fonctionner', `marcher', `progresser'...) could be used as well (although arguably with different connotations). This results in one English word having a multitude of possible French translations, which illustrates how translating into a morphologically richer language can be a complex disambiguation task, especially when relying on very limited contextual information.

\section{Related Work}\label{sec:relsv}
\subsection{Statistical Machine Translation}
\begin{figure}[h!]
\centering
\begin{tikzpicture}[node distance=2cm]
\node (TRAVAIL) [startstop, yshift=-0.5cm] {TRAVAIL};
\node (work) [startstop, right of=TRAVAIL, xshift=2cm, yshift=0cm] {work};
\node (travail) [startstop, left of=TRAVAIL, xshift=-2cm, yshift=2.25cm] {travail};
\node (travailler) [startstop, left of=TRAVAIL, xshift=-2cm, yshift=1.75cm] {travailler};
\node (travaille) [startstop, left of=TRAVAIL, xshift=-2cm, yshift=1.25cm] {travaille};
\node (travailles) [startstop, left of=TRAVAIL, xshift=-2cm, yshift=0.75cm] {travailles};
\node (travaillons) [startstop, left of=TRAVAIL, xshift=-2cm, yshift=0.25cm] {travaillons};
\node (travaillez) [startstop, left of=TRAVAIL, xshift=-2cm, yshift=-0.25cm] {travaillez};
\node (travaillent) [startstop, left of=TRAVAIL, xshift=-2cm, yshift=-0.75cm] {travaillent};
\node (travaillions) [startstop, left of=TRAVAIL, xshift=-2cm, yshift=-1.25cm] {travailliez};
\node (travailliez) [startstop, left of=TRAVAIL, xshift=-2cm, yshift=-1.75cm] {travaillions};
\node (etc) [startstop, left of=TRAVAIL, xshift=-2cm, yshift=-2.25cm] {...};
\draw [arrow] (travail.east) -- (TRAVAIL.west);
\draw [arrow] (travailler.east) -- (TRAVAIL.west);
\draw [arrow] (travaille.east) -- (TRAVAIL.west);
\draw [arrow] (travailles.east) -- (TRAVAIL.west);
\draw [arrow] (travaillons.east) -- (TRAVAIL);
\draw [arrow] (travaillez.east) -- (TRAVAIL.west);
\draw [arrow] (travaillent.east) -- (TRAVAIL.west);
\draw [arrow] (travailliez.east) -- (TRAVAIL.west);
\draw [arrow] (travaillions.east) -- (TRAVAIL.west);
\draw [arrow] (etc.east) -- (TRAVAIL.west);
\draw [arrow] (TRAVAIL.east) -- (work.west);
\end{tikzpicture}
\caption{\textit{Many-to-one} relation between some of the French translations of the English word `work', mapped to their lemma `TRAVAIL'} \label{fig:work2}
\end{figure}
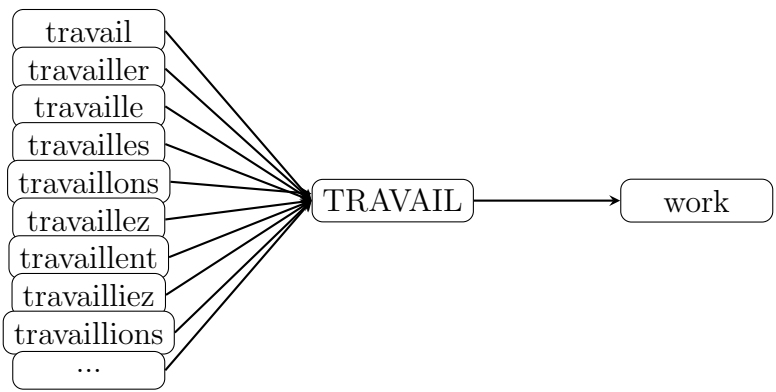

Initially, research on integrating morphological information in SMT aimed to improve  translation quality from a morphologically rich language, such as German or French, into English: a morphologically poor language \citep{Corston2004,Niessen2004,habash2006arabic,Birch2007,Carpuat2009,Wang2012}. The aim was to reduce data sparsity by simplifying the rich(er) source language using morphologically driven preprocessing techniques~\citep{goldwater2005}.
The difficulty when translating from a morphologically rich language into a  morphologically poor one is a \emph{many-to-one} problem that can be made easier for the MT system to solve by converting the actual word form into its lemma or stem in a pre-processing step. Essentially, such techniques make source and target language more alike by simplifying the rich source language into a form that is morphologically more similar to the target language. As shown in Figure~\ref{fig:work2}, by merging `superfluous' morphological variants together, necessary information on one particular target form is grouped together, reducing data sparseness.\footnote{We would, however, like to point out that, in Figure~\ref{fig:work}, that means grouping the noun (`travail') and the verb forms (`travaille',`travailles'...) together. Often research on simplifying the source forms by replacing it with a lemma would integrate POS information in order to overcome such problems.}

\cite{Niessen2004} studied the effects of decomposing German words using lemmas and morphological tags, showing that on relatively small corpus sizes (up to 60,000 parallel sentences) improvements in translation quality are observed. Similar studies that followed obtained positive effects translating from Spanish, Catalan and Serbian~\citep{popovic2004towards}, Czech~\citep{goldwater2005} or Arabic~\citep{habash2006arabic} into English using tokens, lemmas and POS-tags. 

After initial research on simplifying rich target languages, attention slowly shifted to the reverse scenario, i.e. translations into a morphologically richer language. This changes the original problem from a \emph{many-to-one} relationship between source and target  into a \emph{one-to-many} relationship. Several strategies have been proposed to translate from a morphologically poor language into a morphologically richer language. The issue, as stated in \cite{Koehn2005}, is more complex since grammatical features such as number or gender might need to be inferred during the decoding process. Solutions that have been proposed to handle morphology-related difficulties include: (i) preprocessing of the source data, on the assumption that the necessary information to translate an ambiguous word can be found in its source context \citep{Ueffing2003,Avramidis2008,Haque2010}, or (ii) a combination of both pre- and post-processing in a two-step translation pipeline. The two-step translation method usually implies first building a translation model with stems, lemmas or morphemes, and then inflecting them correctly \citep{ElKahlout2006,Virpioja2007,El2012,Fraser2012}. Both pre- and post-processing of source or target language relies on linguistic resources such as POS-taggers, chunkers, parsers, and manually constructed dictionaries, all of which -- assuming them to be available at all -- work to different levels of performance. Instead of having a pre- and/or post-processing step, \cite{Koehn2007_b} proposed to have a tighter integration of linguistic information by introducing factored models. In factored translation models, words are represented as vectors that can contain (apart from the word form) lemmas and POS-tags. However, merely adding lemma and POS information will not provide the translation model with the information necessary to select the correctly conjugated verb form in the target language. Additionally, factored PB-SMT models suffer from data sparsity issues.

\subsection{Neural Machine Translation}

By the end of 2016, NMT became the state-of-the-art for MT. Since then, there have been many studies systematically comparing NMT and PB-SMT's performance, strengths and weaknesses~\citep{Bentivogli2016,Koehn2017six,Pierre2017,Shterionov2018}. One of the main conclusions drawn from such comparisons is that NMT's translations are morphologically more correct. As NMT encodes the entire sentence at once, it is able to handle long distance agreement better than PB-SMT systems. \cite{Pierre2017} observed a jump from 16\% to 72\% in terms of correctly handled morpho-syntactic divergences and commented that improvement was mainly due to NMT's ability to deal with many of the more complex cases of subject-verb agreement. 

There have been studies on integrating more general linguistic features into NMT, which will be discussed in more detail in Chapter~\ref{ch:Supertag}, where we focus particularly on integrating linguistic information into NMT. However, to the best of our knowledge, there is no work in the field of NMT dealing specifically with subject-verb agreement. This might be explained by the fact that NMT handles morphosyntax relatively well, especially when compared to the previous PB-SMT systems. 
NMT's superiority when it comes to dealing with morphosyntax does not, however, imply that NMT systems are infallible when it comes to subject-verb agreement. When looking into more complex cases of subject-verb agreement, one can still encounter examples such as~(\ref{NMTS}) produced by GNMT:

\begin{li}
\label{NMTS}
\item EN: I always assumed that, \textbf{you and I} \emph{would} have loads of fun together, but \emph{could} at some point \emph{start} annoying each other.
\item FR: J'ai toujours supposé que \textbf{vous et moi} \emph{aurions} beaucoup de plaisir ensemble, mais que, à un moment donné, \emph{vous pourriez vous ennuyer}.
\end{li}

In~(\ref{NMTS}), `you and I' (1$^{st}$ person plural) are the subject of the verbs `could' and `start'. The translation produced by GNMT uses a 2$^{nd}$ person plural for both verbs instead. It used to be very easy to trick Google Translate and to find examples of mistakes with respect to subject-verb agreement. Nowadays, their GNMT system is very good at handling subject-verb agreement for the English-French language pair, an observation that we saw further confirmed by the manual evaluation of our test set translated using GNMT in Section~\ref{sec:manual_eval}. It took us a considerable amount of time to find an example where the system outputs an incorrect subject-verb agreement. The main issues remaining are related to ambiguity, where GNMT seems to overgenerate certain forms at the cost of others (e.g. more masculine endings than feminine ones), although theoretically, such translations are not wrong. This issue will be further discussed in Chapter~\ref{ch:Loss}.


\section{Experiments}
In this section, we will discuss in more detail the experiments conducted: (i) by describing how we modeled the source language (Section~\ref{sec:mod_source}), and (ii) by explaining the set-up of the PB-SMT and NMT systems trained (Section~\ref{subsec:expsetup}).

Our approach applies a set of rules to  the `morphologically poor' source-language data in order to render it more `morphologically rich'. Based on source-side information (including POS-tags and the distance between identified subjects and possible main verbs), we modify the identified verb forms in such a way that instead of a \emph{one-to-many} relationship between source and target verb forms,  we create a \emph{one-to-one}  relationship (for most of the verb-forms) between them, by mapping the verb form in the source language to a single correct verb form in the target language. Our method thus makes minimal changes to the source language and so avoids creating extra unnecessary sparsity. Note that while Koehn (\citeyear[p. 313]{Koehn2010}) observes that in general, agreement errors occur ``between multiple words, so simple word features such as POS-tags do not give us sufficient information to detect [them]'', our results demonstrate that, at least as far as subject-verb agreement errors are concerned, POS information can be very useful indeed when combined with some simple pre-processing rules~\citep{vanmassenhove2016improving}.

The verb forms that appear to be specifically difficult to tackle for MT are the 1$^{st}$ and 2$^{nd}$  singular and plural. This is due to the fact that: (i) they are not as common as the 3$^{rd}$  person in written texts, so data-wise are under-represented, and (ii) in English, they share the same verb form with more frequently appearing verb forms, such as the 3$^{rd}$ person plural. However, the context in which the 1$^{st}$  and 2$^{nd}$ person appear is limited since those verb forms can only appear in combination with their specific pronouns (I, you, we). This contrasts with 3$^{rd}$ person singular verb forms which can take NPs, VPs, PPs or even a whole clause as subject, as we demonstrate in (\ref{1S})~\citep{vanmassenhove2016improving}:

\begin{li}
\label{1S}
\item {\em VP as SUBJ}: [$_{VP}$Being a Man Utd fan] makes no sense!
\item {\em PP as SUBJ}: [$_{PP}$In the army] is not a safe place to be.
\item {\em S as SUBJ}: [$_S$That the world is round] is no longer in doubt.
\end{li}

Based on the appearance of a specific pronoun in a sentence, we enriched the closest verb form (within a window size of 4, established empirically) in order to create a \emph{one-to-one} relationship with the source-language verb forms.\footnote{We are aware of the work of \cite{Cai2009} on   subject-verb agreement for English using a dependency grammar approach. While this may be of interest to our work, it is not conducted within the specific remit of MT.} We did this for all pronouns except for the third person verb forms since (i) creating a different verb form for the 3$^{rd}$
person based on the appearance of a specific pronoun would only create additional unnecessary sparsity within our data, and (ii) due to the fact that we changed the verb forms of the 1$^{st}$ and 2$^{nd}$ person and the 3$^{rd}$ person singular already has a different form (s-ending), the 3$^{rd}$ person plural will be the only one left with the base form in the present tense. Our method only requires the use of a POS-tagger on the English side in order to retrieve the conjugated verbs and label them according to the closest subject pronoun. 

We believe that our approach can be used to help generate more correct subject-verb agreements in terms of number and person when translating from a morphologically richer language into a more analytic one. However, for some language pairs, more specific information such as `gender' information might be required in order to select the correct verb form. Issues related to gender in MT will be further addressed in Chapter~\ref{ch:Gender} and Chapter~\ref{ch:Loss}.

The existing methods described all require (a combination of) linguistic resources such as POS-taggers, parsers, morphological analyzers etc. In contrast with the research mentioned above, our work focuses only on subject-verb agreement and not on other problems related to translations into morphologically rich languages (e.g. case or other types of agreement). We will show that improving subject-verb agreement when translating from English to French does not require a two-step translation pipeline where both source and target language are remodeled, since the morphological structure of French is not as complex as Russian~\citep{Toutanova2008,Marevcek2011} or German \citep{Fraser2012}.  Therefore, as in \cite{Avramidis2008} and \cite{Haque2010}, we aim to improve subject-verb agreement by building a system that augments the source-language data with extra information drawn from the source-side context. However, unlike their work, we do this by using only a POS-tagger on the English side \citep{vanmassenhove2016improving}.

We would also like to discuss some of the shortcomings of our approach. As this was a first attempt aiming to experiment with a simple solution towards better subject-verb agreement in PB-SMT, we employed a very basic and low-resource approach. As such, we limited our work to simple subjects, i.e. (i) we did not take conjoined NPs into account (e.g. `you and I' or `he/she and I'), and (ii) the system cannot properly handle distances longer than four words in between a subject pronoun and its verb. Overcoming these shortcomings is impossible without relying on more advanced tools such as syntactic parsers. To give a realistic idea of the improvement that can be obtained with a simple approach, our test set was selected randomly (we did not filter out appearances of phenomena of any kind). 

\subsection{Modeling of the Source Language}\label{sec:mod_source}

We use the Penn TreeBank tagset \citep{Marcus1994} and the default tagger used in the nltk package\footnote{\url{http://www.nltk.org/}} to tag the source  sentences. Once the source sentences contain the information from the POS-tagger, we can in the next step use this information to look for verb forms that agree in person with a subject. These are the non-3$^{rd}$ person singular present (`VBP'), the 3$^{rd}$ person singular present (`VBZ'), the past verb tense (`VBD') and modal verbs (`MD')~\citep{vanmassenhove2016improving}. 

Within the already tagged sentences, we search for 1$^{st}$ and 2$^{nd}$ person
pronouns (`I', `you' and `we'). Once a pronoun is found, we identify the closest verb form (a verb tagged `VBP', `VBZ', `VBD' or `MD') following the pronoun, within a window of size 4. The verbs found are enriched with information of the pronoun as in Table~\ref{workXR}.

\begin{table}[ht]
\begin{center}
\begin{tabular}{ |c|c| } 
 \hline
 I work & I work1sg \\ 
 \hline
 You work & you work2  \\ 
 \hline
 we work & we work1pl \\ 
 \hline
\end{tabular}
\end{center}
\caption{Enriching English surface verb forms with POS information}\label{workXR}
\end{table}	

We distinguish between declarative and interrogative sentences by looking at the last token of the sentence. If it is a question mark, we identify this sentence as an interrogative sentence. For interrogative sentences, the verb typically (but not always) precedes the pronoun, so  in these types of sentences, we first look for a verb appearing before the pronoun (within a window size of 2, established empirically) before looking at the words following it.

Table~\ref{work-final} shows an example of the verb forms obtained after pre-processing.

\begin{table}[ht]
\begin{center}
\begin{tabular}{ |c|c| } 
\hline
I work & I {\bf work1sg}  \\ 
\hline
you work & you {\bf work2}  \\ 
\hline
he {\bf works} & 	  \\ 
\hline
we work & we {\bf work1pl} \\ 
\hline
they {\bf work} &  \\ 
\hline
\end{tabular}
\end{center}
\caption{Final verb forms after pre-processing}\label{work-final}
\end{table}

Although our method does not resolve the ambiguity between the third person singular and third person plural in the past tense, it does reduce the complexity of the disambiguation process by converting a \emph{one-to-many} problem into a \emph{one-to-two} problem. Furthermore, since 3$^{rd}$ person plural and singular are both very common verb forms, their disambiguation is less problematic for the language model to resolve.\footnote{For the 2$^{nd}$ person the ambiguity remains given that the pronoun {\em you} is identical in both singular and plural. However, in French, the 2$^{nd}$ person plural is also ambiguous since it can be both plural or singular (polite form).}

\subsection{Experimental Setup}\label{subsec:expsetup}
In this section, we describe in more detail the specifications of the systems trained and the datasets used for training and testing.

\subsubsection{PB-SMT}
To evaluate our approach, we build two types of PB-SMT systems with the Moses toolkit \citep{Koehn2007}: (i) from the original data, that we refer to as  \emph{baseline}, and (ii) from our morphologically-enriched data, which we refer to in the rest of this chapter as \emph{morphologically-enriched} systems. We then score these PB-SMT
systems using automatic evaluation metrics as well as manual error analysis and compare them. 

For training we use subsets of increasing sizes (respectively 200K, 400K and 600K sentences) of the Europarl parallel corpus \citep{Koehn2005} for the  English--French language pair. Both the baseline data as well as the morphologically-enriched data are tokenized and lowercased using the Moses tokenizer. Sentences longer than 60 tokens are filtered and not used in our model. We use the default Moses settings to train our systems. 

Since we are specifically interested in subject-verb agreement, we want to have as much variety in verb forms as possible for the development set and the test set. Accordingly, we created our development and test sets from the WMT development sets from 2008 until 2013.\footnote{\url{http://www.statmt.org/wmt13/}} We select from these data the sentences that contain 1$^{st}$ person and 2$^{nd}$ person pronouns and 3$^{rd}$ person verb forms. We split the 2098 sentences retrieved into a development set of 1000 sentences and a test set of 1098 sentences. To manually evaluate the performance of the \emph{morphologically-enriched} PB-SMT system against the \emph{baseline} PB-SMT systems, we randomly extract 60 out of the 1098 input  sentences containing at least 10 occurrences of each verb form. Table~\ref{tbl:pronoun_results} gives an overview of the number of pronouns appearing in the development set, test set and manual test set.

\begin{table}[ht]
\begin{center}
 \begin{tabular}{|l|c|c|c|c|c|c|}
    \hline
    {\# of Pronouns in each set}	& DEV SET 		& TEST SET	&MANUAL TEST SET 						\\\hline
    I	 				& 458			& 542		& 28								\\\hline
    YOU  				& 317			& 370		& 32								\\\hline
    HE, SHE, IT				& 382			& 339		& 28								\\\hline
    WE					& 519			& 417		& 18								\\\hline
    THEY				& 82			& 54		& 11								\\
    \hline
 \end{tabular}
\end{center}
 \caption{Number of different pronouns in the development set, test set and manual test set.}\label{tbl:pronoun_results}
\end{table}

For our \emph{morphologically-enriched} system, we use exactly the same training, development and test sets but pre-processed as described in Section~\ref{sec:mod_source}.

\subsubsection{NMT}
For our NMT system, we trained an RNN NMT model using the OpenNMT-py toolkit. The system was trained for 100K steps, saving an intermediate model every 5000 steps. We used byte pair encoding (BPE) with 50,000 operations to deal with out-of-vocabulary words (OOV). We scored the perplexity of each model on the development set and chose the one with the lowest perplexity as our best model, which we used later for translation. The options we used for the neural systems are as follows: 
\begin{itemize}
    \item RNN: size: 512, RNN type: bidirectional LSTM, number of layers of the encoder and of the decoder: 4, attention type: mlp, dropout: 0.2, batch size: 128, learning optimizer: adam and learning rate: 0.0001.
\end{itemize}
The neural system has the learning rate decay enabled and the training is distributed over 4 nVidia 1080Ti GPUs. The selected settings for the RNN systems are optimal according to \cite{Britz2017}.

For training, we used the largest dataset used for the PB-SMT system (600K) sentences, which for NMT, is still a relatively small dataset. 
Additionally, we used GNMT's system to translate the test set for manual evaluation. This way, we can get an idea of how a large state-of-the-art (SOTA) NMT system handles subject-verb agreement.

The test sets used to evaluate the system automatically and the manual test sets are identical to the ones used to evaluate the \emph{baseline} and \emph{morphologically-enriched} PB-SMT systems described in the previous section.

\section{Results}
In this section, the results of our experiments are discussed. We conducted both an automatic evaluation (Section \ref{sec:sec:ae}) and a manual analysis (Section \ref{sec:manual_eval}). We conducted the same experiments with a PB-SMT system and a baseline NMT system trained on the largest dataset used in our PB-SMT experiments (600K sentences). Additionally, we included a manual analysis of the NMT system and of GNMT.\footnote{Our manual analysis was conducted on 26th April, 2019. This is important as the models are frequently updated.}

\subsection{Automatic Evaluation}\label{sec:sec:ae}
\subsubsection{PB-SMT}
We evaluate the \emph{morphologically-enriched} PB-SMT systems against the baseline systems for the three datasets with different sizes. To score each of the PB-SMT systems we built, we use the automatic evaluation metrics BLEU and TER. The results of the automatic evaluation are presented in
Table~\ref{tbl:eval_results}. For comparison, we added the results of the manual evaluation to Table~\ref{tbl:eval_results}. The manual error evaluation will be further described in Section~\ref{sec:manual_eval}.

\begin{table}[ht]
\begin{center}
 {\setlength\tabcolsep{3.5pt} \begin{tabular}{|l|c|c|c|c|c|c|}
    \hline
    \multirow{2}{*}{\# sentences} 	& \multicolumn{2}{c|}{BLEU} 	& \multicolumn{2}{c|}{TER} 	& \multicolumn{2}{c|}{Manual (in \%)} 	\\\cline{2-7}
					& Baseline	& Person-verb	& Baseline	& Person-verb	& Baseline	& Person-verb		\\\hline
    200 000 				& 19.7		& {\bf 19.8}	& 62.9		& {\bf 62.7}	& 77.5		& {\bf 87.6}		\\\hline
    400 000 				& {\bf 20.6}	& 20.4		& 62.2		& {\bf 62.0}	& 78.4		& {\bf 84.5}		\\\hline
    600 000 				& {\bf 21.6}	& 21.5		& 61.4		& {\bf 61.1}	& 77.9		& {\bf 88.2}		\\
    \hline
 \end{tabular}}
\end{center}
 \caption{Evaluation metrics comparing the baseline and the \emph{pronoun-verb} approach.}\label{tbl:eval_results}
\end{table}

BLEU scores, a measure that compares the overlap between a translation and its reference, range from 0 to 100~\citep{Papineni2002}. The higher the BLEU score, the better the translation. While a small improvement is seen for the first data set (+0.1), there is small decrease (-0.2) for the two larger data sets. TER, on the contrary, is an error metric that measures the amount of editing required to change a system output so it matches (one of the) reference(s), i.e. the lower the TER, the better the translation \citep{Snover2006}. As far as TER is concerned, a small improvement is seen for all data sets. 

However, there is an intrinsic problem in using document-level (or even sentence-level) metrics to try to demonstrate improvements in translation quality when one is focused on a single linguistic phenomenon. As in other works on modeling morphology in MT \citep{Avramidis2008, Marevcek2011}, when computing (say) the BLEU score, all {\em n}-grams are weighted equally. However, this does not take into consideration that not every part of a document (or sentence) contributes equally to the overall adequacy and fluency of the translation, which may lead to an incorrect understanding of the system's actual quality. More precisely for our purposes, a subject-verb agreement error that may considerably influence both the grammaticality (fluency) and the semantics (adequacy) of a translation is treated in equal measure to any other error \citep{Callison2006}. Accordingly, it is noteworthy that correcting subject-verb agreement errors leads to translations that are considered better by humans \citep{Marevcek2011}.

Table~\ref{tbl:eval_results} demonstrates a similar trend, where we see that for each data set, our system considerably outperforms the equivalent
baseline when looking at the results of the manual evaluation. For the largest data set, our model improves by 10.3\% absolute (or 13.2\% relative) compared to the equivalent \emph{baseline}. However, based on the automatic evaluation, the improvement is not so clear.

Note too that as is well-known, PB-SMT systems can generate perfectly good translations which do not result in an improved BLEU score, simply because the output translation differs significantly from the reference. One such example appears in (\ref{2}):

\begin{li}\label{2}
 \item {\em Reference}: ``Nous analysons cela car ces id\'{e}es \ldots'' 
 \item {\em Baseline}: ``Nous examin\'{e} pourquoi ces id\'{e}es \ldots'' 
 \item {\em Our system}: ``Nous examinons pourquoi ces id\'{e}es \ldots'' 
\end{li}

Here, the \emph{baseline} system inserts a past participle in the position where the main verb should be. Our \emph{morphologically-enriched} system correctly inserts a 1$^{st}$-person plural verb {\em examinons} (`examine'), which while being semantically correct, differs from the reference {\em analysons} (`analyse'). Another example is (\ref{3}):

\begin{li}
\label{3}
 \item {\em Reference}: ``Je sais qui tu es'', a-t-il dit, gentiment.
 \item {\em Baseline}: ``Je sais qui vous sont'', il a dit aimablement.
 \item {\em Our system}: ``Je sais qui vous \^{e}tes,'', il a dit aimablement.
\end{li}

In this example, the \emph{baseline} system inserts a 3$^{rd}$-person plural form {\em sont} (`are') after the 2$^{nd}$-person plural pronoun {\em vous} (`you'). While our system produces the correct form {\em \^{e}tes}, as the reference contains the 2$^{nd}$-person singular phrase {\em tu es}, there is a significant difference between this and the
output translation, so no additional benefit in terms of BLEU score accrues; indeed, in this example, the incorrect \emph{baseline} translation obtains exactly the same BLEU score as our (arguably) correct morphologically enhanced system. 

\subsubsection{NMT}
We evaluate the NMT system and compare it to the \emph{morphologically-enriched} and \emph{baseline} PB-SMT systems. We use BLEU and TER for automatic evaluation. We only trained an NMT system on 600k sentences, as 200k and 400k are too little to yield results indicative for NMT's potential as its learning curve is steeper~\citep{Koehn2017six}. Nonetheless, it is worth saying that even 600k sentences is a small parallel dataset to train an NMT system on. 

\begin{table}[ht]
\begin{center}
 {\setlength\tabcolsep{5pt}\begin{tabular}{|l|c|c|c|c|c|c|}
    \hline
    \multirow{2}{*}{\# sentences} 	& \multicolumn{3}{c|}{BLEU} 	& \multicolumn{3}{c|}{TER} 	\\\cline{2-7}
					& PB-SMT   & ME-PB-SMT    & NMT           & PB-SMT   &ME-PB-SMT & NMT	       		\\\hline
    600 000 		&  21.6	& 21.5      &\textbf{23.9*}	& 61.4  & 61.1  &\textbf{60.9*}  	\\\hline
 \end{tabular}}
\end{center}
 \caption{Evaluation metrics comparing the baseline PB-SMT with our morphologically-enriched PB-SMT system (ME-PB-SMT) and the NMT system. The \emph{*} indicates results that are significant $(p<0.5)$.}\label{tbl:eval_resultsNMT}
\end{table}

As shown in Table~\ref{tbl:eval_resultsNMT}, in terms of automatic evaluation metrics, the NMT systems significantly\footnote{We used multeval version 0.5.1~\citep{Clark2011} to compute statistical significance.} outperform both PB-SMT systems. However, as discussed and demonstrated previously, the automatic evaluation does not always reflect how well a system performs on a specific linguistic task such as subject-verb agreement. The manual analysis conducted and described in the next section will shed further light on whether NMT indeed outperforms PB-SMT.

\subsection{Manual Error Evaluation}\label{sec:manual_eval} 
\subsubsection{PB-SMT}
In order to discover in what circumstances our \emph{morphologically-enriched} system
improves over the \emph{baseline}, in this section we describe a detailed manual error analysis we conducted which focuses on pronoun-verb agreement cases.\footnote{All the verbs were retrieved and given a mark if they agreed with their subject. There was only one annotator since the correctness of agreement between subject and verb should not be  ambiguous.} We evaluate the outputs of the test sets for the \emph{baseline} and \emph{morphologically-enriched} PB-SMT systems.

For each test set $s$ we compute the correctness of the pronoun-verb translation ratio by dividing the correctly translated pronoun-verb pairs ($t^{correct}$) by all pronoun-verb pairs that should have agreement ($t^{total}$): $C_s=\frac{t_s^{correct}}{t_s^{total}}$.
The higher the correctness of pronoun-verb translation ratio, the better. 

We compared both the \emph{baseline} and the \emph{morphologically-enriched} systems and saw that our approach leads to PB-SMT systems that produce a more correct translation with respect to subject-verb agreement. In the right-most column of Table~\ref{tbl:eval_results} we already observed that all three \emph{morphologically-enriched} systems have a higher correctness-score for pronoun-verb translation. To obtain a better view on how the \emph{morphologically-enriched} systems (referred to in Table~\ref{tbl:detailed_eval_results} as \emph{ME1}, \emph{ME2} and \emph{ME3}) perform compared to the \emph{baselines} (\emph{BS1}, \emph{BS2} and  \emph{BS3}), for every pronoun we count the pronoun-verb pairs that are translated correctly. 

\begin{table}
\begin{center}
 {\setlength\tabcolsep{5pt}\begin{tabular}{|l|c|c|c|c|c|c|}
    \hline
{\% correct SV agreement}	&I		&YOU		&HE, SHE, IT	&WE		&THEY		&TOTAL		\\\hline
BS1	(200k)			 	&78.57		&60.71		&93.33		&61.11		&69.23		&72.59		\\\hline
ME1	(200k)				&92.86		&82.76		&93.10		&88.24		&81.82		&87.75		\\\hline
BS2	(400k)				&85.71		&57.14		&93.55		&70.59		&76.92		&76.78		\\\hline	
ME2	(400k)				&85.71		&79.31		&90.00		&83.33		&81.82		&84.04		\\\hline
BS3	(600k)				&78.57		&57.14		&93.33		&70.59		&76.92		&75.31		\\\hline	
ME3	(600k)				&92.86		&82.76		&93.10		&83.33		&81.82		&86.77 		\\\hline\hline
BS (average over all)			&80.95		&58.33		&93.41		&67.43		&74.36		&74.90		\\\hline	
ME (average over all)			&90.48		&81.61		&92.07		&84.97		&81.82		&86.19		\\
    \hline
 \end{tabular}}
\end{center}
 \caption{\% correctly translated pronoun-verb pairs in \emph {baseline} and \emph {pronoun-verb} approach per pronoun.}\label{tbl:detailed_eval_results}
 \end{table}
From the results in Table~\ref{tbl:detailed_eval_results}, we see that our approach outperforms the \emph{baseline} systems in terms of agreement for all pronouns except for the 3$^{rd}$ person singular, where the \emph{baseline} and \emph{morphologically-enriched} systems score similarly (respectively 93.41\% and 92.07\%). The other subject-verb agreements are more difficult for all \emph{baseline} systems. The biggest improvements in the \emph{morphologically-enriched} system can be noted for the 2$^{nd}$ person and the 1$^{st}$ person plural, where over all datasets, we can see an absolute increase of 23.28\% and 17.54\%, respectively. When averaging over the three datasets for all pronouns, we observe an overall improvement of 11.29\% absolute (15.07\% relative).

In (\ref{4}) we show an example of the translations generated by the two different systems. The \emph{baseline} system translates the verb that agrees with the 1$^{st}$-person plural pronoun \emph{nous} (`we') incorrectly as an infinitive \emph{aider} (`to help'). However, our system translates it to the correct verb form \emph{aidons}.
 \begin{li}
\label{4}
 \item {\em Source}: ``{\bf We help} relatives as much as patients'' says Nathalie Savard, Director of Care.
 \item {\em Baseline}: ``{\bf Nous aider} proches autant que les patients'' affirme Nathalie Savard, directeur de soins. 
 \item {\em Our system}: ``{\bf Nous aidons} proches autant que les patients'' affirme Nathalie Savard, directeur de soins. 
\end{li}

Another example is (\ref{5}):

 \begin{li}
\label{5}
 \item {\em Source}: 'Then {\bf you can} start breaking them,' Jakub told us.
 \item {\em Baseline}: {\bf Vous} ' then {\bf puissent} commencer à les enfreindre,``Jakub nous a dit.
 \item {\em Our system}: {\bf Vous pouvez} ' then commencer enfreindre,'' Jakub nous a dit.
\end{li}
In example~(\ref{5}), our system correctly translates the pronoun-verb pair \emph{you can} as \emph{vous pouvez}. However, the baseline translates the verb incorrectly as a subjunctive 3$^{rd}$ person plural verb \emph{puissent}.

In the next example (\ref{6}), the verb \emph{understood} that agrees with the 1$^{st}$ person singular pronoun \emph{I} is translated by the \emph{baseline} system as a past participle while the \emph{morphologically-enriched} system translates it correctly. This example also illustrates how our system deals with unseen verb-forms. While both systems do not translate the verb \emph{drill} since it is unseen in the training set, our system adds the information of the subject to the verb form \emph{drill1pl}. In our error analysis, we counted these missing verb forms as errors for both systems. Nevertheless, the information that was added to the verb form by our system is correct.

 \begin{li}
\label{6}
 \item {\em Source}: Here {\bf I} finally {\bf understood} why {\bf we drill} all the\ldots
 \item {\em Baseline}: Ici, {\bf je} enfin {\bf compris} pourquoi {\bf nous drill} tous les\ldots
 \item {\em Our system}: Finalement, {\bf j'ai compris} pourquoi {\bf nous ici drill1pl} tous les\ldots
\end{li}

\subsubsection{NMT}

We used the same manual test set used for evaluating our \emph{baseline} and \emph{morphologically-enriched} PB-SMT system to test the NMT system and GNMT. GNMT's system was included as it represents an NMT system trained on a very large dataset. This evaluation can show us what NMT's full potential is when it comes to handling subject-verb agreement.

\begin{table}[ht]
\begin{center}
 {\setlength\tabcolsep{4pt}\begin{tabular}{|l|c|c|c|c|c|c|}
    \hline
{\% correct SV agreement}	&I		&YOU		&HE, SHE, IT	&WE		&THEY		&TOTAL		\\\hline
BS	(600k)				&78.57		&57.14		&93.33		&70.59		&76.92		&75.31		\\\hline	
ME	(600k)				&92.86		&82.76		&93.10		&83.33		&81.82		&86.77 		\\\hline
NMT	(600k)				&95.24		&93.55		&96.72		&92.31		&100		&95.56 		\\\hline
GNMT	            	&100.00		&100.00		&100.00		&100.00		&100.00		&100.00 		\\\hline
 \end{tabular}}
\end{center}
 \caption{\% pronoun-verb pairs with correct agreement in \emph {baseline} (BS), \emph {morphologically-enriched} (ME) and NMT system.}\label{tbl:detailed_eval_resultsNMT}
 \end{table}

The manual evaluation confirms previous research comparing NMT and PB-SMT. Even the NMT system trained on 600k sentences clearly outperforms both PB-SMT systems for all cases. Some observations we made during the manual NMT evaluation:
\begin{itemize}
\item The English pronoun `you' was \textbf{always} translated into the plural (or polite) form `vous' in French by the NMT system. This was not the case for the PB-SMT systems, where `tu' would still appear as a translation for `you', although less frequently than `vous'. The increased appearance of already frequent words and the loss of less frequent ones, will be further discussed in our chapter on the loss of linguistic diversity in NMT (Chapter~\ref{ch:Loss}).  
\item Sometimes, the NMT translation would be fluent although completely inadequate compared to the source sentence. In this evaluation, we only considered the correctness of the agreement in the translation itself (without taking the adequacy of the translation into account), which might inflate the correctness scores of the systems. However, as we evaluated all systems in a similar manner, the systems' scores can be compared against each other. An example of an inadequate yet syntactically correct sentence on its own is given in~(\ref{ex:errorAd}). The reference translation mentions ``a dit Jakub'' (``Jakub said'') while the NMT translation uses ``On nous dit'' (``they told us'').

 \begin{li}
\label{ex:errorAd}
\item {\em Source}: ``\ldots start breaking them,'' \textbf{Jakub} \textbf{told} us.
 \item {\em Ref}: ``\ldots commencer à y contrevenir'', \textbf{a} dit \textbf{Jakub}.
 \item {\em NMT}: ``\ldots commencer à les briser,'' \textbf{On} nous \textbf{dit} ".
\end{li}

For GNMT, we did not encounter any subject-verb issues, although we did observe an inconsistency in~(\ref{ex:errorInc}). While the English Source uses `they' and `them' in a generic way, both the reference human translation and the translation produced by GNMT use an equivalent construction in French using the word `il'.\footnote{`Il' is literally translated in English into `he' but can be used generically like `they' in English.} However, the GNMT translation is not consistent as it first uses `il' but later in the sentence switches to `elle' to refer to the same entity. `Elle' is the female variant of `il' and cannot be used generically. This example is also related to gender bias and gender agreement which is further discussed in Chapter~\ref{ch:Gender}.

\begin{li}
\label{ex:errorInc}
 \item {\em Source}: ``\ldots we should expect them to give us different (smaller) size than what \textbf{they} really \textbf{wear}.''
 \item {\em Ref}: ``\ldots nous devons ensuite compter sur le fait qu'\textit{il} nous \textit{dicte} avec une grande probabilité une taille complètement différente (plus petite) que celle qu'\textbf{il} \textbf{a} en réalité.''
  \item {\em GNMT}: ``\ldots nous devrions nous attendre à ce qu'\textit{il} nous \textit{donne} une taille différente (plus petite) que celle qu'\textbf{elle} \textbf{porte} réellement.''
\end{li}

\end{itemize}

\section{Conclusions}

PB-SMT systems typically have problems in ensuring correct subject-verb agreement when producing translations. This is especially problematic when translating from a  morphologically impoverished language like English into a morphologically rich language; given that a huge proportion of the world's translation requirement is from English into some other language, this problem affects/affected many systems.

Using a simple POS-based model, we annotate source-language verbs with morphological information to turn the problem from a \emph{one-to-many} mapping between English surface forms and their multiple target-language equivalents into a series of \emph{one-to-one} associations. 
Testing this on English-to-French, we see improvements (averaged over the three different data sets) in subject-verb agreement of 11.29\% absolute (or 15.07\% relative) compared to the equivalent Moses \emph{baseline} for our \emph{morphologically-enriched} system, as measured by a human evaluation. We note the problem of relying on automatic metrics when honing in on specific translational phenomena, as well as the well-known problem of improvements in translation quality not always being reflected by increased automatic evaluation scores~\citep{Barreiro2014,vanmassenhove2016improving}.

With the arrival of NMT, the architecture of the SOTA MT systems changed completely and stopped relying on \emph{n}-grams as NMT encodes an entire sentence at once. Research comparing PB-SMT and NMT concluded that NMT outperforms PB-SMT in term of subject-verb agreement and morphosyntax in general~\citep{Bentivogli2016,Koehn2017six}. To account for this, we carried out additional experiments to verify how a relatively small NMT system and GNMT perform when it comes to subject-verb agreement. Our experiments and subsequent manual and automatic evaluation confirm previous comparisons between PB-SMT and NMT, showing that NMT systems, especially a large NMT system such as GNMT, perform remarkably well in terms of subject-verb agreement.

At this stage, we can formulate a partial answer to RQ2. For PB-SMT, we identified that the \emph{n}-grams are an insufficient information source in order to generate correct subject-verb agreements. Additional linguistic knowledge can be integrated in a simple way in order to disambiguate source verb forms, facilitating the generation of the appropriate target forms. NMTs performance indicated that simple sentence-level agreement issues are grasped by its sentence encoders. Nevertheless, in the following chapters, we identify NMT still encounters difficulties with more complex linguistic phenomena. We further address RQ2 in Chapter~\ref{ch:Supertag} and Chapter~\ref{ch:Gender} where we experiment with ways of integrating additional linguistic knowledge into the NMT pipeline.

\chapterbib

%% file: Aspect_SMT_NMT.tex

\newpage
\epigraph{Existence really is an imperfect tense that never becomes a present.}{\textit{Friedrich Nietzsche}}

\chapter{Aspect and Tense in Statistical and Neural Machine Translation}\label{ch:Aspect}

In the previous Chapter, we presented a morphologically-enriched PB-SMT system in order to improve how PB-SMT systems handle basic cases of number agreement between subject and verb. Our manual analysis showed the effectiveness of our approach for PB-SMT. However, when later on comparing the output with translations generated by NMT systems, it became clear that even a baseline NMT architecture outperforms a morphologically-enriched PB-SMT system for this particular morphosyntactic translation difficulty. Furthermore, it appeared that a system such as GNMT did not make a single mistake on the manually evaluated test set in terms of subject-verb agreement.\footnote{The test set was, however, limited to 60 sentences and there were still some inconsistencies, although unrelated to number agreement.} By the time we reached this stage of our research, multiple studies had already compared NMT and PB-SMT systems and it became clear that NMT had some clear advantages over PB-SMT systems, especially when trained on large datasets. As subject-verb agreement is still a relatively easy task to perform, governed by a limited number of syntactic rules, the next step is to investigate how NMT deals with more complex linguistic phenomena. As we focus particularly on verbs and as `tense' and `aspect' have proven to be a difficult task for PB-SMT systems, we decided to verify how well NMT translates verb tenses with the appropriate aspectual value in the target language. Unlike subject-verb agreement, where the verb should be in agreement with its subject, aspect is a grammatical category which value can depend on multiple elements within a sentence or even across sentences: an aspectual network. Furthermore, different languages have different means to express `aspect' which makes it an interesting topic to research from a translation point of view. 

The current chapter addresses RQ1 as we have a deeper look into the knowledge sources of both PB-SMT and NMT systems and study how linguistic theory on aspect is reflected in PB-SMT phrase-tables and NMT encoding vectors. We do so by comparing the information encoded in the knowledge sources and by verifying the performance of a baseline NMT model to a baseline PB-SMT system.

\section{Introduction}

One of the important differences between English and French grammar is related to how their verbal systems handle aspectual information. While the English simple past tense is aspectually neutral, the French and Spanish past tenses are linked with a particular imperfective/perfective aspect. This work examines what PB-SMT and NMT learn about `aspect' and how this is reflected in the translations they produce. We use their main knowledge sources -- phrase-tables in PB-SMT and encoding vectors in NMT -- to examine what kind of aspectual information they encode. Furthermore, we examine whether this encoded `knowledge' is actually transferred during decoding and thus reflected in the actual translations. 

The experiments described in this chapter are based on the translations of the English \emph{simple past} and \emph{present perfect} tenses into French and Spanish imperfective and perfective past tenses. We examine the interaction between the lexical aspect of English simple past verbs and the grammatical aspect expressed by the tense in the French/Spanish translations. It results that PB-SMT phrase-tables contain information about the basic lexical aspect of verbs. Although lexical aspect is often closely related to the grammatical aspect expressed by the French and Spanish tenses, for some verbs (mainly atelic dynamic verbs) more contextual information is required in order to select an appropriate tense. On the one hand, the PB-SMT \emph{n}-grams provide insufficient context to grasp other aspectual factors included in the sentence to consistently select the tense with the appropriate aspectual value. On the other hand, the encoding vectors produced by a baseline NMT system do contain information about the entire sentence. An analysis based on the English NMT encoding vectors shows that a logistic regression model can obtain an accuracy of 90\% when trying to predict the correct tense based on the encoding vectors. However, these positive results are not entirely reflected in the actual translations, i.e. part of the aspectual information is still lost during decoding.

Translating sentences from one language to another is a complex task that requires a profound knowledge of the two languages involved. Translators use their understanding of the morphology, structure and the semantics of both languages in order to select an appropriate translation in a specific context~\citep{Sager1994,Hogeweg2009}. Additional translation difficulties arise when dealing with ``translation mismatches'', a term used in the field of MT to refer to cases where the grammar of one language makes distinctions that are not made by the other. Such a mismatch can apply to a particular utterance or it can be due to more systematic differences between the source and target language systems. The issues related to subject-verb agreement discussed in Chapter~\ref{ch:Agreement} are one example of such differences.

Systematic differences between two languages reveal something about the linguistic systems involved. A better understanding of the systematicity of apparent mismatches between languages and the mechanisms behind them could lead to a more accurate mapping between two language systems. Although ``two languages are more informative than one'' \citep{Dagan1991}, not many corpus-based translation studies focus on specific linguistic phenomena~\citep{Santos2004}. In the field of MT, and more particularly RBMT, comprehensive and detailed modules were integrated in MT systems (such as Eurotra and Rosetta) that dealt with mismatches related to tense and aspect~\citep{VanEynde1988,Appelo1993}. However, within the field of data-driven MT (SMT and NMT), not many studies focus on resolving particular translation problems related to specific cross-linguistic phenomena.

The mismatches that occur within the verbal systems of languages are particularly interesting since verbs are arguably the most important lexical category of language~\citep{Miller1991}. Sentences are governed by verbs and, with the exception of some languages such as Russian (where it is possible to have a sentence that does not contain a verb), languages need verbs to represent the sentence predicate~\citep{Vcech2011}. Furthermore, verbs have a crucial impact on the general structure of sentences and are the most complex and varied forms in language~\citep{Fischer1978}. Thus, incorrect translations of verbs can propagate errors across a sentence.

English and French/Spanish grammar have considerable areas of overlap, but there are some important mismatches that can cause interference. French and Spanish have, for example, a richer inflectional morphology where verbs need to agree in person and number with their subject (or objects in some cases). Although this is a relatively easy task for human translators, these translation mismatches appear to be difficult to learn for PB-SMT systems~\citep{vanmassenhove2016improving}. There is also no one-to-one correspondence between the English and French/Spanish tense systems. Some tenses are formally similar, such as the English \emph{present perfect} and the French \emph{pass\'{e} compos\'{e}}, but are not in usage.\footnote{The difference can already be noted by looking at their labels: the `present' perfect vs the `pass\'{e}' compos\'{e} (`past' composed) tense.} In the case of the \emph{present perfect} and the \emph{pass\'{e} compos\'{e}}, this is due to a shift that occurred in the French past tense system, where the \emph{pass\'{e} compos\'{e}} has largely replaced the usage of the French \emph{pass\'{e} simple} and is thus used as a perfective past tense. Although a similar evolution has been observed in English, where the \emph{present perfect} is used increasingly in \emph{simple past} contexts (presumably by analogy with the French tense)\footnote{Cases of constructional contamination are studied in more detail in~\cite{Pijpops2016} and~\cite{Pijpops2018}.}, the \emph{present perfect} forms cannot express past events to the same extent as the \emph{pass\'{e} compos\'{e}}~\citep{Engel1998}.

One of the main differences between the English and French or Spanish tenses is related to `aspect' and `tense'. Comrie (1976, p.3) defines tense and aspect as follows:

\indent\blockquote{[...] tense relates the time of the situation referred to some other time, whereas aspects are different ways of viewing the internal temporal constituency of a situation.}
~\newline
\\
The English \emph{simple past} is aspectually vague when compared to the French (\emph{pass\'{e} compos\'{e}} and \emph{imparfait}) and Spanish (\emph{pret\'{e}rito indefinido} and \emph{pret\'{e}rito imperfecto}) past tenses. Although both English and French/Spanish tenses express mainly the tense (present, past or future), the \emph{pass\'{e} compos\'{e}}/\emph{pret\'{e}rito indefinido} and the \emph{imparfait}/\emph{pret\'{e}rito imperfecto}, express the same tense (past) but have a different aspectual meaning. Their distinction is purely based on an aspectual difference and is thus an example of a formally marked aspect that forms part of the morphological system of French/Spanish, a ``grammatical aspect'' \citep{Garey1957}. An example of the grammatical aspect expressed by the Spanish past tenses is given in Example~(\ref{gramAspect}).

\begin{li}\label{gramAspect}
\item 
\begin{tabular}{rlll}
     1. & ``He made dinner.'' \\
     2. & ``\emph{Hizo} la comida.'' & [PER.]\\
     3. & ``\emph{Hac\'{i}a} la comida.'' & [IMP].\\
~\newline\\
\end{tabular}
\end{li}

The perfective aspect expressed in Spanish by the \emph{pret\'{e}rito indefinido} is bounded. It presents the event from the outside, as a whole~\citep{Vendler1957,Comrie1976,Dowty1986}. The perfective reading of the English sentence ``He made dinner.'' is one where the event started and finished (it has been `completed') and thus resulted in a `dinner', an interpretation presented by the Spanish perfective translation ``\emph{Hizo} la comida.''. However, the second translation ``\emph{Hac\'{i}a} la comida.'' presents the event from the inside \citep{Comrie1976,Dowty1986,Vendler1957}, without emphasis on the beginning or the end, merely focusing on the fact that `He was making dinner'. This is the imperfective interpretation and translation of that same English sentence.

This does not imply that the English language does not or cannot convey this aspectual difference since other words in a sentence (the semantics of the verbs, nouns, adjectives, adverbs, etc.) can carry aspectual meaning. While some languages have an overt formal category of grammatical aspect, for others it is a covert semantic category on the sentential (or propositional) level \citep{Filip2012}. The difficulty consists thus of making something that is implicit in the one language explicit when translating into another language. 

As different words in a sentence can carry aspectual meaning, the aspect of a sentence should be regarded as a network. Within this network, the basic `lexical aspect' of the verb is a good starting point to determine the aspectual value of a sentence/proposition \citep{Moens1987}. Within the lexical structure of a verb, there can be properties that present some boundary or limit (with respect to the duration of the event described by the verb) while this is not the case for others verbs. A verb like `to explode' or `to sneeze', would, without any further context, be translated with a perfective tense since the action is completed the moment it happens.~\footnote{Verbs such as `to sneeze' or `to explode' are sometimes also classified as semelfactive verbs~\citep{Kiss2011}.} Other verbs like `to own', `to want' do not put such an emphasis on the beginning or the end of an event and are more easily linked with an imperfective aspect. In Example~(\ref{lexAspect2}) we provide some example sentences in French that illustrate this possible interaction between lexical aspect and grammatical aspect. 

\begin{li}\label{lexAspect2}
\item
{\centering
\begin{tabular}{rlll}
(a) 1. &``It exploded.''\\
    2. &``\c{C}a \textbf{a explos\'{e}}.'' & [PER.]\\
~\newline\newline\\
(b) 1. &``It exploded continuously.''\\
    2. &``\c{C}a \textbf{explosait} sans cesse.'' & [IMP.]\\
~\newline\newline\\
(c) 1. &``I owned a car.'' \\
    2. &``Je \textbf{poss\'{e}dais} une voiture.'' & [IMP.]\\
~\newline\newline\\
(d) 1. &``I owned a car for two years.'' \\
    2. &``J'\textbf{ai poss\'{e}d\'{e}} une voiture pendant deux ans.'' & [PER.]\\
~\newline\newline
\end{tabular}}
\end{li}

Whereas the English verb `exploded' in (a) is prototypically linked with a perfective interpretation (and thus translated in French with a \emph{pass\'{e} compos\'{e}}), the verb `owned' in (c) has semantic properties that are more closely related to the imperfective aspect. However, an adverb (\emph{continuously} in (b)) or prepositional phrase (\emph{for two years} in (d)) can change the overall aspectual value of the proposition. The sentences in Example~(\ref{lexAspect2}) illustrate that `grammatical aspect' and `lexical aspect' are intertwined.

Verbs can be grouped together in a taxonomy of aspectual classes according to properties related to their `lexical aspect'. Wilmet's taxonomy~\citep{Wilmet1997} classifies verbs into three different aspectual classes: \emph{stative} verbs (e.g. `to own', `to love', `to believe', `to want'), \emph{telic\footnote{Telic verbs are verbs that refer to events/situations with an inherent goal or ending while atelic verbs do not have this property~\citep{Garey1957}.} dynamic} verbs (`to cough',`to deliver') and \emph{atelic dynamic} verbs (e.g. `to eat', `to write', `to walk'). Unless the context contains important triggers that suggest otherwise, stative verbs will be more likely translated into an imperfective tense. Likewise, telic activity verbs are prototypically translated into a perfective tense. However, this classification is not fixed. It is susceptible to aspectual triggers provided by the context. That is, verbs can transition from one class to another given the right triggers in the context (as illustrated in Example~(\ref{lexAspect2})). This implies that an automatic translation system might need the entire source sentence (or even its surrounding sentences) in order to determine correctly the aspectual value of a verb.

PB-SMT learns to translate phrases (\emph{n}-grams) of the source language to target language phrases based on their co-occurrence frequencies in a parallel corpus. The size of those \emph{n}-grams is usually limited to 6. All source-language phrases and their target-language counterparts are stored in phrase-tables together with their probabilities. Every phrase is seen as an atomic unit and thus translated as such. Given that these units are limited in length and not linguistically motivated but based purely on statistics, important linguistic information is lost. When it comes to determining the aspectual value of a verb or proposition in English, PB-SMT can, on the source side, rely only on the limited information contained in those phrase-tables. 

Neural Machine Translation (NMT)~\citep{Bahdanau2014,Cho2014,Sutskever2014} encodes the entire source sentence in an encoding vector. In an encoder-decoder NMT model, there are two neural networks at work. The first one encodes information about the source sentence into a vector of real-valued numbers (the hidden state). The second neural network decodes the hidden state into a target sentence. Unlike PB-SMT, the neural network responsible for the decoding of the hidden state has access to a vector that contains information about the entire source sentence. This means that the `knowledge source' of NMT systems is the encoding vector which is supposed to contain all the necessary information to correctly translate the source into a target sentence.\footnote{We would like to add that we are talking about translating sentences in isolation.}\\ 
In this study, we want to examine how PB-SMT and NMT, relying on different `knowledge sources' (phrase-tables vs encoding-vectors), deal with a covert semantic category on the sentential level such as `aspect'. For PB-SMT, we know that its knowledge source is in theory, insufficient. The phrase-tables cannot always cover all the necessary contextual information in order to determine the correct aspectual value of a sentence. The phrase-tables can, however, reflect something about the lexical aspect of the verbs it contains. The probabilities stored in the phrase-tables of an English-Spanish or English-French MT system should be able to reflect whether the verb has a strong preference for an imperfective or perfective tense (or not). In order to verify whether this is the case, we compiled a list of 206 English verbs. The verbs were classified into their prototypical or basic aspectual class (following the taxonomy proposed by~\cite{Wilmet1997}). We then verified whether the phrase-tables reflect the connection between the aspectual classes and the grammatical aspect expressed by the French/Spanish tenses. Unlike PB-SMT, NMT does encode the entire sentence at once and should theoretically have sufficient\footnote{When translating paragraphs, information of the previous and/or following sentences might be required in order to determine the correct tense. Hardmeier, Nivre and Tiedemann~\citeyearpar{Hardmeier2012} developed a decoding algorithm that can handle cross-sentence features. A detailed discussion and overview of discourse phenomena can be found in the thesis of Hardmeier~(\citeyear{Hardmeier2014}).} information at its disposition to decode the sentence correctly. However, NMT's encoding vectors that consist of a large number of real-valued numbers are very hard to understand and interpret. Inspired by the work of Shi et al.~\citeyearpar{Shi2016}, we used a logistic regression model trained on the encoding vectors to verify whether the encodings contain aspectual information.

Once we have a better understanding of what aspectual knowledge PB-SMT and NMT decoders have at their disposal, we aim to see how this is reflected in the actual translations. We hypothesize that PB-SMT performs well in the prototypical cases where the grammatical aspect reflects the lexical aspect of a verb. However, it will fail to render the correct grammatical aspect for the more complex cases where other contextual elements (adverbial phrases, prepositional phrases etc.) change the aspectual meaning of the verb. For NMT, formulating a hypothesis is more complicated. We know that the source sentence vectors encode the entire sentence but we do not know whether they implicitly store any aspectual information. Furthermore, even if they do contain aspectual information, it might still be lost during the decoding process.

The remainder of this chapter is structured as follows: Section \ref{relWork} discusses related work on `aspect' in MT; Section \ref{sec:experiments} explains in more detail the set-up of our experiments for PB-SMT and NMT as well as their outcome; the results are described in Section~\ref{sec:resultsaspect}; finally, Section \ref{sec:conclusions} presents the conclusions we draw from our work.

\section{Related Work}\label{relWork}
\subsection{Statistical Machine Translation}
Although tense, aspect and mood (often referred to as `TAM') have received a lot of attention in linguistic fields such as formal semantics and logic~\citep{mccawley1971tense,Richards1982}, within the field of data-driven MT (SMT and NMT) there has been little research on tense, aspect and mood. However, in knowledge-based MT systems such as Eurotra and Rosetta, comprehensive and detailed modules were included in order to deal with TAM-related translation difficulties~\citep{VanEynde1985,Appelo1986,VanEynde1988}. Van Eynde~\citeyearpar{VanEynde1985,VanEynde1988} integrates tense information by mapping the analysis of tense and aspect onto their meanings. Since many meanings can be assigned to one form, the assignment of meaning is followed by a disambiguation step where context factors (e.g. temporal adverbials and the Aktionsart of the basic proposition) are taken into account. Although Eurotra is a transfer-based RBMT system, the representations of tense and aspect are interlingual, which implies that their meaning can simply be copied during the actual transfer. Appelo~\citeyearpar{Appelo1986} intends to solve the problem of translating temporal expressions in natural languages within the Rosetta MT system framework. The Rosetta MT system uses the `isomorphic grammar' method which attunes the grammars of languages such that ``a sentence \textit{s} is a translation equivalent of a sentence \textit{s'} if \textit{s} and \textit{s'} have similar derivational histories'' \cite[p.1]{Appelo1986}.

Within the field of PB-SMT, the works focusing on specific problems related to tense and aspect remain scarce. There has been some research on tense prediction: Ye and Zhang~\citeyearpar{Ye2005} focus on unbalanced levels of temporal reference between Chinese and English. They build a tense classifier in order to predict the English tense given a Chinese verb based on several lexical and syntactic features. Their tense classifier was trained on manually labeled data. However, they did not build their tense classifier into an MT system. Similarly Gong et al.~(\citeyear{Gong2012}) focuses on tense prediction. They used an \emph{n}-gram-based tense model in order to predict the appropriate tense. Focusing particularly on the English-French language pair, Loaiciga et al.~\citeyearpar{Loaiciga2014} developed a method for the alignment of verbs phrases by using GIZA++~\citep{Och2003}, a POS-tagger, a parser and several heuristics. They labeled the verb phrases (VPs) with their tense and voice on both sides of the parallel text. Once the VPs are aligned and labeled, a tense predictor is trained on the labeled data based on several features.

Only a handful of research studies focused on particular problems related to `aspect' in MT: Ye et al.~\citeyearpar{Ye2007} report on a study of aspect marker generation for the English-Chinese language pair. They train an aspect marker classifier based on a maximum entropy model and achieve an overall classification accuracy of 78$\%$.  Meyer et al~\citeyearpar{Meyer2013} worked on the disambiguation of the \emph{pass\'{e} compos\'{e}} and the \emph{imparfait} when translating from English into French by focusing on narrativity. Their Maximum Entropy classifier was trained on data that was manually annotated for narrativity. By training a classifier to predict \emph{narrativity} and by consequently integrating this as a feature into a factored PB-SMT system~\citep{Koehn2007}, Meyer et al.~\citeyearpar{Meyer2013} obtain a small BLEU score improvement.

\subsection{Neural Machine Translation}
No other studies have been done on `tense' and `aspect' for NMT. Tense, aspect and mood have, however, been mentioned in linguistic evaluation studies, published after we conducted our experiments. In 2017, a linguistic evaluation study~\citep{Burchardt2017} comparing PB-SMT and NMT in a more systematic way, included verb tense, aspect and mood as a category in their preliminary version of a test suite for English-German and their following manual evaluation. The analysis shows how the GNMT system clearly outperforms Google's previous PB-SMT system over all categories. An interesting and more relevant observation with respect to our objective, is that the RBMT~\citep{Alonso2003} and the GNMT system are the two best-performing systems on average in terms of the linguistic categories evaluated. The RBMT system is the best system for handling ambiguity and verb tense, aspect and mood translations (82\% vs 96\% correctness). This is possibly the case because verb paradigms are part of the linguistic information RBMT systems are based on~\citep{Burchardt2017}. We would like to point out that: (a) as our evaluation in Chapter~\ref{ch:Agreement}, they solely focus on the phenomena to be evaluated in the respective sentences and thus disregard other errors (e.g. overall fluency of the sentence) and (b) their 2017 evaluation was done on a limited amount of samples and so might not be conclusive as a more extensive and systematic evaluation needs to be conducted in order to draw general observations and allow for more quantitative statements. 

An extension of \cite{Burchardt2017} was published in \cite{Macketanz2018} where they perform a more fine-grained evaluation of the German-English test suite.\footnote{We descibed the results for English-German for~\cite{Burchardt2017} as it is closer to our work, i.e. translating from English into a morphologically richer language. However, in \cite{Burchardt2017} they also present results for German-English and the results for verb tense, aspect and mood are very comparable to those obtained for English-German.} They evaluate 106 German grammatical phenomena and focus on verb-related issues employing three different types of MT systems (RBMT, PB-SMT and NMT) by comparing the performance of 16 different state-of-the art systems. According to their analysis, verb tense, aspect and mood remain problematic linguistic categories overall with average accuracies around 75\%. The one system that stood out in terms of verb-related phenomena was the UCAM or SGNMT  system~\citep{Stahlberg2016,Stahlberg2017} with an accuracy of 86.9\% for verb tense, aspect and mood. UCAM is a hybrid MT system that combines NMT with PB-SMT components. Different to our work, German does not mark aspect formally and has no equivalent to the Romance imperfect-perfect distinction in verb forms.

In order to reveal how much aspectual information is stored in NMT encoding vectors, we were inspired by the work of Shi et al.~\citeyearpar{Shi2016}. Their work uses the high-dimensional encoding vectors of a sequence-to-sequence model (\emph{seq2seq}) to predict sentence-level labels. By training a logistic regression model on a set of labeled encoding vectors, they show that local and global syntactic information is contained within these vectors, but other types of syntactic information is still missing (e.g. subtleties such as attachment errors). Their logistic regression model aiming to predict the voice (active/passive) using encoding cell states achieved an accuracy of 92\%. However, voice is an overt phenomenon in English expressed by specific verbal constructions. It could be, therefore, that the model does not `learn' but just preserves information about the word forms in the encoding vector. To learn more about aspect in NMT we perform a similar experiment with a more covert phenomenon (aspect) that can manifest itself in many different ways (verbal aspect, verb structure, adverbial phrases etc.).

\section{Experiments}\label{sec:experiments}

\subsection{Statistical Machine Translation}

\subsubsection{Compilation of Verbs}

We compiled a list of 206 English verbs from linguistic sources and classified them by their `basic' lexical aspect according to Wilmet's taxonomy~\citep{Wilmet1997}.\footnote{We opted for Wilmet's taxonomy instead of for the more well-known classes of Vendler~\citeyearpar{Vendler1957} (who distinguishes between 4 different aspectual verb classes: \emph{states}, \emph{activities}, \emph{achievements} and \emph{accomplishments}). Wilmet groups together Vendler's \emph{achievements} and \emph{accomplishments} based on their `telic' feature. We believe this makes Wilmet's classes more generalizable and thus more appropriate for our purpose.} According to Wilmet's taxonomy there are three basic lexical aspectual classes: \emph{stative}, \emph{dynamic telic} or \emph{dynamic atelic}. Classifying verbs into their basic lexical class is not always straightforward. One could argue that a verb like e.g. `to run' or `to drive' (and many more) can be both telic or atelic dynamic verbs as illustrated in Example~(\ref{exmp:smt_telic_atelic}):

\begin{li}\label{exmp:smt_telic_atelic}
\item 
\begin{tabular}{rll}
(a) & ``I ran.'' & ATELIC \\
(b) & ``I ran a mile.'' & TELIC\\
~\newline\newline
\end{tabular}
\end{li}

However, our classification is based on a verb's occurrence in its most basic proposition. We classify a verb as stative when it does not undergo any changes in between its initial and final stage. When a change does occur, as in Example~(\ref{exmp:smt_telic_atelic}), we classify the verb as dynamic.  Since the verb `to run' does not denote an inherent end-point in its most basic proposition (\ref{exmp:smt_telic_atelic}a), we classify `to run' as a dynamic atelic verb.

\subsubsection{Description of PB-SMT system}

The PB-SMT systems are built with the Moses toolkit \citep{Koehn2007}. The data is tokenized and lowercased using the Moses tokenizer and lowercaser. Sentences longer than 60 tokens are filtered out. For training, we use the default Moses settings. We trained three systems on 1 million parallel sentences of: (1) the Europarl corpus \citep{Koehn2005}, (2) the News Commentary corpus \citep{Tiedemann2012} and (3) The Open Subtitles corpus \citep{Tiedemann2009} for two language pairs: English--French and English--Spanish.

\subsubsection{Extracting Information from Phrase-Tables}\label{extract}

The core component of phrase-based translation models and the main knowledge source for the PB-SMT decoder are the phrase-tables, which contain the probabilities of translating a word (or a sequence of words) from one language into another. 
All the knowledge that phrase-tables contain is extracted from the word and phrase alignments obtained from the parallel data they were trained on. Table~\ref{exmpl:phrase_table} shows phrase-translations extracted from a phrase-table trained on Europarl data:

\vspace{0.5cm}
\begin{table}[h!]
\centering
 \setlength\tabcolsep{3pt}\begin{tabular}{lllllll}
\noindent
  & \textcolor{white}{$|||$}  & \textcolor{white}{$|||$} p(en\textbar fr) & lex(en\textbar fr) & \textbf{p(fr\textbar en)} & lex(fr\textbar en) &  \\
worked & $|||$ ont abattu & $|||$ 0.2 & 0.0078406 & \textbf{0.00085034} & 4.06111e-05 & \\
worked & $|||$ ont collabor\'{e} & $|||$ 0.0952381 & 0.13217 & \textbf{0.00340136} & 0.0013808  &  \\
worked & $|||$ ont fonctionn\'{e} ,\footnotemark & $|||$ 1 & 0.211969 & \textbf{0.00085034} & 0.000345333 & \\
worked & $|||$ ont travaill\'{e} & $|||$ 0.037037 & 0.183833 & \textbf{0.00085034} & 4.57342e-05 &  \\
worked & $|||$ travaillait & $|||$ 0.275862 & 0.148649 & \textbf{0.00680272} & 0.004663 & \\
worked & $|||$ travaillant & $|||$ 0.00671141 & 0.0110865 & \textbf{0.00170068} & 0.0021195 &  \\
\end{tabular}
\caption{Example of phrase-translation extracted from a phrase-table trained on Europarl data}\label{exmpl:phrase_table}
\end{table}

\footnotetext{Although this is a relatively `clean' fraction of a phrase-table, phrase-tables do not only include a lot of lexical variation (`abbatu' vs `travaill\'{e}') and morphological variation (`travaillait' vs `travaillant') but also often include function words (`the' , `an' , ...) and noise (`,',`it'). [\url{http://www.statmt.org/book/slides/05-phrase-based-models.pdf}]. }

As can be seen in Table~\ref{exmpl:phrase_table}, the possible phrase translations are followed by four scores: the inverse phrase translation probability (p(en\textbar french)), the inverse lexical weighting (lex(en\textbar fr)), the direct phrase translation probability (p(fr\textbar  en)) and the direct lexical weighting (lex(french \textbar english)). We are interested in the probability of the French word (or phrase) given an English word, i.e. the direct phrase translation probability (p(fr\textbar en)). We extracted all \emph{imparfait} and \emph{pass\'{e} compos\'{e}} translations of the English verbs and collected their probabilities. 

We first tagged the phrases with the POS-tagger provided by the python package \verb|polyglot|.\footnote{\url{http://polyglot.readthedocs.io/en/latest/}} The pre-trained POS-tagger recognises the 17 universal parts of speech for several languages including Dutch, French and Spanish (Castilian).\footnote{\url{http://polyglot.readthedocs.io/en/latest/POS.html}} Afterwards, we identified the perfective tenses by searching for phrases that contain a conjugated present tense auxiliary verb `to have' (`avoir', `haber' or `hebben' for French, Spanish and Dutch, respectively) followed by a past participle.\footnote{We allowed for insertion of adverbs and/or pronouns in between the auxiliary and the participle.} To cover the exceptions, for the French pronominal verbs as well as for 14 other verbs (and their derivatives\footnote{Commonly known as the verbs of `La Maison d'\^{e}tre': `aller',`(r)entrer',`passer',`(re)monter',`(re)tomber',`arriver',`(re)na\^{i}tre',`(re)descendre',`(re\textbar de\textbar par)venir',
`(res)sortir',`retourner',`rester',`(re)partir',`mourir'}), the auxiliary verb `\^{e}tre' (`to be') was used. The imperfective tense was identified by extracting only those verbs with particular endings\footnote{French: `-ais',`-ait',`-ions',`-iez',`-aient'; Spanish:`-aba',`-abas',`-\'{a}bamos',`-abais',`-aban',`-ía',`-íamos',`-ías',`-íais',`-ían' Dutch: `-te(n)',`-de(n)'.} characteristic of the imperfective tense of the language in question. We included the irregular conjugations that are not covered by the general endings.\footnote{Verbs such as `ir',`ser' and `ver' (Spanish) and the so-called `strong' Dutch verbs.} Since some verbs in the present indicative, the present subjunctive and the present conditional tense in Spanish, French and Dutch have endings that overlap with those of the imperfective tenses, we cleaned the extracted phrases afterwards and added the false positives to a list in order to remove them from our results.

By then dividing both the added \emph{pass\'{e} compos\'{e}} and \emph{imparfait} values by the total (\emph{imparfait} + \emph{pass\'{e} compos\'{e}}), we normalize the probabilities and obtain the probability of the \emph{pass\'{e} compos\'{e}} and/or \emph{imparfait} given the total \emph{pass\'{e} compos\'{e}} and \emph{imparfait} translations  of  a specific verb. Table~\ref{tbl:prob_results} below illustrates the different tense probabilities of the verbs `promised', `hit', `saw' and `thought' based on the information extracted from PB-SMT phrase-tables trained on 1M sentences of Europarl, the News Commentary Corpus and the OpenSubtitles Corpus for English--French.

\begin{table}[ht]
\begin{center}
 {\setlength\tabcolsep{5pt}\begin{tabular}{|l|c|c|c||c|c|c|}\hline
    \multirow{2}{*}{\% Verbs} 		& \multicolumn{3}{c||}{Imparfait} 		& \multicolumn{3}{c|}{Pass\'{e} compos\'{e}} 			                \\\cline{2-7}
				& NEWS			    &EUROPARL       &OpenSubs 		    &NEWS			    &EUROPARL	        &OpenSubs                   \\\hline\hline		
    promised	& 20.63			    & 6.12       	& 1.07              & \textbf{79.37}	& \textbf{93.88}    & \textbf{98.93}			\\\hline
    hit  		& 7.14			    & 0.00			& 3.31              & \textbf{92.86}    & \textbf{100.00}   & \textbf{96.69}   		    \\\hline
    said			& 16.98			    & 11.47 			& 9.53              & \textbf{83.02}	& \textbf{88.53}    & \textbf{90.47}			\\\hline
    thought		& \textbf{74.32}	& \textbf{72.78}& \textbf{62.70}    & 25.68			    & 27.22             & 37.30			            \\
    \hline
 \end{tabular}}
\end{center}
 \caption{\emph{Imparfait} and \emph{pass\'{e} compos\'{e}} percentages for the verbs \emph{promised}, \emph{hit}, \emph{saw} and \emph{thought}.}\label{tbl:prob_results}
\end{table}

As can be seen in Table~\ref{tbl:prob_results}, these four English verbs (`promised', `hit', `said' and `thought') have a clear preference for one particular tense in French according to the information extracted from the phrase-tables. Except for the verb `thought', which has a stative lexical aspect, the verbs `hit' (telic dynamic), `promised' (telic dynamic) and `said' (atelic dynamic) are more commonly translated into \emph{pass\'{e} compos\'{e}} than into \emph{imparfait}. This is the case in all three corpora. However, we do also observe some differences across the corpora when looking at these particular verbs, e.g. in the NEWS domain the verbs are translated more often into an imperfective tense compared to the Europarl and OpenSubtitles corpora. These differences can be explained by the fact that the Europarl and OpenSubtitles corpora often contain reported speech and the perfective tense (\emph{pass\'{e} compos\'{e}}) is linked more closely to the spoken domain \citep{labelle1987utilisation,Grisot2012}.\footnote{This illustrates as well how the use of particular tenses can depend not only on their immediate context but also on a broader context, paragraph or even on an entire domain.} We ought to note that we are working with parallel corpora and the data contained within such corpora might be susceptible to well-studied phenomena in human translation such as interference of the source on the target \citep{Xiao2015}.\footnote{Furthermore, the original language of the data can be either of the two languages involved (or even another language as it is possible that both the English or French data has already been translated), i.e. when working with an English-French corpus, the original data could have been English that was translated into French, or vice versa. \cite{Ozdowska2009} showed how the different derivations of Europarl sentences can have a significant impact on MT quality.}

Together with the information about the French past tenses the verbs can be translated into, we also extract all the possible \emph{imparfait} and \emph{pass\'{e} compos\'{e}} translations of the verb in order to (later on during our research) use the translations to label English source sentences according to the grammatical aspect expressed by their corresponding reference target sentences. By storing all verbs with their translations, we are able to automatically label sentences and evaluate the outputs of our translation systems. An example of the possible translations extracted from the phrase-tables for one verb is given in Example~(\ref{ex:translations}):

\begin{li}\label{ex:translations}
 \item \textbf{Verb: }``felt'' 
 \item \textbf{Pass\'{e} Comp.: } ``a compris''; ``a consid\'{e}r\'{e}''; ``a cru''; ``a estim\'{e}''; ``a jug\'{e}''; ``a paru''; ``a pens\'{e}''; ``a ressenti''; ``a sembl\'{e}''; ``a senti''; ``a tenu''; ``a trouv\'{e}''; ``ai consid\'{e}r\'{e}''; ``ai estim\'{e}''; ``ai jug\'{e}es''; ``ai not\'{e}es''; ``ai pens\'{e}''; ``ai ressenti''; ``ai senti''; ``ai trouv\'{e}''; ``avons consid\'{e}r\'{e}''; ``avons estim\'{e}''; ``avons jug\'{e}''; ``avons pens\'{e}''; ``avons ressenti''; ``ont consid\'{e}r\'{e}''; ``ont estim\'{e}''; ``ont pens\'{e}''; ``ont ressenti''; ``ont senti''; ``ont trouv\'{e}''; ``s' est senti''; ``s' est sentie''; ``s' est vue'' 
 \item \textbf{Imparfait :}  ``craignaient''; ``estimaient''; ``estimait''; ``jugeait''; ``paraissait''; ``pensaient''
\end{li}

\subsubsection{Aspectual information in PB-SMT}\label{sec:Aspect_info_SMT}

We further analyzed how the lexical aspect of English verbs correlates with the grammatical aspect expressed by the French (see Table~\ref{tbl:english-french-sp} and Table~\ref{tbl:english-french-pp}) and Spanish (see Table~\ref{tbl:english-spanish-sp} and Table~\ref{tbl:english-spanish-pp}) tenses. Based on the information extracted from the Europarl, OpenSubs and News corpora, we analyzed which tense appeared overall more often when translating from an English \emph{simple past}/\emph{present perfect}. We included the results for all verbs that had a `strong' (\textgreater67\%) preference for the perfective tense (\emph{pass\'{e} compos\'{e}}) or the imperfective tense (\emph{imparfait}). 
As can be seen in Table \ref{tbl:english-french-sp} and Table~\ref{tbl:english-french-pp}, the French \emph{pass\'{e} compos\'{e}} appears more often than the \emph{imparfait} in all corpora. Furthermore, with respect to the lexical aspect classes assigned to the English verbs, we see some clear patterns. The \emph{pass\'{e} compos\'{e}} is the most used verb tense when translating an English \emph{simple past} or \emph{present perfect} verb. It also appears to be the more flexible one: both stative and dynamic (telic and atelic) verbs can have a preference for \emph{pass\'{e} compos\'{e}} in French. These findings are in accordance with the literature stating that the perfective viewpoint is dominant in French \citep{Smith2013}. The results also correspond with those from a previous contrastive linguistics study by Grisot and Cartoni~\citeyearpar{Grisot2012} based on a corpus containing texts belonging to different domains (journalistic, judicial, literary and administrative). Grisot and Cartoni~\citeyearpar{Grisot2012} calculated the frequencies of the French tenses given the English \emph{simple past} and \emph{present perfect} and, in our experiments, in both the News and Europarl corpora the \emph{pass\'{e} compos\'{e}} dominated, especially when translating from an English \emph{present perfect} verb. For the English--French data, we also observe that the judicial and spoken domains (Europarl and OpenSubtitles) contain more verbs that have a preference for the perfective tense compared to the News domain.

While the usage of the \emph{pass\'{e} compos\'{e}} extends over different lexical aspect classes, the \emph{imparfait} has a clear preference for stative verbs. English atelic dynamic verbs can also be translated into a French \emph{imparfait}, but, our analysis revealed that none of the telic dynamic \emph{present perfect} verbs were translated as an \emph{imparfait}. Their telicity is difficult to unite with the imperfectivity of the French \emph{imparfait}.

From the analysis above, performed on data extracted from PB-SMT phrase-tables, we conclude that PB-SMT's knowledge source indirectly possesses information about the lexical aspect of verbs. Furthermore, there is indeed a relation between the English lexical aspect assigned to verbs and the grammatical aspect of the French tense they are translated into.


\begin{table}[ht]
\centering
\begin{tabular}{|l||r|r|r|}\hline
ENGLISH - & \multicolumn{3}{c|}{SIMPLE PAST} \\\cline{2-4}
FRENCH & EUROPARL & OPENSUBS & NEWS \\\hline
\multicolumn{1}{|c||}{PASS\'{E} COMPOS\'{E}} & \multicolumn{1}{c|}{73$\%$} & \multicolumn{1}{c|}{73$\%$} & \multicolumn{1}{c|}{63$\%$} \\\hline
stative & 22$\%$ & 15$\%$ & 20$\%$ \\\hline
atelic & 45$\%$ & 45$\%$ & 41$\%$ \\\hline
telic & 33$\%$ & 39$\%$ & 40$\%$ \\\hline\hline
\multicolumn{1}{|c||}{IMPARFAIT} & \multicolumn{1}{c|}{27$\%$} & \multicolumn{1}{c|}{27$\%$} & \multicolumn{1}{c|}{37$\%$} \\\hline
stative & 75$\%$ & 100$\%$ & 100$\%$ \\\hline
atelic & 0$\%$ & 0$\%$ & 0$\%$ \\\hline
telic & 25$\%$ & 0$\%$ & 0$\%$ \\\hline
\end{tabular}
\caption{English lexical verb classes versus the grammatical aspect of French tenses for translations of English simple present verbs.}\label{tbl:english-french-sp}
\end{table}

\begin{table}[ht]
\centering
\begin{tabular}{|l||r|r|r|}\hline
ENGLISH - & \multicolumn{3}{c|}{PRESENT PERFECT} \\\cline{2-4}
FRENCH & EUROPARL & OPENSUBS & NEWS \\\hline
\multicolumn{1}{|c||}{PASS\'{E} COMPOS\'{E}} & \multicolumn{1}{c|}{98$\%$} & \multicolumn{1}{c|}{99$\%$} & \multicolumn{1}{c|}{98$\%$} \\\hline
stative & 11$\%$ & 10$\%$ & 9$\%$ \\\hline
atelic & 49$\%$ & 45$\%$ & 42$\%$ \\\hline
telic & 40$\%$ & 45$\%$ & 49$\%$ \\\hline\hline
\multicolumn{1}{|c||}{IMPARFAIT} & \multicolumn{1}{c|}{2$\%$} & \multicolumn{1}{c|}{1$\%$} & \multicolumn{1}{c|}{2$\%$} \\\hline
stative & 93$\%$ & 84$\%$ & 84$\%$ \\\hline
atelic & 7$\%$ & 16$\%$ & 16$\%$ \\\hline
telic & 0$\%$ & 0$\%$ & 0$\%$ \\\hline
\end{tabular}
\caption{English lexical verb classes versus grammatical aspect of French tenses for translations of English present perfect verbs.}\label{tbl:english-french-pp}
\end{table}

We performed a similar analysis on the information extracted from the English--Spanish phrase-tables trained on the same corpora (Europarl, News, and OpenSubtitles) producing comparable results. These results are summarized in Table~\ref{tbl:english-spanish-sp} and Table~\ref{tbl:english-spanish-pp}. As in French, the Spanish past tenses \emph{pret\'{e}rito indefinido} and \emph{pret\'{e}rito imperfecto} are linked with different grammatical aspect. Unlike French, where the \emph{pass\'{e} compos\'{e}} became an equivalent (and almost completely replaced) the \emph{simple past} tense, the \emph{pret\'{e}rito indefinido} is still widely used. Although Spanish has a past tense that is formally very similar to the \emph{pass\'{e} compos\'{e}} (\emph{pret\'{e}rito perfecto}), the perfective aspect is marked by the \emph{pret\'{e}rito indefinido} while the imperfective aspect is marked by the \emph{pret\'{e}rito imperfecto}.

Looking at Table~\ref{tbl:english-spanish-sp} and Table~\ref{tbl:english-spanish-pp}, it results that the Spanish \emph{pret\'{e}rito indefinido} and \emph{pret\'{e}rito imperfecto} behave similarly to the French past tenses. As in the French data, the perfective tense is dominant in all corpora for both \emph{simple past} and \emph{present perfect} verbs. The perfective past tense (\emph{pret\'{e}rito indefinido}) occurs with verbs from all lexical aspect classes while the imperfect past tense (\emph{pret\'{e}rito imperfecto}) appears only as a translation of stative and atelic dynamic verbs, with a clear preference for stative verbs. 

Both the English--French and English--Spanish tables showed the dominance of the perfective tense over the imperfective one as well as the limited use of imperfective tenses with respect to certain verb classes (specifically the telic dynamic verbs). Given the fact that telic verbs present a completed action and the \emph{imparfait} and \emph{imperfecto} have an imperfective aspect presenting something ongoing, the lexical and grammatical aspect do not match, which explains why we had no occurrences in our data of telic dynamic verbs being translated into a tense with an imperfective aspect. The fact that telic verbs in Romance languages do not combine well with an imperfective tense was also noted by King and Su\~{n}er~\citeyearpar{King1980}.


\begin{table}[ht]
\centering
\begin{tabular}{|l||r|r|r|}\hline
ENGLISH - & \multicolumn{3}{c|}{SIMPLE PAST} \\\cline{2-4}
SPANISH & EUROPARL & OPENSUBS & NEWS \\\hline
\multicolumn{1}{|c||}{INDEF.} & \multicolumn{1}{c|}{75$\%$} & \multicolumn{1}{c|}{71$\%$} & \multicolumn{1}{c|}{74$\%$} \\\hline
stative & 12$\%$ & 11$\%$ & 15$\%$ \\\hline
atelic & 44$\%$ & 43$\%$ & 42$\%$ \\\hline
telic & 45$\%$ & 46$\%$ & 44$\%$ \\\hline\hline
\multicolumn{1}{|c||}{IMPERF.} & \multicolumn{1}{c|}{25$\%$} & \multicolumn{1}{c|}{29$\%$} & \multicolumn{1}{c|}{26$\%$} \\\hline
stative & 82$\%$ & 83$\%$ & 84$\%$ \\\hline
atelic & 18$\%$ & 17$\%$ & 16$\%$ \\\hline
telic & 0$\%$ & 0$\%$ & 0$\%$ \\\hline
\end{tabular}
\caption{English lexical verb classes versus grammatical aspect of Spanish tenses for translations of English simple past verbs.}\label{tbl:english-spanish-sp}
\end{table}

\begin{table}[ht]
\centering
\begin{tabular}{|l||r|r|r|}\hline
ENGLISH - & \multicolumn{3}{c|}{PRESENT PERFECT} \\\cline{2-4}
SPANISH & EUROPARL & OPENSUBS & NEWS \\\hline
\multicolumn{1}{|c||}{INDEF.} & \multicolumn{1}{c|}{81$\%$} & \multicolumn{1}{c|}{97$\%$} & \multicolumn{1}{c|}{97$\%$} \\\hline
stative & 14$\%$ & 9$\%$ & 15$\%$ \\\hline
atelic & 39$\%$ & 44$\%$ & 41$\%$ \\\hline
telic & 47$\%$ & 47$\%$ & 44$\%$ \\\hline\hline
\multicolumn{1}{|c||}{IMPERF.} & \multicolumn{1}{c|}{19$\%$} & \multicolumn{1}{c|}{3$\%$} & \multicolumn{1}{c|}{3$\%$} \\\hline
stative & 83$\%$ & 0$\%$ & 50$\%$ \\\hline
atelic & 17$\%$ & 100$\%$ & 50$\%$ \\\hline
telic & 0$\%$ & 0$\%$ & 0$\%$ \\\hline
\end{tabular}
\caption{English lexical verb classes versus grammatical aspect of Spanish tenses for translations of English present perfect verbs.}\label{tbl:english-spanish-pp}
\end{table}

In order to verify our method, we performed the same experiment with English--Dutch data.\footnote{We used the exact same training data for our Dutch MT system as for our English--French and English--Spanish systems. However, the NEWS corpus is not available in Dutch which is why we do not provide data from the news domain.} English and Dutch have a similar verb system with tenses that are inflected similarly. There are some cases where translating the \emph{simple past} or \emph{present perfect} tense from English into Dutch might result in a negative transfer, but Dutch past tenses (unlike the French and Spanish ones) do not grammaticalize aspect. The results shown in Table~\ref{tbl:english-dutch-sp} and Table~\ref{tbl:english-dutch-pp} confirm this. There is almost a one-to-one correspondence between the English \emph{simple past} and the Dutch \emph{onvoltooid verleden tijd} (OVT): in Europarl, 90\% of the English simple past verbs are translated by a Dutch OVT, and in the OpenSubtitles the figure rises to 97\%. Similarly, the English \emph{present perfect} `matches' the Dutch \emph{voltooid tegenwoordige tijd} (VTT) with 97\% (Europarl) and 95\% (OpenSubtitles).


\begin{table}[ht]
\centering
\begin{tabular}{|l||r|r|}\hline
ENGLISH - & \multicolumn{2}{c|}{SIMPLE PAST} \\\cline{2-3}
DUTCH & EUROPARL & OPENSUBS \\\hline
\multicolumn{1}{|c||}{OVT.} & \multicolumn{1}{c|}{\bf{90$\%$}} & \multicolumn{1}{c|}{\bf{97$\%$}} \\\hline
stative & \bf{42$\%$} & \bf{27$\%$} \\\hline
atelic & \bf{36$\%$} & \bf{46$\%$} \\\hline
telic & \bf{22$\%$} & \bf{27$\%$} \\\hline\hline
\multicolumn{1}{|c||}{VTT} & \multicolumn{1}{c|}{10$\%$} & \multicolumn{1}{c|}{3$\%$}  \\\hline
stative & 0$\%$\footnotemark & 0$\%$  \\\hline
atelic & 70$\%$ & 0$\%$  \\\hline
telic & 30$\%$ & 100$\%$   \\\hline
\end{tabular}
\caption{English lexical verb classes versus Dutch tenses for English simple past verbs.}\label{tbl:english-dutch-sp}   
\end{table}

\begin{table}[ht]
\centering
\begin{tabular}{|l||r|r|}\hline
ENGLISH - & \multicolumn{2}{c|}{PRESENT PERFECT} \\\cline{2-3}
DUTCH & EUROPARL & OPENSUBS \\\hline
\multicolumn{1}{|c||}{OVT.} & \multicolumn{1}{c|}{3$\%$} & \multicolumn{1}{c|}{5$\%$} \\\hline
stative & 60$\%$ & 25$\%$ \\\hline
atelic & 40$\%$ & 50$\%$ \\\hline
telic & 0$\%$ & 25$\%$ \\\hline\hline
\multicolumn{1}{|c||}{VTT} & \multicolumn{1}{c|}{\bf{97$\%$}} & \multicolumn{1}{c|}{\bf{95$\%$}}  \\\hline
stative & \bf{19$\%$} & \bf{17$\%$}  \\\hline
atelic & \bf{43$\%$} & \bf{42$\%$}  \\\hline
telic & \bf{38$\%$} & \bf{41$\%$}   \\\hline
\end{tabular}
\caption{English lexical verb classes versus Dutch tenses for English present perfect verbs.}\label{tbl:english-dutch-pp}
\end{table}

\footnotetext{\label{percentages}As mentioned in Section \ref{sec:Aspect_info_SMT}, Tables (\ref{tbl:english-dutch-sp} and \ref{tbl:english-dutch-pp}) only present the verbs that had a `strong' (\textgreater67\%) preference for a particular tense. Although it is possible to translate a stative English simple past verb into a Dutch VTT, there were no stative English simple past verbs (in both Europarl and the OpenSubtitles corpora) that had a strong inclination for the VTT.}

\paragraph{Intermediate Conclusion}

According to Vinay~\citeyearpar{Vinay1995} and Filip~\citeyearpar{Filip2012}, aspect should be regarded as a non-lexical property that cannot be assigned to separate words but constitutes a property of an entire sentence. Although we do not disagree with this from a theoretical point of view, we do believe (and see this confirmed by the results in Section \ref{sec:Aspect_info_SMT}) that in practice the lexical aspect of a verb often correlates with the grammatical aspect expressed by Spanish and French past tenses. Some verbs (especially the stative and telic activity verbs) do show a clear preference for a perfective or imperfective tense.

Nonetheless, contextual triggers (e.g. adverbs/adverbial phrases) can influence and change the aspectual `value' of a sentence and thus cause the verb to be translated into the tense with the opposite aspect of its lexical aspect. In the sentences in Example (\ref{exAspect}a)\footnote{The example sentences (in Examples~(\ref{exAspect}) and (\ref{exAspect2})) were extracted from the English-French OpenSubtitles Corpus.}, the telic verb `to lose' is translated in a `neutral' context into the French perfective past tense. In Example~(\ref{exAspect}b), the same verb is translated into the French imperfective past tense due to the aspectual meaning of the adverbial phrase `all the time'. In this case, in a PB-SMT system an \emph{n}-gram of size 5 would be needed in order to have the verb `lost' with the adverbial phrase `all the time' appearing in one phrase. Even if the size of the \emph{n}-grams is set to 5, you would need an exact match with the 5-gram `lost Manny all the time' to be able\footnote{We want to stress the fact that you would be able to retrieve it but there is still no guarantee that that particular 5-gram would eventually be selected.} to retrieve that particular translation.

\begin{li}\label{exAspect}
~\newline
\begin{tabular}{llll}
 (a) & ``I lost.''                          & & TELIC\\
     & ``J'ai perdu.''  \\
 (b) & ``I lost Manny all the time''        & & ATELIC\\
     & ``Je perdais Manny tout le temps.'' 
\end{tabular}
~\newline
\end{li}

Nevertheless some verbs do not show such a clear preference. An atelic activity verb such as `walked', `run' or `eat' can easily be translated into both the perfective and imperfective French and Spanish tenses (even without any contextual triggers). This is illustrated in Example~(\ref{exAspect2}).

\begin{li}\label{exAspect2}
~\newline
\begin{tabular}{llll}
 (a)& ``He ran.'' \\
    & ``Il courrait.'' & ATELIC\\
 (b)& ``He ran.''\\ 
    & ``Il a couru.'' & TELIC
\end{tabular}
~\newline
\end{li}

\noindent
The information extracted from the English--French and English--Spanish phrase-tables from different corpora is in accordance with our hypothesis that verbs (belonging to different lexical aspect classes) often have a preference for a tense connected to a specific grammatical class. However, PB-SMT does not have any means to extract contextual aspectual triggers from the context except for the (limited) \emph{n}-grams stored in the phrase-tables. Therefore, we hypothesize that PB-SMT performs well in terms of selecting the correct past tense in French and Spanish for verbs that have a strong lexical aspect. However, verbs that do not have such a clear lexical aspect and that rely more on the context to select the correct past tense in French and Spanish are most likely to cause more difficulties for a baseline PB-SMT system. The actual translations of our PB-SMT system will be evaluated in Section~\ref{sec:resultsaspect}.

\subsection{Neural Machine Translation}

We will start by describing the NMT systems and the data set we trained on. Afterwards, our logistic regression model will be described in more detail followed by a discussion of its results.

\subsubsection{Description of the NMT system}

We carried out experiment with an encoder-decoder NMT model trained with the toolkit \verb|nematus|~\citep{Sennrich2017}. Our model was trained with the following parameters: \emph{vocabulary size:} $45 000$, \emph{maximum sentence length:} $60$, \emph{vector dimension:} $1 024$, \emph{word dimension:} $500$, \emph{learning optimizer:} \verb|adadelta|. In order to by-pass the OOV problem and reduce the number of dictionary entries we use word-segmentation with BPE. We ran the BPE with $89 500$ operations.

\subsubsection{Aspectual information in NMT}

NMT does store information about the entire sentence in its encoding vectors, unlike PB-SMT. A recent work by Shi et al.~\citeyearpar{Shi2016} uses the high-dimensional encoding vectors of a sequence-to-sequence model (\emph{seq2seq}) to predict sentence-level labels. They show that much syntactic information is contained within these vectors, but, other types of syntactic information is still missing. They trained an NMT system on 110M tokens of bilingual (English--French) data. They created a set of 10K English sentences that were labeled for voice (active or passive) and converted them with their learned NMT encoder into their corresponding encoding vectors (1000-dimensions). A logistic regression model was then trained on 9K sentences to learn to predict voice and tense based on the English encoding vectors. They tested their logistic regression model on 1K sentences and achieved an accuracy of 92.8\% for voice predictions. Our work on discovering how much aspectual information is contained within an NMT system is inspired by this work as we also used a logistic regression model to predict `aspect' based on the encoding vectors. However, it also differs from their experiments in two ways:

\paragraph{(1)}First, a different `voice' or `tense' in a sentence manifest themselves `overtly' in the English source sentence, i.e., with a different verbal construction for passive and active voices and different verbal forms for all the English tenses. The passive will always be characterized by a +[be]-construction, such as in Example~(\ref{exMood}):

\begin{li}\label{exMood}
~\newline\newline
\begin{tabular}{lll}
 (a) & ``I taught French (to ...)'' & ACTIVE\\
 (b) & ``I was taught French (by ...)'' & PASSIVE
\end{tabular}
~\newline
\end{li}

This is not the case for aspect in English. The simple past in English is neutral\footnote{Not explicitly marked.} with respect to aspect but, as illustrated before in Example (\ref{lexAspect2}) and (\ref{exAspect}), contextual information (e.g. adverbs, adverbial phrases) can carry aspectual meaning. Determining the aspect of a sentence, if at all possible,\footnote{Sometimes information from the broader context (other sentences/paragraphs) is needed in order to determine the aspect correctly.} is a more complex task: aspectual meaning can be conveyed by words with different parts of speech and can furthermore be combined together in complex ways to create aspectual meaning of a verbal expression. We therefore hypothesize that predicting the aspect of sentences based on their NMT encoding vectors is a more complex task.

\paragraph{(2)}Second, the work of Shi et al.~\citeyearpar{Shi2016} shows that encoding vectors capture certain linguistic information. However, their study does not include any information on the actual effect of this on the translation. Is this information also `decoded' correctly?

In the next Section~\ref{sec:LRM}, we explain how we trained a logistic regression model in order to predict the aspectual value based on encoding vectors. The results and our analysis is presented in Section~\ref{sec:resultsaspect}.

\subsubsection{Logistic Regression Model}\label{sec:LRM}

In order to train our logistic regression model on the NMT encoding vectors, we need labeled English data. Since our NMT model is specifically trained to translate from English into French or Spanish, and French and Spanish require the past tense to make a distinction between two different past tenses that are each associated with an opposite aspectual value, we semi-automatically labeled the English sentences based on the aspectual value of the tense in the French/Spanish reference translations. As explained in Section \ref{extract}, we did not only extract information about the preference of specific verbs for the one aspectual tense or the other but also the specific translations of the verbs themselves. Since our NMT system is trained on 3M OpenSubtitles sentence pairs, we trained a PB-SMT system on the same data and extracted the possible imperfective (\emph{imparfait} and \emph{pret\'{e}rito imperfecto}) and perfective (\emph{pass\'{e} compos\'{e}} and \emph{pret\'{e}rito indefinido}) translations of our verbs. We used a separate set of the OpenSubtitles corpus to extract sentences with (i) verbs in the English simple past, and (ii) French or Spanish reference sentence containing an \emph{imparfait}/\emph{pret\'{e}rito imperfecto} or \emph{pass\'{e} compos\'{e}}/\emph{pret\'{e}rito indefinido} verb. Based on the appearance of either an imperfective tense or a perfective tense in the reference translations, we automatically labeled the corresponding English sentences and limited the length of the sentence to 10 tokens.\footnote{We limited the sentence length since we wanted only one label per sentence (i.e. one main verb) in order to be able to train a logistic regression model on the sentence vectors and the imperfective/perfective labels.} 

We randomly selected 40K labeled sentences, and generated for every sentence their encoding vector with the NMT system described above. Next, we trained a logistic regression using the python machine learning toolkit \verb|scikit learn|\footnote{\url{http://scikit-learn.org/stable/}} with the default settings.

\section{Results}\label{sec:resultsaspect}
\subsection{Logistic Regression Model}
To test our logistic regression model we compiled 4 test sets of increasing difficulty. Each of the test sets contains 2K sentences. The reason why we compiled 4 different test sets is because of the results we obtained and described in Section \ref{extract}. Our results showed  that some verbs have a very strong basic lexical aspect which links them to a particular tense in French (`surprised', `jumped' (IMP: 0\% and PC: 100\%) or `weighed', `sounded' (IMP: 100\% vs PC: 0\%)). Other verbs can easily be associated with an imperfective or perfective aspectual value and thus rely more on other factors apart from their lexical aspect in order to disambiguate between the two tenses (`reigned' (67\% vs 33\%), `lived' (44\% vs 56\%)). Since we assumed that especially those verbs that do not have a strong lexical aspect (and thus no strong preference for a particular tense) are harder to translate, we created 4 test sets. The first test set is the `general' one that contains all types of verbs, while the second test set  does not include verbs that have more than 80\% or less than 20\% preference for a particular tense. It thus only contains verbs whose preference for either tense is between 20\%-80\%. Similarly, test set 3 and test set 4 contain verbs with a 30\%-70\% and 40\%-60\% preference, respectively, i.e. the more `ambiguous' verbs in terms of aspect. We compared the predictions of our logistic regression model with the reference labels for the 4 test sets. To check our logistic regression model, we also computed a naive baseline performance, which represents the highest accuracy that would be obtained if all predictions consisted of only either 0s and 1s.\footnote{The baseline represents the ratio of perfective to imperfective labels.} The results of the logistic regression model for French and Spanish can be found in Table~\ref{resultsLogReg}.

\begin{table}[ht]
\centering
\begin{tabular}{|c|c|c|c|c|c|}\hline
     &  & \textbf{100--0} & \textbf{80--20} & \textbf{70--30} & \textbf{60--40}\\\hline
French  & \textbf{LogReg} & 90.95$\%$ & 86.10$\%$ & 86.20$\%$ & 77.10$\%$\\\cline{2-6}
        & \textbf{Baseline} & 64.55$\%$ & 64.60$\%$ & 74.55$\%$ & 72.60$\%$\\\hline\hline
Spanish & \textbf{LogReg} & 87.05$\%$ & 73.95$\%$ & 65.10$\%$ & 66.10$\%$\\\cline{2-6}
        & \textbf{Baseline} & 76.70$\%$ & 52.40$\%$ & 60.65$\%$ & 60.70$\%$\\\hline
\end{tabular}
\caption{Prediction accuracy of the Logistic Regression Model on the French and Spanish vectors compared to an accuracy baseline.}\label{resultsLogReg}
\end{table}

The results in Table \ref{resultsLogReg} confirm our hypothesis that the more ambiguous verbs in terms of aspect are harder to predict for the logistic regression model since the prediction accuracy lowers over the test sets. In the general test set (referred to as ``100--0'' in Table \ref{resultsLogReg}) the accuracy of the logistic regression model is 90.95\% for French. The accuracy drops when excluding verbs with a strong lexical aspect (test set ``80--20'' and ``70--30'') to 86.10\% and 86.20\%, respectively. The lowest score prediction accuracy is 77.10\% for the fourth test set (60--40), containing verbs that, without any further context, are almost equally likely to be translated into either of both French past tenses. 

For Spanish, we observe a similar trend. The logistic regression model's  accuracy drops considerably when comparing the most general test set (``100--0'') with the other test sets. 

The fact that a logistic model can extract certain aspectual information from the NMT encoding vectors does not guarantee that the decoder is able to. Therefore, in the next section we will examine in more detail the actual outputs of our NMT/PB-SMT systems on the same 4 test sets.

\subsection{Aspect in NMT/PB-SMT Translations}

A logistic regression model trained to label English source vectors with a particular tense achieved an accuracy of 90.95\% (English--French) and 87.05\% (English--Spanish), which are promising results. So far, however, we have not yet looked at the actual outputs of both systems. Accordingly, we now examine how the PB-SMT and NMT translations compare (in terms of selecting an imperfective or perfective French/Spanish tense) with respect to the reference translations.

We translated the 4 test sets described in Section \ref{sec:resultsaspect} with NMT and PB-SMT systems trained on the same 3M OpenSubtitles sentences. In Section \ref{extract} we explained that, together with the \emph{perfective} and \emph{imperfective} preference of verbs, we also extracted all the translations stored in the PB-SMT phrase-tables. As Example~(\ref{ex:translations}) in Section \ref{extract} shows, these translations do not contain all possible correct translations (often only the third person of verbs is represented, e.g. `felt': `craignaient', `estimaient', `jugeait' etc.). Therefore, we first of all made sure our translations included all possible forms (in terms of persons) of the translations extracted. We also included the `female' and `plural' forms of the French participle. One such translation expansion is partially shown in Example~(\ref{ex:translation_expansion}):

\begin{li}\label{ex:translation_expansion}
 \item \textbf{Verb: }``felt'' 
 \item \textbf{Pass\'{e} Comp.: } ``ai compris''; ``ai comprise''; ``ai comprises''; ``as compris''; ``as comprise''; ``as comprises''; ``\emph{a compris}''; ``a comprise''; ``a comprises''; ``avons compris''; ``avons comprise''; ``avons comprises''; ``avez compris''; ``avez comprise''; ``avez comprises''; ``ont compris''; ``ont comprise''; ``ont comprises''; ``\emph{ai consid{e}r\'{e}}''; ``ai consid{e}r\'{e}e''; ``ai consid{e}r\'{e}es''; ``as consid{e}r\'{e}''; ... 
 \item \textbf{Imparfait: } ``craignais''; ``craignait'', ``craignions''; ``craigniez''; ``\emph{craignaient}''; ``estimais''; ``\emph{estimait}''; ``estimions'', ``estimiez''; ``\emph{estimaient}''; ...
\end{li}

Based on the verbs in the source sentence and their translations, we were able to automatically evaluate the outputs of our translation systems. The results of our translated test sets for French are presented in Table~\ref{tbl:translationResultsFR} and for Spanish in Table~\ref{tbl:translationResultsES}. As we did with the logistic regression model, we observe that also for the translations, our test sets present different difficulty levels. Both NMT and PB-SMT perform best on the data containing all types of verbs (test set 1 ``100--0''). Performance decreases over the other three test sets with +/- 12\% for both NMT and PB-SMT. The logistic regression model indicated we can accurately (90.95\%) predict the correct grammatical aspect for the French tense in the general test set based on the vector encodings. We do not see this same accuracy reflected in the actual translations of the general test set (79\%). This implies that part of the aspectual information is lost during decoding. 

~\newline
\begin{table}[ht]
\centering
\setlength\tabcolsep{3pt}
\begin{tabular}{|l|l|c|c|c|c|c|c|c|c|}\hline
 & & \multicolumn{2}{c|}{100--0} &  \multicolumn{2}{c|}{80--20} &  \multicolumn{2}{c|}{70--30} &  \multicolumn{2}{c|}{60--40} \\\cline{3-10}
 Ref. & Trans. & SMT & NMT & SMT & NMT & SMT & NMT & SMT & NMT\\\hline
IMP & PC & 114 & 103 & 214 & 221 & 246 & 229 & 248 & 170\\\hline
PC & IMP & 26 & 40 & 72 & 93 & 86 & 103 & 93 & 171\\\hline
\multicolumn{2}{|c|}{Correct (in $\%$)} & 77.90$\%$ & 79.05$\%$ & 72.50$\%$ & 71.55$\%$ & 72.05$\%$ & 70.85$\%$ & 65.55$\%$ & 67.20$\%$\\\hline
\end{tabular}
~\newline
\caption{Translation accuracy PB-SMT vs NMT for the OpenSubtitles test sets for the English-French language pair.}\label{tbl:translationResultsFR}
\end{table}

Surprisingly, the performance of PB-SMT and NMT is very comparable on all test sets although their knowledge sources are different. This indicates that the lexical aspect of a verb plays an important role when selecting the tense with the correct grammatical aspect. This statement is consistent with the observations in Ye et al.~\citeyearpar{Ye2007} and Olsen et al.~\citeyearpar{Olsen2001}. Ye et al.~\citeyearpar{Ye2005,Ye2006,Ye2007} reported on the high utility of lexical aspectual features in selecting a tense. Similarly, Olsen et al.~\citeyearpar{Olsen2001} reported on the significance of the telicity of verbs in order to reconstruct the tense for Chinese-to-English translation. 

For English--Spanish (Table \ref{tbl:translationResultsES}), we obtained similar results. However, the results are overall lower than the ones obtained for the English--French systems. The tenses in the translations only correspond to the tense of the reference translations in 46.50\% of cases for PB-SMT,~\footnote{\label{perc_random} Since the perfective tense appears more in the test sets than the imperfective one and PB-SMT regularly opts for an imperfective tense (while the reference is perfective), percentages can be below 50$\%$.} and 57.70\% for NMT on the general test set (``100--0''). We analyzed some of the translations in order to identify why the results are lower than for the English--French systems and saw that often, our NMT and PB-SMT systems opted for another Spanish tense: the \emph{pret\'{e}rito perfecto compuesto}.\footnote{The \emph{pret\'{e}rito perfecto compuesto} is formally similar to the English \emph{present perfect}. It is conjugated by using the verb `haber' (have) followed by a past participle.} In the future, we would like to further extend our work in order to cover this additional tense. Unlike the English--French results, the English--Spanish NMT systems consistently outperform the PB-SMT systems in terms of selecting the same tense as the reference.

\begin{table}[ht]
\centering
\setlength\tabcolsep{3pt}
\begin{tabular}{|l|l|c|c|c|c|c|c|c|c|}\hline
 & & \multicolumn{2}{c|}{100--0} &  \multicolumn{2}{c|}{80--20} &  \multicolumn{2}{c|}{70--30} &  \multicolumn{2}{c|}{60--40} \\\cline{3-10}
Ref. & Trans. & SMT & NMT & SMT & NMT & SMT & NMT & SMT & NMT\\\hline
IMP & PC & 71 & 84 & 191 & 209 & 365 & 358 & 208 & 169\\\hline
PC & IMP & 98 & 84 & 225 & 178 & 251 & 229 & 361 & 348\\\hline
\multicolumn{2}{|c|}{Correct (in $\%$)} & 46.50$\%$ & 57.70$\%$ & 58.25$\%$ & 62.45$\%$ & 49.60$\%$ & 51.95$\%$ & 46.70$\%$ & 49.15$\%$\\\hline
\end{tabular}
~\newline
\caption{Translation accuracy PB-SMT vs NMT for the OpenSubtitles test sets for the English-Spanish language pair.}\label{tbl:translationResultsES}
\end{table}

For both English--French and English--Spanish language pairs, we see that the tense prediction of the logistic regression model is more accurate than the tenses in the NMT outputs. This is most likely due to the fact that the logistic regression model is trained for one specific task while the decoder has to take care of multiple tasks simultaneously such as word translations, word reordering, etc.

\section{Conclusions}\label{sec:conclusions}
In this chapter, we investigated what kind of aspectual information PB-SMT and NMT could grasp and how this is reflected in their translations. For PB-SMT, we saw the lexical aspect of verbs reflected in their `knowledge source', i.e. the phrase-tables. PB-SMT's knowledge is limited to the size of the \emph{n}-grams in the phrase-tables, so they cannot cover other aspectual factors that appear in a sentence (in case they fall out of the \emph{n}-gram range). We hypothesized this would be particularly problematic for those verbs that do not have a `strong' lexical aspect, which we saw confirmed by the results of the experiments conducted.

Unlike PB-SMT, NMT does have the means to store information about the entire source sentence. By using a logistic regression model trained on the encoding vectors, we discovered that NMT encoding vectors indeed capture aspectual information. Nevertheless, the evaluation of the actual outputs of the NMT and PB-SMT systems in terms of \emph{imperfective} or \emph{perfective} tense choice revealed that NMT and PB-SMT perform very similarly on all test sets. Although aspect can accurately (90.95\%) be predicted from the encoding vectors by a logistic regression model, the NMT decoder loses some of this information during the decoding process.

With respect to the first research question (RQ1), we observed how PB-SMT phrase-tables enable us to learn to some extent about the prototypical aspectual values of verbs. However, this information is limited as it is `static'. It represents the probability of an English verb to be translated in one tense or another, without any additional knowledge. In practice, while translating, the prototypical aspectual value of a verb can change depending on the context it appears in. The limiting \emph{n}-gram context does not allow for such contextual changes. For NMT, the encoding vectors do store more global sentence information but loses its advantage during the decoding.

As NMT does have the potential to incorporate more of the necessary linguistic information into their so-called knowledge sources, directing the NMT system a bit more during the decoding process could help improve our systems. We hypothesize that providing the NMT system with additional linguistic information that would allow it to generalize better over the seen information could help the systems improve the actual decoded output they produce. In the next chapter, we will try to integrate additional linguistic information that we believe could help the NMT system learn and generalize better over the seen output.

\chapterbib

%% file: SuperTags.tex


\newpage
\epigraph{We must think things not words, or at least we must constantly translate our words into the facts for which they stand, if we are to keep to the real and the true.}{\textit{Oliver Wendell Holmes Jr. }}

\chapter{Integrating Semantic Supersenses and Syntactic Supertags into Neural Machine Translation Systems}\label{ch:Supertag}

In Chapter~\ref{ch:Agreement} and Chapter~\ref{ch:Aspect}, we compared NMT's and PB-SMT's performance based on two (automatic) translation difficulties: subject-verb number agreement and more complex tense- and aspect-related issues. Although NMT clearly outperforms PB-SMT for the former, the latter remains difficult even for a state-of-the-art NMT system. From our analysis, it appears that using a logistic regression classifier one could in most cases accurately predict the correct tense with the appropriate aspectual value from the encoding vector. Nevertheless, by analyzing the actual translations, we observed that some of that information is lost during the NMT decoding process. For that reason, we contend that NMT systems can still be improved by integrating additional linguistic information. Providing an NMT system with more specific and more general information might facilitate its handling of more complex linguistic patterns. Accordingly, in this chapter, we integrate simple and more complex syntactic features as well as, and in combination with, high-level semantic features. This answers the second part of RQ2 partially as we explore one way of integrating linguistic knowledge into NMT models by employing factored models. Another way of integrating linguistic features for NMT will be explored in Chapter~\ref{ch:Gender} while the integration of linguistic features for PB-SMT has already been discussed in Chapter~\ref{ch:Agreement}.

\section{Introduction}

Compared to PB-SMT, NMT performs particularly well when it comes to word reorderings and translations involving morphologically rich languages~\citep{Bentivogli2016}. Although NMT seems to partially `learn' or generalize some patterns related to syntax from the raw, sentence-aligned parallel data, more complex linguistics phenomena (e.g. prepositional-phrase attachment or tense and aspect) remain problematic~\citep{Bentivogli2016,vanmassenhove2017investigating}. Recent work showed that explicitly~\citep{Sennrich2016,Aharoni2017,Bastings2017,Nadejde2017} or implicitly~\citep{Eriguchi2017} modeling extra syntactic information into an NMT system on the source (and/or target) side could lead to improvements in translation quality. Sennrich and Haddow~\citeyearpar{Sennrich2016} integrated morphological information, POS-tags and dependency labels in the form of features on the source side of the NMT model, while Nadejde et al.~\citeyearpar{Nadejde2017} introduced syntactic information in the form of CCG supertags on both the source and the target side. Moreover, Nadejde et al.~\citeyearpar{Nadejde2017} showed that a shared embedding space, where syntax information and words are tightly coupled, is more effective than multitask training.

When integrating linguistic information into an MT system, following the central role assigned to syntax by many linguists, the focus has been mainly on the integration of syntactic features. Although there is a body of research on semantic features for PB-SMT~\citep{Wu2009,Liu2010,Aziz2011,baker2012,jones2012,Bazrafshan2013}, at the time our research was conducted, no work had been done on enriching NMT systems with more general semantic features at the word-level yet.\footnote{By the time of compiling this thesis (May 2019), there have been some other publications on leveraging semantics for NMT. This research will be discussed in Section~\ref{subsec:relwNMT}.} This might be explained by the fact that NMT models already have a means of learning semantic similarities through word-embeddings, where words are represented in a common vector space \citep{Mikolov2013}. However, making some level of semantics more explicitly available at the word or sentence level can provide the MT system with a higher level of abstraction beneficial to learn more complex constructions. Furthermore, we hypothesize that a combination of both syntactic and semantic features would help the NMT system learn more difficult semantico-syntactic patterns. 

To illustrate a more challenging semantico-syntactic pattern for MT, consider the translation presented in Example~(\ref{exmp:catmisPB-SMT}) originally used in \cite{jones2012} to demonstrate how a (back then) state-of-the-art German--English PB-SMT system is unable to preserve basic meaning structure:\footnote{\cite{jones2012} used Google Translate, which was back then a PB-SMT system. They generated this translation on 08/31/2012\citep{jones2012}.}

\begin{li}\label{exmp:catmisPB-SMT}
\item (a)  Source:   Anna fehlt ihrem Kater.
\item (b)  PB-SMT:     *Anna is missing her cat.
\item (c)  Ref:     Anna's cat is missing her.
\end{li}

Due to the different realization of arguments in English and German, the PB-SMT system fails in this instance. It keeps the same argument order when translating into English and thus loses the basic meaning structure. The French translation of the verb `to miss' (manquer \`a) has a similar argument structure as the German verb in Example~(\ref{exmp:catmisPB-SMT}). When translating an equivalent phrase from French to English with GNMT, the same problem occurs.\footnote{GNMT used on 06/05/2019.} This French-English translation is given in Example~(\ref{exmp:catmisFr}).

\begin{li}\label{exmp:catmisFr}
\item (a)  Source:   Anna manque à son chat.
\item (b)  GNMT:     *Anna misses her cat.
\item (c)  Ref:      Anna's cat is missing her.
\end{li}

Additionally, when translating the German sentence presented in Example~(\ref{exmp:catmisPB-SMT}) using GNMT~\footnote{GNMT used on 06/05/2019} we obtain the following translation (Example~(\ref{exmp:catmisNMT})):

\begin{li}\label{exmp:catmisNMT}
\item (a)  Source:   Anna fehlt ihrem Kater.
\item (b)  GNMT:     *Anna is missing her hangover.
\item (c)  Ref:     Anna's cat is missing her.
\end{li}

In Example~(\ref{exmp:catmisNMT}), we observe that GNMT does not only fail to preserve the basic meaning structure but also that it opts for a rather strange translation of the German word `Kater' in this particular context. The word `Kater' is ambiguous and can refer to either `a male cat' or `a hangover', so on a word-level `hangover' is indeed a correct translation. However, when opting for the second alternative, the correct translation should actually be `Anna's hangover is missing her', and as `hangover' is inanimate,\footnote{As language is flexible, one could imagine a very specific context where a `hangover' is somehow personified. However, we can clearly see how this translation is not the preferred one without such context, especially not given that the other possible translation is animate (`cat').} this is a highly unlikely translation. Examples~(\ref{exmp:catmisPB-SMT}),~(\ref{exmp:catmisFr}) and~(\ref{exmp:catmisNMT}) illustrate how both syntax and semantics are important, but also how both PB-SMT and NMT have difficulties dealing with preserving relatively basic meaning structures even in short sentences.

To apply semantic abstractions at the word-level that enable a characterisation beyond what can be superficially derived, coarse-grained semantic classes can be used. Inspired by Named Entity Recognition which provides such abstractions for a limited set of words, supersense-tagging uses an inventory of more general semantic classes for domain-independent settings~\citep{Schneider2015}. In this chapter, we investigate the effect of integrating supersense features (26 for nouns, 15 for verbs) into an NMT system. To obtain these features, we used the \emph{AMALGrAM 2.0} tool~\citep{Schneider2014,Schneider2015} which analyses the input sentence for Multi-Word Expressions (MWE) as well as noun and verb supersenses. The features are integrated using the framework of Sennrich et al.~\citeyearpar{Sennrich2016b}, replicating the tags for every subword unit obtained by BPE. We further experiment with a combination of semantic supersenses and syntactic supertag features (CCG syntactic categories~\citep{Steedman2000} using EasySRL~\citep{Lewis2015}) and less complex features such as POS-tags, assuming that the semantic supersense tags have the potential to be useful especially in combination with syntactic information.

The remainder of this chapter is structured as follows: first, in Section~\ref{sec:RelW}, the related work for PB-SMT and NMT is discussed. Next, Section~\ref{sec:SemAndSyn} presents the semantic and syntactic features used. The experimental set-up is described in Section~\ref{sec:Experiment} followed by the results in Section~\ref{sec:Results}. Finally, we present our main conclusions in Section~\ref{sec:ConcFut}.

\section{Related Work}\label{sec:RelW}
Section~\ref{subsec:relwPB-SMT} discusses the related work for PB-SMT. To the best of our knowledge, there had been no work on explicitly integrating semantic information in NMT at the time our experiments were conducted. Relevant work that was published simultaneously with, or after, our work was conducted is discussed in Section~\ref{subsec:relwNMT}.

\subsection{Statistical Machine Translation}\label{subsec:relwPB-SMT}
In PB-SMT, on the syntax level, various linguistic features such as stems~\citep{Toutanova2008}, lemmas~\citep{Marevcek2011,Fraser2012}, POS-tags~\citep{Avramidis2008}, dependency labels~\citep{Avramidis2008} and supertags~\citep{Hassan2007,Haque2009} are integrated using pre- or post-processing techniques often involving factored phrase-based models~\citep{Koehn2007}. Compared to factored NMT models, factored PB-SMT models have some disadvantages: (a) adding factors increases the sparsity of the models, (b) the \emph{n}-grams limit the size of context that is taken into account, and (c) features are assumed to be independent of each other. However, adding syntactic features to PB-SMT systems led to improvements with respect to word order and morphological agreement~\citep{Williams2012,Sennrich2015}. 

On the semantics level, \cite{Wu2009} were the first to use semantic parsing to improve PB-SMT models. They present a novel hybrid semantic PB-SMT model by using a two-pass architecture. For the first pass, they use a conventional PB-SMT model. The second pass consists of employing a shallow semantic parser that produces semantic frame and role labels for reordering. \cite{Liu2010} used semantic role features for a Tree-to-String transducer. While the previous works focused solely on the integration of semantic information, \cite{Aziz2011} integrated both shallow syntactic and semantic information for PB-SMT. However, their experiments showed that there was no improvement with respect to the model with shallow syntactic information. They attribute this to sparsity and representation issues as multiple predicates share arguments within a given sentence~\citep{Aziz2011}. \cite{baker2012} applied a new modality and negation annotation scheme to PB-SMT using a syntactic framework that allowed for enrichment with semantic annotations. A similar method was proposed in \cite{Bazrafshan2013}, where semantic information was added to the syntactic tree. However, rather than focusing on modality and negation, they worked with predicate-argument structure that represented the overall structure of each verb~\citep{Bazrafshan2013}. Finally, \cite{jones2012} used a graph-structured meaning representation to create a semantically-driven PB-SMT system.

\subsection{Neural Machine Translation}\label{subsec:relwNMT}


One of the main strengths of NMT is its strong ability to generalize. The integration of linguistic features can be handled in a flexible way without creating sparsity issues or limiting context information within the same sentence, both of which were issues for PB-SMT. Furthermore, the encoder and attention layers can be shared between features. By representing the encoder input as a combination of features~\citep{Alexandrescu2006}, Sennrich and Haddow \citeyearpar{Sennrich2016} generalized the embedding layer in such a way that an arbitrary number of linguistic features can be explicitly integrated. They then investigated whether features such as lemmas, subword tags, morphological features, POS-tags and dependency labels could be useful for NMT systems or whether their inclusion is redundant. 
Also focusing on the syntax level, Shi et al. \citeyearpar{Shi2016} show that although NMT systems are able to partially learn syntactic information, more complex patterns remain problematic. Furthermore, sometimes information is present in the encoding vectors but is lost during the decoding phase~\citep{vanmassenhove2017investigating}.

\cite{Sennrich2016b} show that the inclusion of linguistic features leads to improvements over the NMT baseline for English--German (0.6 BLEU), German--English (1.5 BLEU) and English--Romanian (1.0 BLEU). When evaluating the gains from the features individually, it appears that the gain from different features is not fully cumulative. Nadejde et al.~\citeyearpar{Nadejde2017} extend their work by including CCG supertags as explicit features in a factored NMT system. They also propose a novel approach where syntax from the target language is integrated at the word level in the decoder by interleaving CCG supertags in the target word sequence. They show that CCG supertags improve the translation quality (measured in terms of BLEU) for German--English and Romanian--English.
Moreover, they experiment with serializing and multitasking and show that tightly coupling the words with their syntactic features leads to improvements in translation quality as measured by BLEU, whereas a multitask approach (where the NMT predicts CCG supertags and words independently) does not perform better than the baseline system. A similar observation was made by Li et al \citeyearpar{Li2017}, who incorporate the linearized parse trees of the source sentences into Chinese--English NMT systems. They propose three different sorts of encoders: (a) a parallel RNN, (b) a hierarchical RNN, and (c) a mixed RNN. Like Nadejde et al.~\citeyearpar{Nadejde2017}, Li et al~\citeyearpar{Li2017} observe that the mixed RNN (the simplest RNN encoder), where words and label annotation vectors are simply stitched together in the input sequences, yields the best performance with a significant BLEU improvement ($+$1.4 BLEU, a relative improvement of 4\%). 

Eriguchi et al.~\citeyearpar{Eriguchi2016} integrated syntactic information in the form of linearized parse trees by using an encoder that computes vector representations for each phrase in the source tree. They focus on source-side syntactic information based on Head-Driven Phrase Structure Grammar~\citep{Sag1999} where target words are aligned not only with the corresponding source words but with the entire source phrase. 
Compared to Sennrich et al. \citeyearpar{Sennrich2016} and Nadejde et al. \citeyearpar{Nadejde2017}, they focus more on exploiting the unlabelled structure of syntactic annotation and less on the disambiguation power of functional dependency labels. The approach of Eriguchi et al. \citeyearpar{Eriguchi2016} is effective for handling one-to-many alignments but cannot handle long-distance dependencies.
Wu et al.~\citeyearpar{Wu2017} focus on incorporating source-side long-distance dependencies by enriching each source state with global dependency structure.

Similarly to syntactic features, we hypothesize that semantic features in the form of semantic `classes' can be beneficial for NMT by providing it with an extra ability to generalize and thus better learn more complex semantico-syntactic patterns (specifically when combined with other syntactic features). At the time our experiments were conducted and published, there was, to the best of our knowledge, no other work on integrating semantic structures into NMT.

Independent from our research but published at the same time and event,\footnote{ACL 2018, Melbourne, Australia.}~\cite{Marcheggiani2018} experimented with the integration of semantic features into an NMT system for English--German. Their work shows how the integration of predicate argument structure of the source sentences into a standard attention-based NMT model~\citep{Bahdanau2014} is beneficial for the English--German language pairs. They observe better results with semantic features ($+$1.1 BLEU or a 4.7\% relative improvement) compared to syntactic ones ($+$0.6 BLEU or a 2.6\% relative improvement) and obtain a further gain ($+$1.6 BLEU points or a 6.9\% relative improvement compared to their baseline) when combining them, a conclusion similar to our findings.

Other research on incorporating semantics in NMT that was published after our experiments includes the work of \cite{Song2019} where the usefulness of Abstract Meaning Representation for NMT is examined. They show that a  significant improvement can be achieved over an attention-based sequence-to-sequence NMT baseline. Another set of experiments that is less directly related to our work but still worth mentioning is a paper by \cite{Shah2018} on `generative' NMT. Their work is based on the idea that a sentence's real meaning can be captured by looking at that same sentence in multiple languages. Unlike other work on NMT, their model is designed to learn the \emph{joint} distribution of the target and the source. To achieve this, they use a latent variable (that represents the meaning in a language-agnostic way) to generate the same sentence in multiple languages. They argue that, this way, the latent variable is encouraged to capture the semantic meaning of the sentence. They show that their method is particularly effective on longer sentences and achieves comparable BLEU scores to the more standard NMT models that model a \emph{conditional} distribution of the target sentence given the source.

\section{Semantics and Syntax in Neural Machine Translation}\label{sec:SemAndSyn}
We present in detail the semantic (Section~\ref{sec:sem}) and syntactic (Section~\ref{sec:syn}) features that we integrated in an NMT system.

\subsection{Supersense Tags}\label{sec:sem}
The novelty of our work is the integration of explicit semantic features \emph{supersenses} into an NMT system. Supersenses are a term which refers to the top-level hypernyms in the WordNet~\citep{Miller1995} taxonomy, sometimes also referred to as \emph{semantic fields} \citep{Schneider2015}. The supersenses cover all nouns and verbs with a total of 41 supersense categories~\citep{Schneider2015}, 26 for nouns and 15\footnote{16 if you count the separate tag for auxiliary verbs.} for verbs (see~\ref{exmp:supersenses}):

\begin{li}\label{exmp:supersenses}
\item  Nouns:\\  ACT, ANIMAL, ARTIFACT, ATTRIBUTE, BODY, COGNITION, COMMUNICATION, EVENT, FEELING, FOOD, GROUP, LOCATION, NATURAL OBJECT, PERSON, PHENOMENON, POSSESSION, QUANTITY, RELATION, STATE, SUBSTANCE, TIME, MOTIVE, PROCESS, PLANT, OTHER, SHAPE
\item  Verbs:\\  `a,\footnote{Used for auxiliary verbs.} body, change, cognition, communication, competition, consumption, contact, creation, emotion, motion, perception, possession, social, stative, weather
\end{li}

To obtain the supersense tags we used the \emph{AMALGrAM (A Machine Analyzer of Lexical Groupings and Meanings) 2.0} tool\footnote{\url{https://github.com/nschneid/pysupersensetagger}} which in addition to the noun and verb supersenses analyzes English input sentences for MWEs. \cite{Schneider2015} argue it is important to treat MWEs as a unit when providing supersense tags because of their semantically holistic nature.\footnote{The type of MWEs considered by the AMALGrAM tool include idioms, light verb constructions, verb-particle constructions and many compounds~\citep{Baldwin2010}. A more complete study on the role of MWEs in MT can be found in~\cite{Monti2018}.} An example of a sentence annotated with the AMALGrAM tool is given in (\ref{exmp:sst}):\footnote{All the examples are extracted from our data used later on to train the NMT systems.}

\begin{li}\label{exmp:sst}
\item (a)   ``He seemed to have little faith in our democratic structures, suggesting that various articles could be misused by governments.''\\
\item (b)   ``He \textbf{seemed$|$cognition} to have$|$stative little \textbf{faith$|$COGNITION} in our democratic structures$|$ARTIFACT , suggesting$|$communication that various articles$|$COMMUNICATION could be$|$`a misused$|$social by\\ governments$|$GROUP .''
\end{li}

As can be noted in (\ref{exmp:sst}), some supersenses (such as \emph{cognition}) exist for both nouns and verbs. However, the supersense tags for verbs are always lowercased while the ones for nouns are capitalized. This way, the supersenses also provide syntactic information useful for disambiguation as in (\ref{exmp:sst_disambig}), where the ambiguous word \emph{work} is correctly tagged as a noun (with its capitalized supersense tag \emph{ACT}) in the first part of the sentence and as a verb (with the lowercased supersense tag \emph{social}). Similarly, in Example~(\ref{exmp:sst}), the verb `seemed' is tagged with a lowercased supersense \emph{cognition}, while the noun `faith' received the uppercased tag \emph{COGNITION}. 

\begin{li}\label{exmp:sst_disambig}
\item Input:   ``In the course of my work on the opinion, I in fact became aware of quite a number of problems and difficulties for EU citizens who live and work in Switzerland.''\\
\item Tagged:  ``In the course$|$EVENT of my work~$|$ACT on the opinion~$|$COGNITION , I \textbf{in\textunderscore fact} became$|$stative aware of quite \textbf{a\textunderscore number\textunderscore of} problems~$|$COGNITION and difficulties~$|$COGNITION for \textbf{EU\textunderscore citizens$|$GROUP} who live$|$social and work$|$social in Switzerland$|$LOCATION .''\\
\end{li}

As the factored NMT input requires a tag for every word, we add a \emph{none} tag to all words that did not receive a specific tag. The final version of the original sentence shown in Example~(\ref{exmp:sst_disambig}), padded with \emph{none} tags, is given in Example~(\ref{exmp:nonepad}):

\begin{li}\label{exmp:nonepad}
\item Final:  ``in$|$none the$|$none course$|$EVENT of$|$none my$|$none \textbf{work$|$ACT} on$|$none the$|$none opinion$|$COGNITION ,$|$none I$|$none in$|$mew fact$|$mew \textbf{became$|$stative} aware$|$none of$|$none quite$|$none \textbf{a$|$mew number$|$mew of$|$mew} problems$|$COGNITION and$|$none difficulties$|$COGNITION for$|$none EU$|$GROUP citizens$|$GROUP who$|$none live$|$social and$|$none \textbf{work$|$social} in$|$none Switzerland$|$LOCATION .$|$none''
\end{li}



Since the semantically holistic nature of MWEs and supersenses naturally complement each other, Schneider and Smith \citeyearpar{Schneider2015} integrated the MWE identification task \citep{Schneider2014} with the supersense tagging task of Ciaramita and Altun~(\citeyear{Ciaramita2006}). In Example (\ref{exmp:sst_disambig}), the MWEs \emph{in fact}, \emph{a number of} and \emph{EU citizens} are retrieved by the tagger. 

We add this semantico-syntactic information in the source as an extra feature in the embedding layer following the approach of Sennrich and Haddow~\citeyearpar{Sennrich2016}, who extended the model of Bahdanau et al.~(\citeyear{Bahdanau2014}). A separate embedding is learned for every source-side feature provided (the word itself, POS-tag, supersense tag etc.). These embedding vectors are then concatenated into one embedding vector and used in the model instead of the simple word-embedding one~\citep{Sennrich2016}. 

To reduce the number of OOV words, we follow the approach of Sennrich et al.~(\citeyear{Sennrich2016b}) using a variant of BPE for word segmentation capable of encoding open vocabularies with a compact symbol vocabulary of variable-length subword units. For each word that is split into subword units, we copy the features of the word in question to its subword units. In~(\ref{exmp:bpe}), we give an example with the word `stormtroopers' that is tagged with the supersense tag `GROUP'. It is split into five subword units, so the supersense tag feature is copied to all of its five subword units. 

\begin{li}\label{exmp:bpe}
\item Input:   ``the stormtroopers''\\
SST:   ``the stormtroopers$|$GROUP''\\
BPE:   ``the stor@@ m@@ tro@@ op@@ ers''\\
Output:   ``the$|$\textbf{none} stor@@$|$\textbf{GROUP} ... op@@$|$\textbf{GROUP} ers$|$\textbf{GROUP}''
\end{li}

For the MWEs we decided to copy the supersense tag to all the words of the MWE (if provided by the tagger), as in~(\ref{exmp:mwe_sst}). Transferring the tags to the relevant units of the MWE was done automatically by leveraging the MWE indicators provided by the AMALGrAM tool in combination with hand-written rules. If the MWE did not receive a particular tag, we added the tag \emph{mwe} to all its components, as in Example~(\ref{exmp:mwe_nosst}).

\begin{li}\label{exmp:mwe_sst}
\item Input:   ``EU citizens''\\
SST:   ``EU\textbf{\textunderscore} citizens$|$GROUP''\\
Output:   ``EU$|$GROUP citizens$|$GROUP''
\end{li}

\begin{li}\label{exmp:mwe_nosst}
\item Input:   ``a number of''\\
SST:   ``a\textbf{\textunderscore} number \textbf{\textunderscore} of''\\
Output:   ``a$|$\textbf{mwe} number$|$\textbf{mwe} of$|$\textbf{mwe} ''
\end{li}


\subsection{Supertags and POS-tags}\label{sec:syn}
We hypothesize that more general semantic information can be particularly useful for NMT in combination with more detailed syntactic information. To support our hypothesis we also experimented with syntactic features (separately and in combination with the semantic ones): POS-tags and CCG supertags. 

The POS-tags are generated by the Stanford POS-tagger~\citep{Toutanova2003}. For the supertags we used the EasySRL tool~\citep{Lewis2015} which annotates words with CCG-tags. CCG-tags provide global syntactic information on the lexical level. This kind of information can help resolve ambiguity in terms of prepositional attachment, among others. An example of a CCG-tagged sentence is given in~(\ref{exmp:ccg}):\\

\begin{li}\label{exmp:ccg}
\item  It$|$NP is$|$(S[dcl]\textbackslash NP)/NP a$|$NP/N modern$|$N/N form$|$N/PP of$|$PP/NP \\colonialism$|$N .$|$.
\end{li}

\section{Experiments}\label{sec:Experiment}
This section describes the data used in our experiments to train the NMT models as well as the different settings for the NMT systems.

\subsection{Data sets}
Our NMT systems are trained on 1M parallel sentences of the Europarl corpus for English--French (EN--FR) and English--German (EN--DE)~\citep{Koehn2005}. We evaluate the systems on 5K sentences extracted from Europarl and the newstest2013. Two different test sets are used in order to show to what extent additional semantic and syntactic features can help the NMT system translate different types of data. We hypothesize that providing supersenses is particularly useful when testing on a domain that differs from the training data.

\subsection{Description of the Neural Machine Translation System}
We used the \verb|nematus| toolkit \citep{Sennrich2017} to train encoder-decoder NMT models with the following parameters: \emph{vocabulary size:} $35 000$, \emph{maximum sentence length:} $60$, \emph{vector dimension:} $1 024$, \emph{word embedding layer:} $700$, \emph{learning optimizer:} \verb|adadelta|.

We keep the embedding layer fixed to 700 for all models in order to ensure that the improvements are not simply due to an increase in the parameters in the embedding layer. As such, rather than giving an advantage to our linguistically-enriched system, we are `sacrificing' part of the word-embedding vector space to integrate the additional linguistic information. In order to bypass the OOV problem and reduce the number of dictionary entries, we use word segmentation with BPE. We ran the BPE algorithm with $89,500$ operations. We trained all our BPE-ed NMT systems with CCG-tag features, supersense tags (SST), POS-tags and the combination of syntactic features (POS or CCG) with the semantic ones (SST). All systems are trained for $150,000$ iterations and evaluated after every $10,000$ iterations.

As~\cite{Sennrich2016} use a subword structure similar to the IOB format, consisting of four symbols (IOB and E), we experiment with and without them. While `O' is used when a symbol corresponds to a `complete' word, `B' marks the begininng of a word, `I' the inside and `E' the end. An example of this format is illustrated in~(\ref{exmp:IOBE}), where the first word `histrionics' is split into 4 `subwords'\footnote{As can be seen, such subwords are statistically motivated, rather than linguistically.} and tagged accordingly with the IOBE format. 

\begin{li}\label{exmp:IOBE}
\item Input:  ``Histrionics do not help''
\item BPE :  ``His@@ tri@@ on@@ ics do not help''
\item + IOBE:  ``His@@$|$B tri@@$|$I on@@$|$I ics$|$E do$|$O not$|$O help$|$O''
\item + POS:  ``His@@$|$B$|$NNS tri@@$|$I$|$NNS on@@$|$I$|$NNS ics$|$E$|$NNS do$|$O$|$VBP not$|$O$|$RB help$|$O$|$VB''
\end{li}

\section{Results}\label{sec:Results}
In this section, we discuss the results obtained for the EN--FR and EN--DE NMT systems.

\subsection{English--French}
First of all, we present the result obtained on the in-domain Europarl dataset. As we are interested in the effect of the features on the learning process of the NMT system, we present its intermediate results for the $150,000$ training iterations. In Table~\ref{tbl:scoresSingleFeatBIO}, the BLEU scores can be found for the baseline system (BASE) as well as for single features (IOBE, POS, SST and CCG), and the combination of syntactic features and semantic supersenses (POS+SST and CCG+SST). As can be seen, the system that combines POS-tags and supersense features (POS+SST) is the one that most often obtains the highest BLEU score. POS-tags (POS) and supersenses (SST) also appear to be useful single features.

\begin{table}[ht!]
\centering
\begin{tabular}{|l|c|c|c|c|c|c|c|}
\hline
\bf{\#Iter}	    &\bf{BASE}	&\bf{IOBE}	    &\bf{POS}	    &\bf{SST}       &\bf{CCG}       &\bf{POS+SST}   &\bf{CCG+SST}	    \\ \hline
\bf{10k}	    &	11.91	&	11.3    	&	10.14	    &	11.99 	    &	8.49	    &   \bf{13.8} 	&	7.03    	    \\ \hline
\bf{20k}	    &	31.61	&	29.32   	&	31.88	    &	32.24	    &	30.37       &   \bf{33.21} 	&	30.96    	    \\ \hline
\bf{30k}	    &	31.87	&	36.97   	&	37.84	    &	38.43       &	38.42       &   \bf{38.52} 	&	37.78    	    \\ \hline
\bf{40k}	    &	36.84	&	40.44   	&	41.00       &	40.29	    &	40.82       &   \bf{41.21} 	&	40.97    	    \\ \hline
\bf{50k}	    &	40.10	&	41.47   	&	41.81	    &  	\bf{42.14}	&	41.99       &   41.51       &	42.27    	    \\ \hline
\bf{60k}	    &	41.69	&	\bf{43.43}	&	43.11	    &	43.42	    &	42.91       &   43.26 	    &	43.38    	    \\ \hline
\bf{70k}	    &	42.81	&	43.58   	&	43.75       &	43.39	    &	42.75	    &   \bf{43.76} 	&	43.72    	    \\ \hline
\bf{80k}	    &	44.18	&	44.26   	&	\bf{44.53}	&	44.30	    &	44.23	    &   43.92 	    &	44.01    	    \\ \hline
\bf{90k}	    &	44.3	&	44.4    	&	44.84   	&	44.80	    &	\bf{44.94}	&   44.92 	    &	44.67    	    \\ \hline
\bf{100k}	    &	44.26	&	44.88   	&	44.94	    &	44.70	    &	45.03	    &   \bf{45.16} 	&	45.13    	    \\ \hline
\bf{110k}	    &	45.04	&	\bf{45.49} 	&	44.77	    &	45.12	    &	44.9	    &   44.94 	    &	45.44    	    \\ \hline
\bf{120k}	    &	45.14	&	45.20   	&	\bf{45.52}  &	44.93	    &	44.74	    &   45.36 	    &	44.99    	    \\ \hline
\bf{130k}	    &	45.37	&	45.46   	&	\bf{45.77}  &	45.32	    &	45.34	    &   45.41 	    &	45.36    	    \\ \hline
\bf{140k}	    &	45.51	&	45.78   	&	45.41	    &	45.78	    &	45.42       &   \bf{46.1} 	&	45.66    	    \\ \hline
\bf{150k}	    &	45.56	&	45.34   	&	45.39	    &	\bf{45.79}	&	45.09       &   45.49 	    &	45.64    	    \\ \hline

\end{tabular}
\caption{BLEU scores for the EN--FR data over the 150k training iterations for the baseline system (BASE) and single features (EBOI, POS, SST and CCG) as well as two combinations of syntactic and semantic features (POS+SST and CCG+SST) evaluated on the in-domain Europarl set.}\label{tbl:scoresSingleFeatBIO}
\end{table}

Additionally, it is clear how the linguistically-enriched systems (whether this being with POS-tags, SST-tags, CCG-tags or the combinations thereof), lead to very big improvements in the beginning stages of the training process. For instance, when looking at the $3^{rd}$ iteration ($30k$ in Table~\ref{tbl:scoresSingleFeatBIO}), the baseline (BASE) obtains a BLEU score of $31.87$, while all of the systems with linguistic features achieve scores above $37$ BLEU (up to $38.52$, a 20.9\% relative improvement for the NMT system enriched with POS- and SST-tags). The difference between the BASE system and the enriched systems gradually disappears as the training process continues. Still, the system obtaining the highest overall BLEU score is the $POS+SST$ system with $46.1$ BLEU (a $+0.54$ absolute improvement and a $1.2$\% relative improvement over the highest baseline system).

Similar experiments were conducted on the newstestset2013, an out-of-domain dataset.\footnote{News data is maybe not always considered to be very different from Europarl, but in this case we are comparing how the system performs on Europarl data (an exact match with the training data) and News data. Therefore, in this scenario, we consider the News data as out-of-domain data.} 

\begin{table}[h!]
\centering
\begin{tabular}{|l|c|c|c|c|c|c|c|}
\hline
\bf{\#Iter}		&\bf{BASE}		&\bf{EBOI}	    &\bf{POS}	    &\bf{CCG}	    &\bf{SST}		&\bf{SST+POS}	&\bf{SST+CCG}   \\ \hline
\bf{10k}		&\bf{3.47}		&2.43		    &2.35		    &2.74		    &1.81		    &2.35		    &2.39		    \\ \hline
\bf{20k}		&11.11		    &10.04		    &\bf{13.18}		&11.11		    &10.25		    &\bf{13.18}		&8.09		    \\ \hline
\bf{30k}		&11.95		    &14.46		    &\bf{15.8}	    &14.44		    &14.56		    &\bf{15.8}	    &15.83		    \\ \hline
\bf{40k}		&13.33		    &17.14		    &17.51		    &17.65  	    &16.23		    &17.51		    &\bf{17.75}	    \\ \hline
\bf{50k}		&16.29		    &17.58		    &\bf{19.17}		&18.56		    &18.29		    &\bf{19.17}	    &18.64		    \\ \hline
\bf{60k}		&18.63		    &19.07		    &19.51		    &19.5		    &19.26		    &\bf{20.01}	    &19.91		    \\ \hline
\bf{70k}		&19.32		    &19.17		    &20.19		    &20.14		    &19.74		    &20.49		    &\bf{20.52}	    \\ \hline
\bf{80k}		&20.51		    &\bf{21.11}		&21.03  		&20.27		    &20.66		    &20.92		    &20.95		    \\ \hline
\bf{90k}		&20.91		    &21.11		    &21.11		    &\bf{21.56}     &20.94		    &21.27		    &20.82		    \\ \hline
\bf{100k}		&20.11		    &21.20		    &21.70		    &\bf{21.73}		&21.02		    &21.56		    &20.88		    \\ \hline
\bf{110k}		&20.97		    &21.17		    &22.05		    &22.53		    &22.09		    &\bf{22.54}		&22.18		    \\ \hline
\bf{120k}		&21.88		    &21.09		    &22.09		    &\bf{22.48}		&21.10		    &22.34		    &22.27		    \\ \hline
\bf{130k}		&21.96		    &22.43		    &22.63		    &22.00		    &21.81		    &22.28		    &\bf{22.76}	    \\ \hline
\bf{140k}		&22.27		    &22.60		    &\bf{22.93}		&22.20		    &22.38		    &22.81	        &22.74		    \\ \hline
\bf{150k}		&22.45		    &22.87		    &22.99		    &22.50		    &22.83		    &22.78		    &\bf{23.00}	    \\ \hline

\end{tabular}
\caption{BLEU scores for the EN--FR data over the 150k training iterations for the baseline system (BASE) and single features (EBOI, POS, SST and CCG) as well as two combinations of syntactic and semantic features (POS+SST and CCG+SST) evaluated on the out-of-domain News set}\label{tbl:scoresSingleFeatBIONEWS}
\end{table}

First of all, from Table~\ref{tbl:scoresSingleFeatBIONEWS} we can see that the BLEU scores are a lot lower compared to the ones obtained on the in-domain data (see Table~\ref{tbl:scoresSingleFeatBIO}). Secondly, the semantic and syntactic feature combinations seem more useful in the out-of-domain scenario as hypothesized. Like in the previous Table (Table~\ref{tbl:scoresSingleFeatBIO}), supersenses in combination with POS-tags (SST+POS) are a useful feature combination, but unlike in Table~\ref{tbl:scoresSingleFeatBIO}, this time also the SST+CCG often leads to good BLEU scores compared to the other features and the baseline system (BASE).

For both test sets, the NMT system with supersenses (SST) converges faster than the baseline (BPE) NMT system. As we hypothesized, the benefits of the features added were more apparent on the newstest2013 than on the Europarl test set.

\begin{figure}
 \centering
 \includegraphics[scale=1]{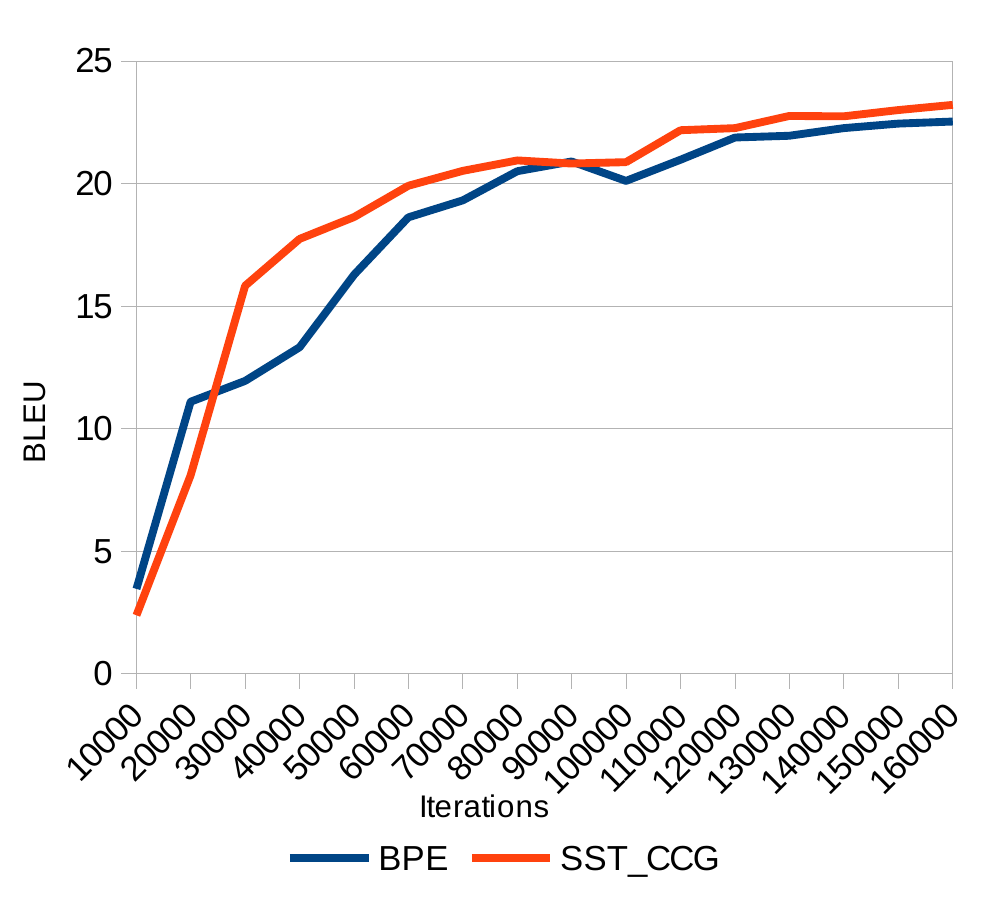}
\captionof{figure}{Baseline (BPE) vs Combined (SST--CCG) NMT Systems for EN--FR, evaluated on the newstest2013.}
\label{pic:en_fr_all}
\end{figure}

Figure~\ref{pic:en_fr_all} compares the BPE-ed baseline system (BPE) with the supertag-supersensetag system (CCG--SST) automatically evaluated on the newstest2013 (in terms of BLEU) over all iterations. Not only does the NMT with features improve over the baseline system, it also has a more robust and consistent learning curve. 

\begin{figure}
 \centering
 \includegraphics[scale=1]{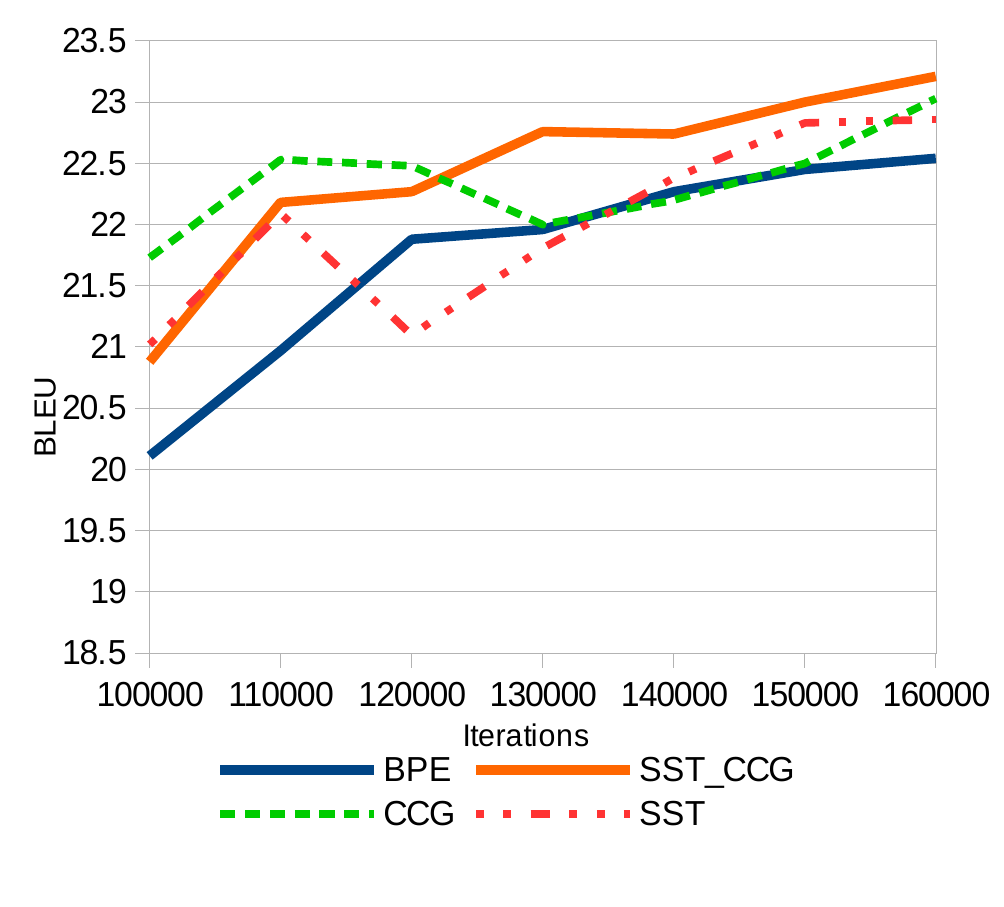}
 \captionof{figure}{Baseline (BPE) vs Syntactic (CCG) vs Semantic (SST) and Combined (SST--CCG) NMT Systems for EN--FR, evaluated on the newstest2013.}
 \label{pic:en_fr_zoom}
\end{figure}

To see in more detail how our semantically-enriched SST system compares to an NMT system with syntactic CCG supertags and how a system that integrates both semantic features and syntactic features (SST+CCG) performs, a more detailed graph is provided in Figure~\ref{pic:en_fr_zoom} where we zoom in on later stages of the learning process for the out-of-domain data. Although Sennrich and Haddow \citeyearpar{Sennrich2016} observe that features are not necessarily cumulative (possibly as the information from the syntactic features partially overlapped), the system enriched with both semantic and syntactic features outperforms the two separate systems as well as the baseline system on an out-of-domain test set in the final stages of the learning process. The best CCG-SST model (23.00 BLEU) outperforms the best BPE-ed baseline model (22.45 BLEU) by 0.55 BLEU (see Table~\ref{tbl:BLEU-FR}), a relative improvement of 2.4\%. Moreover, the benefit of syntactic and semantic features seems to be more than cumulative at some points, confirming the idea that providing both information sources can help the NMT system learn semantico-syntactic patterns. This supports our hypothesis that semantic and syntactic features can be particularly useful when combined. However, for the in-domain data, the benefits of combining supersenses with syntactic supertags is less clear, although as observed in Table~\ref{tbl:scoresSingleFeatBIO} a system using a combination of syntactic and semantic features (in the form of POS-tags with supersenses) was the system that obtained the highest overall BLEU score in 6 out of the 15 evaluations points.\footnote{After every 10k iterations of a total of 150k.}

\begin{table}[h!]
\centering
\begin{tabular}{|l|c|c|c|c|c|}
\hline
\bf{BLEU}	    &\bf{BPE}	&\bf{CCG}	&\bf{SST}	&\bf{POS+SST}   &\bf{CCG+SST}	\\ \hline
\bf{Best Model}	&22.45	    &22.53	    &22.83	    &22.93          &\bf{23.00}	    \\ \hline
\end{tabular}
\caption{Best BLEU scores for Baseline (BPE), Syntactic (CCG), Semantic (SST) and Combined (SST--CCG) NMT systems for EN-FR evaluated on the newstest2013}\label{tbl:BLEU-FR}
\end{table}

\subsection{English--German}
The results for the EN--DE system are very similar to those for the EN--FR system: the model converges faster and we observe the same trends with respect to the BLEU scores of the different systems. Table~\ref{tbl:scoresSingleFeatBIODE} show the results for multiple single features (BPE, BOI, POS, CCG, SST) as well as the combination of syntactic and semantic features (POS+SST and CCG+SST) for German on the in-domain Europarl set. Compared to the results for EN--FR in-domain, the BLEU scores for German are a lot lower. From these scores, we see initially how especially POS-tags (POS) are useful. The first 50k iterations, the POS model gives the best BLEU score. In later stages of the training process, more complex features such as supersenses combined with supertags (SST+CCG) obtain high scores. Looking at the last iterations in Table~\ref{tbl:scoresSingleFeatBIODE}, we see how both POS+SST and SST+CCG lead to the highest BLEU scores, i.e. the two systems that combine syntactic features with semantic ones. The differences in usefulness of different feature sets throughout the learning/training process could be explained by the fact that more general information/features are useful at the start of the learning process, while more complex patterns are only employed by the system in later stages. 

\begin{table}[H]
\centering
\begin{tabular}{|l|c|c|c|c|c|c|c|}
\hline
\bf{\#Iter} &	\bf{BASE}&	\bf{BOI}&	\bf{POS}    &	\bf{CCG}    &	\bf{SST}&	\bf{POS+SST}&	\bf{SST+CCG}	\\ \hline
\bf{10k}	&	2.55	 &	3.37	&	\bf{4.22}	&	3.00	    &	2.77	&	3.70		&	3.37		\\ \hline
\bf{20k}	&	8.10	 &	10.39	&	\bf{12.81}	&	11.85	    &	9.93	&	12.12		&	11.43		\\ \hline
\bf{30k}	&	15.77	 &	16.14	&	\bf{16.59}	&	15.9	    &	15.71	&	16.01		&	16.57		\\ \hline
\bf{40k}	&	18.17	 &	17.67	&	\bf{18.59}  &	18.22	    &	18.14	&	18.47		&	18.16		\\ \hline
\bf{50k}	&	18.79	 &	18.26	&	\bf{19.7}   &	19.13	    &	18.6	&	19.66		&	19.34		\\ \hline
\bf{60k}	&	19.23	 &	19.07	&	19.77	    &	18.77	    &	19.89	&	\bf{20.12}	&	20.01	    \\ \hline
\bf{70k}	&	19.86	 &	20.73	&	20.95       &	\bf{21.21}	&	20.47	&	20.64		&	21.16		\\ \hline
\bf{80k}	&	20.93	 &	21.25	&	21.47	    &	20.13	    &	20.86	&	20.90		&	\bf{21.51}	\\ \hline
\bf{90k}	&	21.56	 &	21.51	&	\bf{21.58}  &	20.04	    &	21.27	&	21.04		&	21.38		\\ \hline
\bf{100k}	&	21.53	 &	21.63	&	22.03	    &	21.63	    &	21.84	&	22.21		&	\bf{21.98}	\\ \hline
\bf{110k}	&	21.66	 &	22.17	&	\bf{21.97}	    &	21.92	    &	21.77	&	21.75	&	21.90		\\ \hline
\bf{120k}	&	21.89	 &	21.95	&	22.39	    &	21.9	    &	22.22	&	22.01		&	\bf{22.29}	\\ \hline
\bf{130k}	&	22.25	 &	22.27	&	22.19	    &	22.4	    &	22.51	&	22.53		&	\bf{22.57}	\\ \hline
\bf{140k}	&	22.27	 &	22.22	&	22.66	    &	22.27	    &	22.17	&	22.44		&	\bf{22.85}	\\ \hline
\bf{150k}	&	22.32	 &	22.16	&	22.5	    &	22.47	    &	22.77	&	\bf{22.81}	&	22.57		\\ \hline
\end{tabular}
\caption{BLEU scores for EN--DE data over the 150k training iterations for the baseline system (BASE) and single features (EBOI, POS, SST and CCG) as well as two combinations of syntactic and semantic features (POS+SST and CCG+SST) evaluated on the out-of-domain News set}\label{tbl:scoresSingleFeatBIODE}
\end{table}

\begin{figure}[H]
 \centering
 \includegraphics[scale=1]{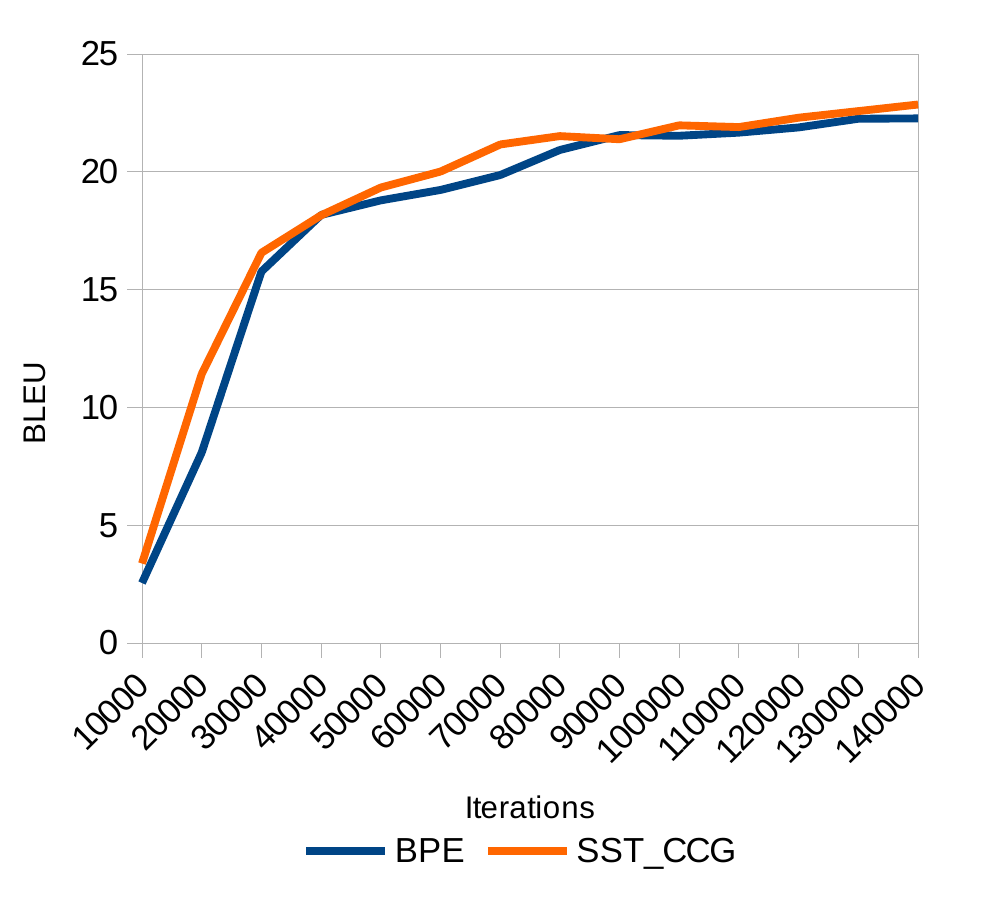}
\captionof{figure}{Baseline (BPE) vs Combined (CCG--SST) NMT Systems for English--German, evaluated on the Europarl test set.}
\label{pic:en_de_all}
\end{figure}

Figure~\ref{pic:en_de_all} compares the BPE-ed baseline system (BPE) with the NMT system enriched with SST and CCG-tags (SST--CCG). Although less clear than for the French data, the SST+CCG learning curve is overall higher than the baseline BPE (BPE). 

\begin{figure}
 \centering
 \includegraphics[scale=1]{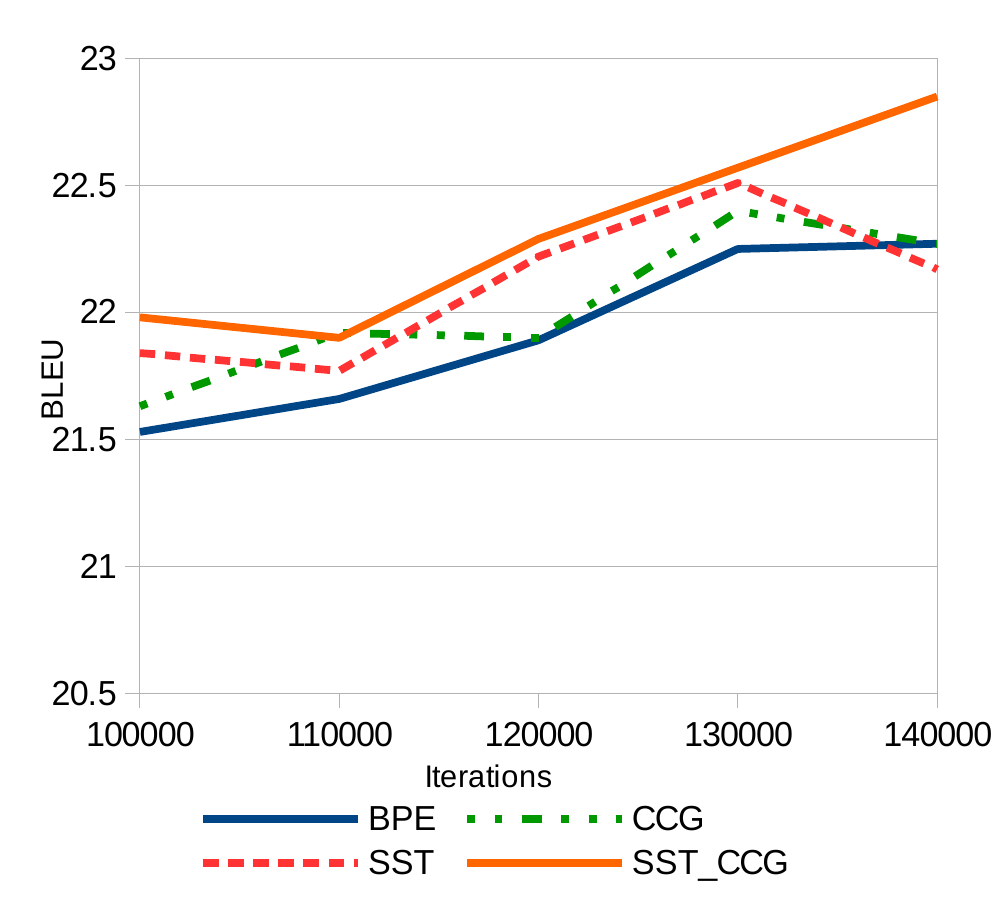}
 \captionof{figure}{Baseline (BPE) vs Syntactic (CCG) vs Semantic (SST) and Combined (CCG--SST) NMT Systems for EN--DE, evaluated on the Europarl test set.}
 \label{pic:en_de_zoom}
\end{figure}

In the last iterations, we see in Figure~\ref{pic:en_de_zoom} how the two systems enriched with supersense tags and CCG-tags have small improvements over the baseline. However, their combination (SST--CCG) leads to a more robust NMT system with a higher BLEU score (see Table~\ref{tbl:BLEU-DE}).

\begin{table}[H]
\centering
\begin{tabular}{|l|c|c|c|c|}
\hline
\bf{BLEU}	&	\bf{BPE}	&	\bf{CCG}	&	\bf{SST}	&	\bf{CCG+SST}	        \\ \hline
\bf{Best Model}	    &	22.32	&	22.47	&	22.77	&	\bf{22.85}	\\ \hline
\end{tabular}
\caption{Best BLEU scores for Baseline (BPE), Syntactic (CCG), Semantic (SST) and Combined (SST--CCG) NMT systems for EN-DE evaluated on the Europarl test set.}\label{tbl:BLEU-DE}
\end{table}

\section{Conclusions}\label{sec:ConcFut}

Although NMT outperforms PB-SMT for relatively simple agreement issues (Chapter~\ref{ch:Agreement}), when looking at more complex patterns related to aspect and tense (Chapter 2), there are still many issues remaining. NMT has the potential to generalize and encode information over the entire sentence but does not always decode all the information properly. This motivated us to experiment with the integration of general linguistic features on the sentence level. We aimed at integrating both higher-level semantics features (in the form of supersenses) and more fine-grained syntactic features (in the form of POS-tags and CCG-tags). This partially answers RQ2. We will further explore different ways of integrating features in NMT in Chapter~\ref{ch:Gender}.

From our experiments, it results that, in terms of automatic evaluation, integrating linguistic features of various kinds can lead to improvements over a baseline BPE-ed state-of-the-art NMT system. In some cases, combining both syntactic and semantic features led to better results than using them separately. This particularly seemed to be the case when using out-of-domain data to evaluate the systems. That can be explained by the fact that the system needs to generalize better over the seen data in order to be able to transfer that knowledge to a different domain. Although the results are promising, the BLEU improvements are small. It is also worth noting that automatically tagging entire corpora with POS-tags and CCG-tags is a time-consuming task and could potentially propagate errors. Furthermore, tools such as CCG-taggers are only available for some languages. 

In addition, we attempted a manual analysis to see where the improvements came from, but as we were using multiple features combined with each other, it was very hard to determine what was going on exactly and pinpoint where the improvements came from. Therefore, in the next chapter, we decided to focus again on a single linguistic phenomenon. We have covered the topic of subject-verb number agreement in Chapter~\ref{ch:Agreement}, and came to the conclusion that NMT is particularly good at morphosyntax, especially compared to the previous phrase-based model. However, when looking further into subject-verb agreement and inspired by the (back then) recent paper published by \cite{Rabinovich2017} on gender domain-adaptation for PB-SMT, we decided to delve into the topic of subject-verb agreement once more, this time focusing on `natural' gender agreement instead of number agreement.

\chapterbib

%% file: GenderNMT.tex
~\newpage
\epigraph{Raw data is both an oxymoron and a bad idea; to the contrary, data should be cooked with care.}{\textit{Geoffrey C. Bowker}}

\chapter{Gender Agreement in Neural Machine Translation}\label{ch:Gender}
%


As shown in Chapter~\ref{ch:Agreement}, NMT is particularly good at getting simple subject-verb number agreement right. However, gender agreement differs from number agreement as it is more complex and it more often requires additional information that is expressed in the broader context of the sentence. Apart from having retained a limited amount of features related to natural gender,\footnote{nouns: `actress' vs `actor', `waitress' vs `waiter'...; pronouns: `she' vs `he', `hers' vs `his'... Occasionally, female pronouns can also be used to refer to ships, airplanes and sometimes countries.} 
English is a relatively gender-neutral language as there is no grammatical gender. Other languages, such as Romance languages or Slavic languages, do mark natural and grammatical gender formally. As a result, they require the human or machine translator to pick between a male, female or neuter variant. To illustrate, consider the English sentences presented in Example~(\ref{gendermonkey})~\citep{Hutchins1992}.

\begin{li}
\label{gendermonkey}
\item (a) \textbf{The monkey} ate the banana because \emph{it} was hungry.
\item (b) The monkey ate \textbf{the banana} because \emph{it} was ripe.
\item (c) The monkey ate the banana because \textbf{it} was tea-time.
\end{li}

The English pronoun `it' refers to `the monkey' in (\ref{gendermonkey}a), `the banana' in (\ref{gendermonkey}b) and the time of the action in (\ref{gendermonkey}c). When translating these sentence into French, `it' has to agree (in number and gender) with its antecedent . As `it' refers to something different in each of these sentences, the French translations differ.\footnote{We would like to note that GNMT (June 2019) translated all three sentences correctly for both French and German, but failed to make the correct agreement for~(\ref{gendermonkey2}b) in Dutch.} In~(\ref{gendermonkey2}a) `it' refers to the male noun `singe' (`monkey') and is thus translated into `il'; in~(\ref{gendermonkey2}b) `it' refers to the female noun `banane' (`banana') and is translated into `elle', the appropriate translation for `it' in~(\ref{gendermonkey2}c) is `ce'.\footnote{Here reduced to `c' because of an elision.} 

\begin{li}
\label{gendermonkey2}
\item (a) Le singe a mangé la banane parce qu'il avait faim.
\item (b) Le singe a mangé la banane parce qu'elle était mûre.
\item (c) Le singe a mangé la banane parce que c'était l'heure du thé.
\end{li}


In this chapter, we will start by introducing some of the problems related to gender and (machine) translation. After having highlighted some of the issues and showing NMT's inability to handle those consistently, we propose a novel method of integrating gender information into the NMT pipeline and discuss some of its advantages and shortcomings. We also hint at the underlying cause, and will delve deeper into this topic in Chapter~\ref{ch:Loss}. We already experimented with the integration of linguistic features in the previous chapter (Chapter~\ref{ch:Supertag}). In the current chapter, we continue doing so, aiming to provide a more complete answer to RQ2 by identifying additional shortcomings of NMT systems and by providing features that can potentially resolve them.

\section{Introduction}

When translating from one language into another, original author traits are partially lost, both in human and machine translations~\citep{Mirkin2015,Rabinovich2017}. However, in the field of MT one of the most observable consequences of this missing information are morphologically incorrect variants due to a lack of agreement in number and gender with the subject. Such errors harm the overall fluency and adequacy of the translated sentence. Furthermore, gender-related errors are not just harming the quality of the translation as getting the gender right is also a matter of basic politeness. Current systems have a tendency to perpetuate a male bias which amounts to negative discrimination against half the population and this has been picked up by the media.\footnote{\url{https://slator.com/technology/he-said-she-said-addressing-gender-in-neural-} \url{machine-translation-racist-and-sexist-biases-research}} 

Human translators rely on contextual information to infer the gender of the speaker in order to make the correct morphological agreement. However, most current MT systems do not; they simply exploit statistical dependencies on the sentence level that have been learned from large amounts of parallel data. Furthermore, sentences are translated in isolation. As a consequence, pieces of information necessary to determine the gender of the speakers might be lost. In such cases the MT system will opt for the statistically most likely variant, which depending on the training data, will either be the male or the female form. Additionally, in the field of MT, training data often consists of both original and translated parallel texts: large parts of the texts have already been translated, which, as studied by Mirkin et al.~(\citeyear{Mirkin2015}), does not preserve the original demographic and psychometric traits of the author, making it very hard for an NMT system to determine the gender of the author.

With this in mind, a first step towards the preservation of author traits would be their integration into an NMT system. As `gender' manifests itself not only in the agreement with other words in a sentence, but also arguably\footnote{As will be further discussed in Section~\ref{sec:relworkG}, there is no consensus with respects to this claim in the field of linguistics. Furthermore, the way gender presents itself in a language is highly language-dependent.} in the choice of context-based words or on the level of syntactic constructions, the sets of experiments conducted in this chapter focus on the integration of a gender feature into NMT for multiple language pairs. 

Within the field of Machine Learning and AI, there is a strong belief that, from the moment we have very large data sets available and we add enough depth to our deep learning algorithms, we can `let the data decide' in one way or another. Because of the nature of the problem we are trying to tackle in combination with the human data we are feeding to the algorithms, simply adding more depth or more data will not be sufficient. We illustrate this with some example translations generated by GNMT.\footnote{May 2019}

\begin{li}
\label{genderIssues1}
\item {\em EN N.}: I am beautiful.
\item {\em ES M.}: Soy hermos\textbf{o}.
\item {\em EN N.}: I am a surgeon.
\item {\em ES M.}: Soy cirujan\textbf{o}
\item {\em EN N.}: I am a beautiful surgeon.
\item {\em ES F.}: Soy un\textbf{a} hermos\textbf{a} cirujan\textbf{a}
\end{li}

In Example~(\ref{genderIssues1}), the English sentences (EN) are all Neutral (N). However, when translating them into a language such as Spanish (ES), the (human or machine) translator needs to pick either the male (M) or female (F) variant of words such as `beautiful' (`hermoso' (M) or `hermosa' (F)) and surgeon (`cirujano' (M) or `cirujana' (F)). A human translator would without any further context have to provide both translation options or justify translating it into a default male or female gender variant. Instead, the translations provided by GNMT switch (arguably) arbitrarly between the male and female variants. Its choice is based on what it has learned from large amounts of data. Although the GNMT system outputs a male variant for both `I am beautiful' and `I am a surgeon', when assigning the attribute `beautiful' to `surgeon', the Spanish translation suddenly becomes female. Although all the translations generated are theoretically correct, they are inconsistent and are reflecting biases picked up from the raw human data.

Similar to the previous example, Example~(\ref{genderIssues2}) shows a comparable inconsistency when translating into French.

\begin{li}
\label{genderIssues2}
\item {\em EN N.}: I am smart.
\item {\em FR M.}: Je suis \textbf{intelligent}.
\item {\em EN N.}: I am beautiful.
\item {\em FR M.}: Je suis \textbf{beau}.
\item {\em EN N.}: I am beautiful but not smart.
\item {\em ES F.}: Je suis \textbf{belle} mais pas \textbf{intelligente}.
\end{li}

`I am smart' and `I am intelligent' are translated into the male form in French. When adding the coordinating conjunctive `but' that indicates a contrast between the two clauses, the French translation suddenly becomes female.

Another example that illustrates how a simple colour choice can affect the translation is given in~(\ref{genderIssues3}):

\begin{li}
\label{genderIssues3}
\item {\em EN N.}: I'm happy with my blue toy.
\item {\em FR M.}: Je suis heureu\textbf{x} de mon jouet bleu.
\item {\em EN N.}: I'm happy with my pink toy.
\item {\em FR M.}: Je suis heureu\textbf{se} de mon jouet rose.
\end{li}

It is hard to determine what drives the GNMT system to opt for one variant or another. We encountered examples where adding a comma, a fullstop, or simply switching the position of words led to changes with respect to gender agreement. Therefore, it is also dangerous to simply assume that, for example, the colour `pink' is associated with women, while `blue' is associated with men. Still, these kinds of examples do illustrate how one simple change to a sentence can lead to significant changes in the translations.

Example~(\ref{genderIssues4}) shows how GNMT loses information that is necessary when translating from a gender-marking language into another gender-marking one: Bulgarian (BG) to French. The Bulgarian sentence \selectlanguage{russian}`Щастлива съм'\selectlanguage{english} (`I am happy') is marked for the female gender with the word \selectlanguage{russian}`щастлив\textbf{а}'\selectlanguage{english}. The French translation, however, uses the male form for the word `happy', i.e. `heureu\textbf{x}'.

\begin{li}
\label{genderIssues4}
\item {\em BG F.}: \selectlanguage{russian}Щастлива съм.\selectlanguage{english}
\item {\em FR M.}: Je suis heureu\textbf{x}\footnote{The `.' was missing in the GNMT translation so we decided to leave it out here as well. The translations were generated with GNMT in May 2019.}
\item {\em EN N.}: I'm happy.
\end{li}

Unlike the other examples presented, where we saw how GNMT produces inconsistent translations, this last example shows a mistake with respect to gender agreement. The gender expressed in the source language is lost during the translation and as a result, the female variant is changed into a male one. We hypothesize that GNMT uses a pivot language for BG--FR. This pivot language is most likely English. English does not mark gender in this sentence, so, `I am happy' is translated into the default male form in French.

The GNMT team announced on the 6th of December 2018 that they updated their translation framework and integrated gender features into their pipeline in order to reduce gender bias. Nevertheless, male/female translations for the examples covered here are not yet provided or dealt with.\footnote{GNMT updated its translation for single word queries from English into French, Italian, Portuguese and Spanish. On the sentence level, Turkish 3rd person neutral sentences are translated into both the male/female variant in English (`he'/`she').} All four examples show that so far, gender is still not handled consistently (Examples~(\ref{genderIssues1}),(\ref{genderIssues2}) and (\ref{genderIssues3})) and sometimes even completely ignored (Example~(\ref{genderIssues4})).

We will start by discussing the related work in Section~\ref{sec:relworkG}. Gender terminology is covered in Section~\ref{sec:genderTerm}. Section~\ref{sec:data} describes and analyses the datasets that were compiled. The experiments conducted are discussed in Section~\ref{sec:experimentsG} and the results are presented in Section~\ref{sec:results}. Finally, our conclusions are presented in Section~\ref{sec:conclusionsGender}.

\section{Related Work}\label{sec:relworkG}
First, we describe some of the work on gender and language that has been conducted in the field of linguistics (Section~\ref{subsec:relwLing}). Next, we briefly comment on the usage of statistical and/or neural models for gender prediction (Section~\ref{subsec:genderPr}). We then describe the work on gender and personalization in PB-SMT conducted prior to our work in Section~\ref{sec:relworkGenderSMT}. Finally, Section~\ref{sec:relworkGenderNMT} describes the related work that has been conducted in NMT, appearing simultaneously or after our publication(s). 

\subsection{Linguistics}\label{subsec:relwLing}
One of the pioneering works for gender and language is the work `Language and Woman's place'~\citep{lakoff1973language}, where Lakoff describes how male and female spoken language differ. Her research identifies certain characteristic of female speech such as the usage of hedges (e.g. `It seems like,') and tag questions (e.g. `, aren't you?'). Since her work, there has been a large body of theoretical and more empirical studies in the field of (socio-)linguistics on various aspects related to language and gender. Several studies, like Lakoff's initial work, identify characteristics typical to female and male discourse. We will discuss some of the studies and characteristics attributed to male or female writing and/or speech. However, as the picture is not always clear, we also highlight how some of this research provides contradictory evidence. 

The work by \cite{Mondorf2002} focuses on marked gender differences in syntax, although linked to semantic types of clauses. On a semantic dimension, her analysis revealed that clauses that express a low commitment to the truth of the proposition expressed (e.g. causal and purpose) are favoured by women. Causal and purpose clauses can be seen as attenuators as they explain a given statement or reasoning. Omitting them can turn a conversation into a confrontation~\citep{gunthner1992}. An example sentence containing a purpose clause is given in~(\ref{ex:causal}).

\begin{li}
\label{ex:causal}
\item ``I'll just break one open \emph{so that you can see the rich mushroom filling}.''~\citep{Mondorf2002}
\end{li}


In contrast, concessive clauses that typically express the highest degree of speaker commitment are more frequently used by men. An example of a concessive clause is provided in~(\ref{ex:concessive}) where the speaker first admits not knowing much about the Croydon Council but later on still claims ``they're wrong''.

\begin{li}
\label{ex:concessive}
\item ``Though I don’t know much about the Croydon Council, I’m sure they’re wrong about that.''~\citep{Mondorf2002}
\end{li}

Furthermore, men preferred the usage of preposed adverbial clauses, while postposed clauses were used more by women~\citep{Mondorf2002}\footnote{An example of a preposed adverbial clause: ``Before I can go on a vacation, I have to finish this thesis.''. The same sentence with a postposed adverbial clause would be: ``I have to finish this thesis before I can go on a vacation.''.}. As such, male language use can be seen as more assertive compared to female language.

Another empirical study conducted by Newman et al.~(\citeyear{Newman2008}) reported systematic differences between men and women when using language. Men use more articles, quantifiers and spatial words while women use more personal pronouns, intensive adverbs and emotion words. Furthermore, women are more likely to discuss topics related to family or social life. As can be seen, some of these difference in language are more syntactic in nature (e.g. usage of articles), while others are more related to semantics and topic preferences (emotion words). The differences related to gender were larger on tasks that place little constraints on language use~\citep{Newman2008}.

A more recent study by Park et al.~(\citeyear{Park2016}) uses social media to explore language usage differences across self-identified males/females. They examined differences in topic preferences as well as more general characteristics such as affiliation and assertiveness. Their results indicate that, in terms of topic preferences, most language differed little across gender. However, they did find substantial gender differences in terms of affiliative language\footnote{Some of the examples give to illustrate affiliative language are words such as `friend', `blessed' and `wonderful'~\citep{Park2016}.} and slight differences in terms of assertiveness.\footnote{Words such as `genuine', `let's' and `absolutely' were marked as assertive~\citep{Park2016}.} On the one hand, self-identified females were more compassionate, warmer and surprisingly (as it contradicts previous findings~\citep{Mondorf2002,Leaper2007} that indicated men use more assertive language compared to women) slightly more assertive in their language use. Self-identified males, on the other hand, used a colder, more hostile and less personal language.

Mulac et al.~(\citeyear{Mulac1988}) show in their study that, while men use more directives, women use questions more commonly. However, a more recent study~\citep{Mulac2000} by the same first author concluded that men ask more questions. We ought to note that both studies were conducted in different domains, i.e. the first one was focused on dyadic interactions~\footnote{A dyadic interaction refers to an interaction between two people.} while the second study focused on male and female managers giving professional criticism in role play.

For a more complete overview of gender differences in language, we refer to the literature review of Mulac et al.~(\citeyear{Mulac2001}). Although well-studied, from our (limited) literature review it appears that empirical investigations have yet to converge to a more coherent picture of differences between male and female speech and writing. This conclusion was also drawn in a more elaborate study by~\cite{Newman2008}.

\subsection{Gender Prediction}\label{subsec:genderPr}
As obtaining high accuracies on gender prediction tasks suggests that there is indeed a difference between male and female language, we briefly discuss some of the work on author profiling (AP) focusing particularly on gender prediction. However, similar to the results obtained in linguistic studies, accuracies vary greatly between different gender prediction tasks depending on the domain, language and the amount of tokens provided.

The main focus in AP has been on predicting gender using in-domain data for training and testing. The yearly PAN evaluation campaigns\footnote{\url{pan.webis.de}} have led to the development of state-of-the-art in-domain gender prediction models on Twitter data for English achieving accuracies up to $80\%-85\%$ \citep{alvarez2015inaoe,rangel2015overview,basile2017,rangel2017overview}. Such high accuracies would indeed suggest there is a detectable difference between male and female language. 

PAN 2016 differed from previous gender prediction tasks as it was the first shared task focusing on cross-genre gender prediction. Twitter data was provided for training while the test data was another `unknown' type of social media text. It should be noted, however, that although the test data differed from the training data, all the data still belonged to the broader `social media' domain. The best scores recorded for gender prediction were $62\%$, $73\%$ and $76\%$ for Dutch, Spanish and English, respectively~\citep{rangel2016overview}. An additional analysis of the cross-genre results by Medvedeva et al.~(\citeyear{medvedeva2017analysis}) revealed that the portability of cross-genre models is only successful when the subdomains are close enough. The PAN-RUS Profiling at FIRE’17 focused on predicting gender across different domains (Twitter, Facebook, essays and reviews) obtaining accuracies between $65\%-93\%$~\citep{litvinova2017overview} depending on the domain. Similarly, in order to capture more domain-independent and thus deeper gender-specific features, the EVALITA 2018 Campaign~\citep{Caselli2018} organized a cross-genre prediction task across five domains (Children Writings, Twitter, YouTube, News, and Personal Diaries) with low accuracies ranging between 51\% (YouTube) and 64\% (Children Writings).

The 2019 CLIN shared task focused on cross-genre gender prediction. An important difference between the two previous tasks on real cross-genre gender prediction and the 2019 CLIN shared tasks is that, unlike Russian and Italian, gender agreement with the first person is very rare in Dutch.\footnote{Exceptions would be certain sentences where the noun agrees in gender with the subject, e.g., `Ik ben een acteur' (masc.) vs `Ik ben een actrice' (fem) [English: `I am an actor/actress']} In Russian and Italian, verbs, adjectives and nouns (can) reflect the gender of the speaker, which facilitates gender prediction. We participated in the CLIN shared task and won both the in-domain and out-domain tasks despite low accuracies for the cross-genre domain setting (53\%--58\%)~\citep{vanmassenhove2019abi}. Our winning model  consisted of  a  weighted  ensemble  model  combining  25 models achieving the highest accuracy on our development set. The ensemble model combined various neural models: self-attention models, LSTMs with attention and pre-trained SpaCy models~\citep{spacy2}. We also experimented with more traditional statistical models and linguistic feature engineering but the neural models outperformed all other approaches.

The mixed results obtained in cross-genre prediction go hand in hand with the empirical linguistic studies discussed in the previous section (Section~\ref{subsec:relwLing}). In AP, gender prediction seems to achieve relatively high accuracies only for certain languages in specific in-domain settings. Like the linguistic studies, the results are not consistent enough to draw a coherent picture or to determine whether there are indeed more general differences within a language, let alone universal differences between male and female speech.

\subsection{Statistical Machine Translation}\label{sec:relworkGenderSMT}
Differences in male and female language use have been studied within various fields related to computational linguistics, including NLP for AP, conversational agents, recommendation systems etc. Within the field of PB-SMT, Mirkin et al.~(\citeyear{Mirkin2015}) motivated the need for more personalized MT. Their experiments show that PB-SMT is detrimental to the automatic recognition of linguistic signals of traits of the original author/speaker. Their work suggests using domain-adaptation techniques to make MT more personalized but does not include any actual experiments on the inclusion of author traits in MT.

The work by Bawden et al.~(\citeyear{Bawden2016}) focuses on speech-like texts and has two main contributions: (i) they create a contextualized parallel corpus of spontaneous dialogues based on the TVD dataset~\citep{Roy2014}, and (ii) they conduct an exploratory experiment on the adaptation of PB-SMT systems based on the gender of the speaker~\citep{Bawden2016}. They experiment with a number of changes in order to adapt a PB-SMT system towards a certain gender using: (i) specific tuning data,  (ii) gender-specific phrase-tables, (iii) gender-specific language models and (iv) a combination of gender-specific phrase-tables and language-models. Their best set-up (which differs for the male/female test sets) leads to an improvement of +0.17 BLEU on the male test set, and +1.09 BLEU on the female test set. When looking into the results, they noticed the results were not to be attributed to improvements in gender agreement but due to lexical choices followed by additions and deletions. They hypothesize that some of the BLEU score improvements might be due to differences in sentence length.

Rabinovich et al.~(\citeyear{Rabinovich2017}) conducted a more elaborate series of experiments very similar to the work by Bawden et al.~(\citeyear{Bawden2016}). Their work on preserving original author traits focuses particularly on gender. As suggested by Mirkin et al.~(\citeyear{Mirkin2015}) and similar to the experiments by Bawden et al.~(\citeyear{Bawden2016}), they treat the personalization of PB-SMT systems as a domain-adaptation task where the female and male gender are treated as two separate domains. They applied two common simple domain-adaptation techniques in order to create personalized PB-SMT: (i) using gender-specific phrase-tables and language models, and (ii) using a gender-specific tuning set. Although their models did not improve over the baseline, their work provides a detailed analysis of gender traits in human and machine translation. 

\subsection{Neural Machine Translation}\label{sec:relworkGenderNMT}

At the time of publication our work was, to the best of our knowledge, the first to attempt to build a speaker-informed NMT system. Our approach is similar to the method proposed in Zero-Shot Translation~\citep{GoogleZST} where an artificial token is inserted at the beginning of the sentence, indicating the desired target language. Sennrich et al.~(\citeyear{Sennrich2016control}) used a similar approach to control politeness adding an `informal' or `polite' tag indicating the level of politeness expressed to the training sentences.

Independent from our work but presented simultaneously at the same event is the work on extreme personalization by Michel and Neubig~\citeyearpar{Michel2018}. Using the data compiled by Rabinovich et al.~\citeyearpar{Rabinovich2017}, they proposed a simple and parameter-efficient adaptation technique unique to each particular user. The adaptation of the NMT system consists of changing the bias of the output softmax to a particular user. As such, their approach can be seen as an extreme version of domain adaptation. Their results showed that such adaptations can allow the model to better reflect linguistic variations, achieving a maximum gain of +0.83 BLEU on English-to-German using a small proportion of the bias parameters. Furthermore, the work by ~\cite{elaraby2018} presents a technique for the translation of speech-like texts focusing particularly on English-to-Arabic. They train a baseline system on generic data (4M parallel sentences) and use a set of gender-labelled sentences (900K) in order to tune the system towards generating translations with correct gender agreement. The labelled sentences were obtained by using an Arabic POS-tagger and a set of rules to identify the gender of the speaker/listener based on the endings of specific Arabic words. They obtain a +2.14 BLEU improvement on a gender-labelled test set with the approach.


More recently,~\cite{Moryossef2019} presented a simple yet effective black-box approach to control the NMT system's translations in case of gender ambiguity. Instead of appending a token, they concatenate unambiguous artificial antecedents with information on the speaker and the interlocutors to ambiguous English sentences. As an example, the English sentence `I love you' is not only ambiguous in terms of the gender of the speaker (`I') but also with respect to the gender and the number of the addressee (`you'). By adding a parataxis construction such as `She told him:' such ambiguity is removed and they then simply rely on the NMT system's ability to handle coreferences in order to generate a correct translation. The translation will then, in most cases, also contain a parataxis construction, which can be easily removed as it is grammatically isolated from the rest of the sentence. They achieve an improvement of $+2.3$ BLEU for English to Hebrew, and a more detailed syntactic analysis reveals that their method enables a certain control over the morphological realization in the target sentences.

~\cite{Bau2019} developed an unsupervised method in order to discover whether individual neurons capture specific linguistic phenomena. Their goal is to be able to control the NMT output in a more systematic way by (i) identifying such neurons, (ii) revealing what they capture specifically and (iii) activating or deactiving the aforementioned neurons in order to control the NMT translations in a predictable way. They experiment with three linguistic properties: tense, number and gender. From their experiments, it appears that gender is the most difficult property to control, with a 21\% success rate using the top-five identified neurons. They hypothesize that the inability to successfully handle gender can be explained by the fact that gender is a property that is distributed, which makes controlling it a hard task. 

Finally, ~\cite{Font2019} use two debiasing techniques on pre-trained GloVe embeddings~\citep{pennington2014glove} and employ them in a Transformer~\citep{Vaswani2017} translation architecture. Their experiments on English to Spanish show gains up to 1 BLEU point. Although there has been a large body of research on debiasing word embeddings for NLP~\citep{bolukbasi2016man,zhao2017men} and counterfactual data augmentation~\citep{zhao2018learning}, we believe that they do not offer a solution to the problem we are aiming to tackle. First of all, debiasing techniques would not offer a solution for MT as even MT systems trained on debiased data will still have to pick a gendered morphological variant in case of ambiguity. Although these might be less biased than the outputs we currently observe, merely debiasing does not offer any control over the generated translations. Second, adjectives of past participles that agree with the natural gender of the subject of a sentence can also appear in agreement with the grammatical gender of (inanimate and animate) nouns. Moreover, a recent paper has shown that current debiasing techniques only superficially remove bias ~\citep{Gonen2019}. Another recent paper~\citep{Nissim2019} demonstrates that due to theoretical problems in previous work on bias in word embeddings, some of the most widely used biased analogies\footnote{Such as, ``Man is to doctor as woman is to nurse''} do not hold up. They argue that, instead of looking for sensational claims, the data should be observed as is (which is already biased enough).

Finally, \cite{Monti2019} provides an overview of issues related to gender in Machine Translation and \cite{Sun2019} a literature review of work related to gender bias in the field of NLP.

\section{Gender Terminology}\label{sec:genderTerm}
Before delving into the experiments conducted, we would like to add a note on the gender terminology used. It is important to make a distinction between the different usages of the term `gender'. Gender, in linguistics, can either refer to:

\begin{itemize}
    \item \textbf{Natural Gender}\\
    Natural gender is either neuter, feminine or masculine. The definition of natural gender according to Collins Dictionary 2018\footnote{Consulted in May 2019 on~\url{ http://www.collinsdictionary.com}} is given below:
    \begin{displayquote}
    Gender based on the sex or, for neuter, the lack of sex of the referent of a noun, as English girl (feminine) is referred to by the feminine pronoun she, boy (masculine) by the masculine pronoun he, and table (neuter) by the neuter pronoun it.
    \end{displayquote}
    There are few features in English that link words to the natural gender of specific people or animals. In such cases the way English handles gender is also referred to as `lexical gender' as the male/female forms are used only when the referent is lexically classified as carrying the `male' or `female' semantic property.
    \item \textbf{Grammatical Gender}\\
    While in English everything is neuter unless it is animate (and it has a natural gender), many other languages are grammatically gendered (French, Dutch, German, Bulgarian,...). Such languages assign an arbitrary gender to every word, which can appear very confusing when unfamiliar with this concept. In Mark Twain's book ``A Tramp Abroad'', one of the chapters is called `The Awful German Language'. In this section, the author provides a literal translation of a German text into English:
    \begin{displayquote}
    ``It is a bleak Day. Hear \textit{the Rain}, how \textbf{he} pours, and \textit{the Hail}, how \textbf{he} rattles; and see \textit{the Snow}, how \textbf{he} drifts along, and oh \textit{the Mud}, how deep \textbf{he} is! Ah \textit{the poor Fishwife}, \textbf{it} is stuck fast in the Mire; \textbf{it} has dropped \textbf{its} Basket of Fishes; and \textbf{its} Hands have been cut by the Scales as \textbf{it} seized some of the falling Creatures; and \textit{one Scale} has even got into \textbf{its} Eye. And \textbf{it} cannot get \textbf{her} out.~\cite[p.~274]{Twain1880}
    \end{displayquote}
    
    This passage illustrates how grammatical gender can be a confusing term. In the last sentence, ``And it cannot get her out'', `it' refers to the animate female referent `the Fishwife' while `she' refers to one of the fish scales. This demonstrates how the grammatical gender of a word and the natural gender of its referent do not necessarily agree. It is worth noting that grammatical gender also manifests itself when determiners, pronouns, or adjectives are inflected in agreement with the (grammatical or natural) gender of the noun that they refer to. How, when and whether this agreement occurs is highly language-dependent and can also interact with other categories such as number and/or case. Grammatical gender should be considered a grammatical category, something close to a noun class. This also explains how some languages have more than 3 grammatical genders. As an extreme example, Swahili, for instance, has 18 noun classes.\footnote{Not all linguists consider all these noun classes grammatical genders, however.}
    
    \item \textbf{Social Gender}
    Social gender is the bias of an unspecified noun towards a certain gender. As an example, `nurse' denotes stereotypically a female person, while `computer scientist' denotes stereotypically a male one~\citep{Hellinger2015}.
\end{itemize}

In our work, these three categories of gender all come into play. However, our main objective is to incorporate information on the natural gender of a speaker in order to resolve issues related to grammatical gender agreement with the natural gender of the speaker. As biases are present in our day-to-day communication, social gender can be picked up from the training data and exacerbated. As such, all three categories are somewhat connected throughout this chapter.

\section{Compilation of Datasets}\label{sec:data}
One of the main obstacles for more personalized MT systems is finding large enough annotated parallel datasets with speaker information. Rabinovich et al.~(\citeyear{Rabinovich2017}) published an annotated parallel dataset for English--French (EN--FR) and English--German (EN--DE). However, for many other language pairs no sufficiently large annotated datasets are available.

To address the aforementioned problem, we published online a collection of parallel corpora licensed under the Creative Commons Attribution 4.0 International License\footnote{To view a copy of this license, visit \url{http://creativecommons.org/licenses/by/4.0/}} for 20 language pairs~\citep{vanmassenhove2018europarl}.\footnote{\url{https://github.com/evavnmssnhv/Europarl-Speaker-Information}} 

We followed the approach described by Rabinovich et al.~(\citeyear{Rabinovich2017}) and tagged parallel sentences from Europarl~\citep{Koehn2005} with speaker information (name, gender, age, date of birth, euroID and date of the session) by retrieving speaker information provided by tags in the Europarl source files. The Europarl source files contain information about the speaker on the paragraph level and the filenames contain the data of the session. By retrieving the names of the speakers together with meta-information on the members of the European Parliament (MEPs) released by Rabinovich et al.~(\citeyear{Rabinovich2017}) (which includes among others name, country, date of birth and gender predictions per MEP), we were able to retrieve demographic annotations (gender, age, etc.). An overview of the language pairs as well as the amount of annotated parallel sentences per language pair is given in Table~\ref{tbl:numberSents}.

\begin{table}[h!]
\centering
\begin{tabular}{|ll|ll|}
\hline
\bf{Languages}  &   \bf{\# sents}    &  \bf{Languages}  &   \bf{\# sents}   \\ \hline
\bf{EN--BG}	&	306,380	    &   \bf{EN--IT}	&	1,297,635	\\ \hline
\bf{EN--CS}	&	491,848	    &	\bf{EN--LT}	&	481,570	    \\ \hline
\bf{EN--DA}	&	1,421,197	&	\bf{EN--LV}	&	487,287	    \\ \hline
\bf{EN--DE} &	1,296,843	&	\bf{EN--NL}	&	1,419,359	\\ \hline
\bf{EN--EL}	&	921,540	    &	\bf{EN--PL}	&	478,008	    \\ \hline
\bf{EN--ES}	&	1,419,507	&	\bf{EN--PT}	&	1,426,043	\\ \hline
\bf{EN--ET}	&	494,645	    &	\bf{EN--RO}	&	303,396	    \\ \hline
\bf{EN--FI} &	1,393,572	&	\bf{EN--SK}	&	488,351	    \\ \hline
\bf{EN--FR}	&	1,440,620	&	\bf{EN--SL}	&	479,313	    \\ \hline
\bf{EN--HU}	&	251,833	    &	\bf{EN--SV}	&	1,349,472	\\ \hline

\end{tabular}
\caption{Overview of annotated parallel sentences per language pair.}\label{tbl:numberSents}
\end{table}

\subsection{Analysis of the EN--FR Annotated Dataset}

We first analysed the distribution of male and female sentences in our data. In the 10 different datasets we experimented with, the percentage of sentences uttered by female speakers is very similar, ranging between 32\% and 33\%. This similarity can be explained by the fact that Europarl is a multilingual corpus with a big overlap between the different language pairs.

We conducted a more focused analysis on one of the subcorpora (EN--FR) with respect to the percentage of sentences uttered by males/females for various age groups to obtain a better grasp of what kind of data we are using for training. As can be seen from Figure~\ref{fig:analysisFM}, with the exception of the youngest age group (20--30), which represents only a very small percentage of the total amount of sentences (0.71\%), more male data is available in all age groups. Furthermore, when looking at the entire dataset, 67.39\% of the sentences are produced by male speakers. Moreover, almost half of the total number of sentences are uttered by the 50--60 age group (43.76\%). 
~\newline
~\newline
\begin{table}[h!]
\centering
\begin{tabular}{|l|l|l|}
\hline
\bf{Age group}	&	\bf{Female}	&	\bf{Male}	\\	\hline
20-30	&	64.47\%	&	35.53\%	 \\	\hline
30-40	&	38.99\%	&	61.01\%	 \\	\hline
40-50	&	36.00\%	&	64.00\%	 \\	\hline
50-60	&	32.79\%	&	67.21\%	 \\	\hline
60-70	&	26.20\%	&	73.80\%	 \\	\hline
70-80	&	16.59\%	&	83.41\%	 \\	\hline
80-90	&	0.00\%	&	100.00\% \\	\hline
\end{tabular}
\caption{Percentage of female and male sentences per age group (EN--FR).}\label{tbl:analysisFM}
\end{table}
~\newline

The analysis shows that there is indeed a gender imbalance in the Europarl dataset, which will be reflected in the translations that MT systems trained on this data produce. Figure~\ref{fig:analysisFM} visualizes the data presented in Table~\ref{tbl:analysisFM}. The visualization shows more clearly the positive trend with respect to gender balance in the Europarl data.

{\begin{figure}[h!]
\centering
\includegraphics{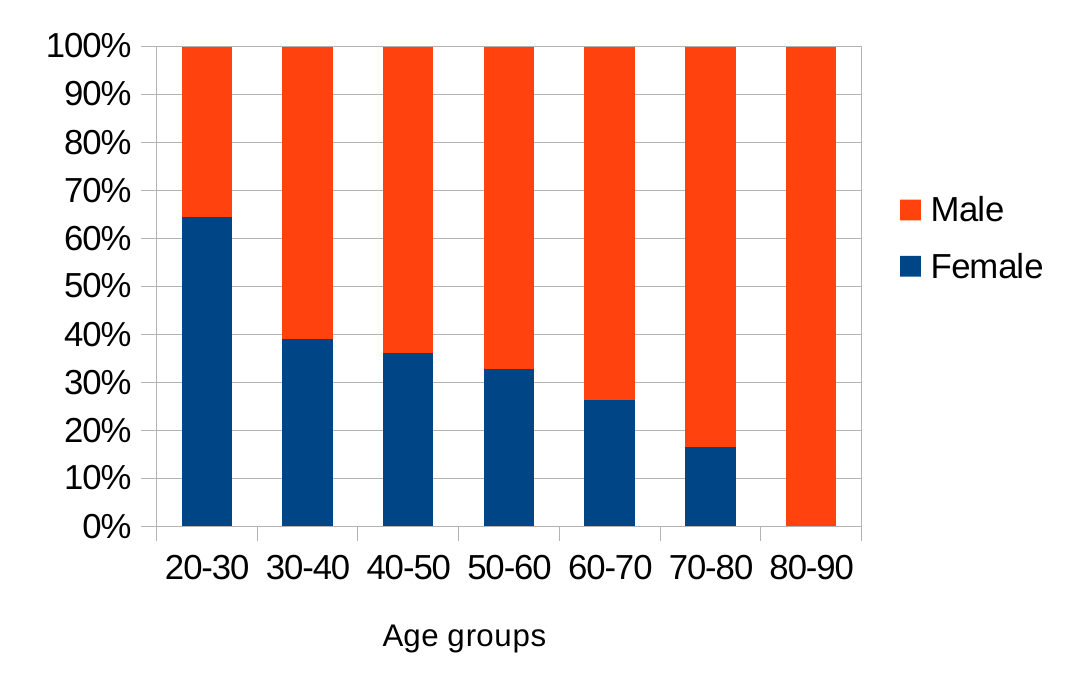}
\caption{Percentage of female and male speakers per age group.}\label{fig:analysisFM}
\end{figure}}

\section{Experiments}~\label{sec:experimentsG}
We briefly describe the datasets used for the set of experiments carried out in Section~\ref{subsec:dataG}, followed by a detailed description of the NMT systems we trained in Section~\ref{subsec:descrG}.
\subsection{Datasets}\label{subsec:dataG}
We carried out a set of experiments on 10 language pairs (the ones for which we compiled more than 500k annotated Europarl parallel sentences):  English-German (EN--DE), English--French (EN--FR), English-Spanish (EN--ES), English-Greek (EN--EL), English--Portuguese (EN--PT), English--Finnish (EN--FI), English--Italian (EN--IT), English--Swedish (EN--SV), English--Dutch (EN--NL) and English--Danish (EN--DA). We augmented every sentence with a tag on the English source side, identifying the gender of the speaker, as illustrated in~(\ref{ExampleComplex}). This approach for encoding sentence-specific information for NMT has been successfully exploited to tackle other types of issues, multilingual NMT systems (e.g. Zero Shot Translation~\citep{GoogleZST}) or domain adaptation~\citep{Sennrich2016control}.

\begin{li}
\label{ExampleComplex}
``FEMALE Madam President, as a...''
\end{li}

For each of these language pairs we trained two NMT systems: a baseline and a tagged one. We evaluated the performance of all our systems on a randomly selected 2K general test set. Moreover, we further evaluated the EN--FR systems on 2K male-only and female-only test sets to investigate the system performance with respect to gender-related issues. We also looked at two additional male and female test sets in which the first person singular pronoun appeared. 

\subsection{Description of the NMT Systems}~\label{subsec:descrG}

We used the OpenNMT-py toolkit~\citep{Klein2017} to train the NMT models. The models are sequence-to-sequence encoder-decoders with LSTMs as the recurrent unit~\citep{Cho2014,Sutskever2014,Bahdanau2014} trained with the default parameters. In order to by-pass the OOV problem and reduce the number of dictionary entries, we use word-segmentation with joint BPE.\footnote{Joint BPE means that the BPE is applied on the concatenation of source and target data.} We ran the BPE algorithm with 89,500 operations~\citep{Sennrich2015}. All systems are trained for 13 epochs and the best model is selected for evaluation.

\section{Results}\label{sec:results}
In this section we discuss some of the results obtained. We hypothesized that the male/female tags would be particularly helpful for French, Portuguese, Italian, Spanish and Greek, where adjectives and even verb forms can be marked by the gender of the speaker. According to the literature, since women and men also make use of different syntactic constructions and make different word choices, we also tested the approach on other languages that do not have morphological agreement with the gender of the speaker such as Danish (DA), Dutch (NL), Finnish (FI), German (DE) and Swedish (SV). 

First, we wanted to see how our tagged systems performed on the general test set compared to the baseline. In Table~\ref{tbl:BLEU}, the BLEU scores for 10 baseline and 10 gender-enhanced NMT systems are presented.

\begin{table}[h!]
\centering
\begin{tabular}{|l|c|c|}
\hline
\bf{Systems}	&	\bf{EN}		&	\bf{EN-TAG}	            \\	\hline
\bf{FR} 		&	37.82	    &   \bf{39.26\rlap{*}}      \\	\hline
\bf{ES}	    	&	42.47		&	42.28 	 	            \\	\hline
\bf{EL}		    &	31.38		&	\bf{31.54}	 	        \\	\hline
\bf{IT}		    &	31.46		&	\bf{31.75\rlap{*}}	 	\\	\hline
\bf{PT}		    &	36.11		&	\bf{36.33}	 	        \\	\hline\hline
\bf{DA}		    &	36.69		&	\bf{37.00\rlap{*}}      \\	\hline
\bf{DE}		    &	28.28		&	28.05 	 	            \\	\hline
\bf{FI}	    	&	21.82		&	21.35\rlap{*}	 	    \\	\hline
\bf{SV}		    &	35.42		&	35.19 	 	            \\	\hline
\bf{NL}	    	&	28.35		&	28.22                   \\	\hline
\end{tabular}
\caption{BLEU scores for the 10 baseline (denoted with \textbf{EN}) and the 10 gender-enhanced NMT (denoted with \textbf{EN-TAG}) systems. Entries labeled with * present statistically significant differences (p $<$ 0.05). Statistical significance was computed with the MultEval tool~\citep{Clark2011}. }\label{tbl:BLEU}
\end{table}

While most of the BLEU-scores in Table~\ref{tbl:BLEU} are consistent with our hypothesis, showing (significant) improvements for the NMT systems enriched with a gender tag (EN-TAG) over the baseline systems (EN) for French, Italian, Portuguese and Greek, the Spanish enriched system surprisingly does not (--0.19 BLEU). As hypothesized, the Dutch, German, Finnish and Swedish systems do not improve. However, the Danish (EN--DA) enriched NMT system does achieve a significant +0.31 BLEU improvement. 

We expected to see the strongest improvements in sentences uttered by female speakers as, according to our initial analysis, the male data was over-represented in the training. To test this hypothesis, we evaluated all systems on male-only and female-only test sets. Furthermore, we also experimented on test sets containing the pronoun of the first person singular as this form is used when a speaker refers to himself/herself. The results on the specific test set for the EN--FR dataset are presented in Table~\ref{tbl:testsets}. As hypothesized, the biggest BLEU score improvement is observed on the female test set, particularly for the test sets containing first person singular pronouns (F1).
~\newline
\begin{table}[h!]
\centering
\begin{tabular}{|l|c|c|}
\hline
\bf{Test Sets}	    &	\bf{EN}		&	\bf{EN-TAG}	        \\	\hline
\bf{FR (M)}		    &	37.58		&	\bf{38.71*}	 	    \\	\hline
\bf{FR (F)} 		&	37.75	    &	\bf{38.97*}  	    \\	\hline
\bf{FR (M1)}	    &	39.00	    &	\bf{39.66*}	 	    \\	\hline
\bf{FR (F1)}		&	37.32		&	\bf{38.57*}   	 	\\ \hline
\end{tabular}
\caption{BLEU-scores on EN--FR comparing the baseline (EN) and the tagged systems (EN--TAG) on 4 different test sets: a test set containing only male data (M), only female data (F), 1st person male data (M1) and first person female data (F1). All the improvements of the EN-TAG system are statistically significant (p $<$ 0.5), as indicated by *.}\label{tbl:testsets}
\end{table}

We had a closer look at some of the translations.\footnote{We used the tool provided by Tilde \url{https://www.letsmt.eu/Bleu.aspx} to see where the BLEU score between the baseline and our tagged systems varied the most.} There are cases where the gender-informed (TAG) system improves over the baseline (BASE) due to better agreement. Interestingly, in Example~(\ref{exmpl:pres}) the French female form of the noun `vice-president' (`vice-pr\'{e}sidente') appears in the translation produced by the BASE system while the male form is the correct one. The gender-informed system does make the correct agreement by using the male variant (`vice-pr\'{e}sident') instead. In Example~(\ref{exmpl:heureux1}) the speaker is female but the baseline system outputs a male form of the adjective `happy' (`heureux').

\begin{li}
\item\begin{tabular}{ll}
(EN) & Being \emph{vice-president}...\\
(Ref) & En tant que \emph{vice-pr\'{e}sident}...\\
(BASE) & En tant que \emph{vice-pr\'{e}sidente}...\\
(TAG) & En tant que \emph{vice-pr\'{e}sident}...\\
\end{tabular}\label{exmpl:pres}
\end{li}

\begin{li}
\item\begin{tabular}{ll}
(EN) & ... I am \emph{happy} that...\\
(Ref) & ... je suis \emph{heureuse} que...\\
(BASE) & ... je suis \emph{heureux} que...\\
(TAG) & ... je suis \emph{heureuse} que...\\
\end{tabular}\label{exmpl:heureux1}
\end{li}

However, we also encountered cases where the gender-informed system fails to produce the correct agreement, as in Example~(\ref{exmpl:not}), where both the BASE and the TAG system produce a male form (`embarass\'e') instead of the correct female one (`embarass\'ee' or `g\^{e}n\'{e}e'). 

\begin{li}
\item\begin{tabular}{ll}
(EN) & I am \emph{embarrassed} that...\\
(Ref) & je suis \emph{g\^en\'{e}e} que...\\
(BASE) & je suis \emph{embarass\'{e}} que...\\
(TAG) & je suis \emph{embarass\'{e}} que...\\
\end{tabular}\label{exmpl:not}
\end{li}

For some language pairs the gender-informed system leads to a significant improvement even on a general test set. This implies that the improvement is not merely because of better morphological agreement, as these kinds of improvements are hard to measure with BLEU, especially given the fact that Europarl consists of formal spoken language and does not contain many sentences using the first person singular pronoun. From our analysis, we observe that in many cases the gender-informed systems have a higher BLEU score than the baseline system due to differences in word choices as in Example~(\ref{exmpl:wordc}) and Example~(\ref{exmpl:wordc2}), where both translations are correct, but the gender-informed system picks the preferred variant.

The observations with respect to differences in word preferences between male and female speakers are in accordance with corpus linguistic studies, which have shown that gender not only has an effect on morphological agreement, but also manifests itself in other ways as males and females have different preferences when it comes to different types of constructions, word choices etc.~\citep{Newman2008,Coates2015}. This also implies that, even for languages that do not mark gender overtly (i.e. grammatically), it can still be beneficial to take the gender of the author/speaker into account.

\begin{li}
\item\begin{tabular}{ll}
(EN) & I think that ...\\
(Ref) & Je pense que ...\\
(BASE) & Je crois que...\\
(TAG) & Je pense que...\\
\end{tabular}\label{exmpl:wordc}
\end{li}

Although more research is required in order to draw general conclusions on this matter, from other linguistic studies it appears that it is indeed the case that there is a relation between the use of the word `pense' (`think') / `crois' (`believe') and the gender of the speaker. To see whether there is a difference in word choice and whether this is reflected in our data, we compiled a list of the most frequent French words for the male data and the female data. Our analysis reveals that `crois' is, in general, used more by males (having position 303 in the most frequent words for males, but only position 373 for females), while `pense' is found at a similar position in both lists (position 151 and 153). These findings are in accordance with other linguistic corpus studies on language and gender stating that women use less assertive speech~\citep{Newman2008}. `Croire' and `penser' are both verbs of cognition but there is a difference in the degree of confidence in the truth value predicated: the verb `croire' denotes more confidence in the truth of the complement clause than the verb `penser' does. In future work, we would like to perform a more detailed analysis of other specific differences in lexical choices between males and females on multiple language pairs.

\begin{li}
\item\begin{tabular}{ll}
(Ref) & J'ai plusieurs remarques...\\
(BASE) & J'ai un nombre de commentaires...\\
(TAG) & J'ai plusieurs remarques...\\
\end{tabular}\label{exmpl:wordc2}
\end{li}

\section{Conclusions}\label{sec:conclusionsGender}

In this chapter, we experimented with the incorporation of speaker-gender tags during the training of NMT systems in order to improve morphological gender agreement with the speaker/writer. Being able to generate the correct morphological variant according to the preferred gender of the speaker/writer is particularly important when translating documents containing speeches, dialogues, moviescripts, etc. We focused particularly on language pairs that express grammatical gender but included other language pairs as well, as linguistic studies have shown that the style and syntax of language used by males and females differs~\citep{Coates2015}. The findings presented in this chapter, combined with what has been discussed previously in Chapter~\ref{ch:Agreement} and Chapter~\ref{ch:Supertag}, finalizes our answer to RQ2 in this thesis.

From the experiments, we see that informing the NMT system by providing tags indicating the gender of the speaker can indeed lead to significant improvements over state-of-the-art baseline systems, especially for those languages expressing grammatical gender agreement. However, while analyzing the EN--FR translations, we observed that the improvements are not always consistent and that, apart from morphological agreement, the gender-aware NMT system differs from the baseline in terms of word choices. 

Changing the translations in terms of word choices instead of enabling the NMT system to deal with issues related to morphological agreement is a(n) (arguably) undesired side-effect. We do not necessarily want female speech to be translated as less assertive in terms of word choices. Furthermore, it is interesting that the NMT system picks up differences in word choices, but does not learn the limited and relatively easy set of rules for gender agreement. 

In this chapter, we briefly addressed the topic of bias in the data and output of NMT systems. Some of the examples that were shown throughout the introduction indicated that NMT systems pick up gender biases that are present in the training data. A handful of studies already indicated that NMT does not only pick up bias from human data; it also exacerbates the already existing biases. In the next chapter, we delve into the topic of algorithmic bias and see how this relates to many of the current remaining issues with MT.

\chapterbib

%% file: LexicalLoss.tex
\tikzstyle{startstop} = [rectangle, rounded corners, minimum width=2cm, minimum height=0.5cm,text centered, draw=black, fill=white!30]
\tikzstyle{transl} = [rectangle, minimum width=2cm, minimum height=0.5cm,text centered, draw=black, fill=red!30]
\tikzstyle{arrow} = [thick,->,>=stealth]

~\newpage
\epigraph{Translating is always about making sacrifices; however, nothing essential should be sacrificed.}{Enrique D\'{i}ez-Canedo}

\chapter{Loss and Decay of Linguistic Richness in Neural and Statistical Machine Translation}\label{ch:Loss}

In the previous chapter, we focused on correcting issues related to gender agreement with the natural gender of the speaker. From the examples we analyzed, we observed that both GNMT and our trained NMT engines seemingly randomly switch between male and female endings in cases of ambiguity. We implemented a solution that improved over the baseline in terms of automatic evaluation metrics and morphological agreement but still lacked consistency. Moreover, we observed that the approach changed some of the translations in terms of word choices. Such side-effects are, although interesting, (arguably) not desirable.

We delved deeper into gender-related issues and encountered a study by \cite{Prates2018}, revealing that GNMT systems overgenerate male or female nouns (referring to professions), even when taking into account the already existing bias in the data.\footnote{In their case, `data' refers to demographic data and not the actual training data used for the NMT systems.} As such, one could speak of an algorithmic bias on top of already biased data, exacerbating the agreement issues. However, aside from briefly alluding to algorithmic bias and suggesting its existence, no further empirical evidence was provided in \cite{Prates2018}. If indeed, PB-SMT and NMT systems do overgeneralize certain seen patterns, words and constructions considerably at the cost of less frequent ones, this not only has consequences for lexical richness but also for these systems' inability to deal with (gender and number) agreement (Chapter~\ref{ch:Agreement} and Chapter~\ref{ch:Gender}), more complex tense patterns (Chapter~\ref{ch:Aspect}) and linguistic richness and complexity in general, as \emph{one-to-many} mappings are inherent to the translation task.

Therefore, in this chapter, we present an empirical approach to quantifying the loss of linguistic richness. We conduct general experiments showing how lexical richness is affected by current NMT systems (RNN, Transformer) and compare this to PB-SMT systems as well as to the original training data the MT systems were trained on. Our automatic and in-depth analysis of specific words shows that NMT systems are unable to deal with lexical richness at the word-level. As we do not believe this phenomenon is limited to lexical richness, we instead speak of a loss of ``linguistic richness''.

Diversity in the translations has not been a priority so far in the MT community as the main focus has understandably been the creation of adequate and fluent translations. However, we argue that this lack of diversity is the underlying cause of many remaining issues that occur in MT and some of them do not just limit the lexical richness of the translations generated, but also affect the adequacy of the sentences produced (complex syntactic patterns, gender bias issues, complex tenses, etc.). As such, this chapter addresses our third and final research question (RQ3) related to identifying the underlying cause of (some of) the limitations of NMT.

Although we do not provide a solution to this problem as NMT systems' ability to generalize is exactly what makes them successful, we do believe it is important to highlight the side-effects of NMT's ability to learn as, apart from limiting the randomness in the language it generates, it could be the underlying cause of its limits. We therefore think it is important to attempt quantifying how much is `lost in (machine) translation'. 



\section{Introduction}

\cite{berman2000translation} observed that the translation process consists of deformation processes, one of which he refers to as `quantitative impoverishment', a loss of lexical richness and diversity. Although mitigated by a human translator, this loss is to some extent inevitable as it is hard to respect the multitude of signifiers and constructions when translating one language into another. While \cite{berman2000translation} studied the decrease of lexical richness of human translations from a theoretical point of view, \cite{Kruger2012} demonstrated using empirical methods that there is indeed a lexical loss when comparing translations to original texts. In the field of MT, \cite{klebanov2013} showed that PB-SMT suffers considerably more from lexical loss than data translated by humans in a study focused on lexical tightness and text cohesion. We are not aware of any other research in this direction. 

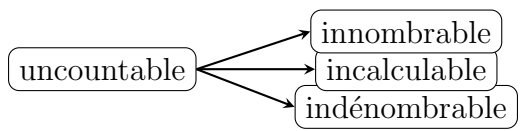
\begin{figure}[h!]
\centering
\begin{tikzpicture}[node distance=2cm]
\node (sw) [startstop, yshift=-0.5cm] {uncountable};
\node (tw1) [startstop, right of=sw, xshift=2cm, yshift=-0.5cm] {ind\'{e}nombrable};
\node (tw2) [startstop, right of=sw, xshift=2cm, yshift=0cm] {incalculable};
\node (tw3) [startstop, right of=sw, xshift=2cm, yshift=0.5cm] {innombrable};
\draw [arrow] (sw.east) -- (tw1.west);
\draw [arrow] (sw.east) -- (tw2.west);
\draw [arrow] (sw.east) -- (tw3.west);
\end{tikzpicture}
\caption{\emph{One-to-many} relation between the English source word `uncountable' and some of its possible French translations `innombrable', `incalculable' and `ind\'{e}nombrable'.} \label{fig:otm}
\end{figure}

\begin{figure}[h!]
\centering
\begin{tikzpicture}[node distance=2cm]
\node (see) [startstop, yshift=-0.5cm] {see};
\node (voir) [startstop, right of=see, xshift=2cm, yshift=2cm] {voir};
\node (vois) [startstop, right of=see, xshift=2cm, yshift=1.5cm] {vois};
\node (voyons) [startstop, right of=see, xshift=2cm, yshift=1cm] {voyons};
\node (voyez) [startstop, right of=see, xshift=2cm, yshift=0.5cm] {voyez};
\node (voient) [startstop, right of=see, xshift=2cm, yshift=0cm] {voient};
\node (voie) [startstop, right of=see, xshift=2cm, yshift=-0.5cm] {voie};
\node (voies) [startstop, right of=see, xshift=2cm, yshift=-1cm] {voies};
\node (voyions) [startstop, right of=see, xshift=2cm, yshift=-1.5cm] {voyions};
\node (voyiez) [startstop, right of=see, xshift=2cm, yshift=-2cm] {voyiez};
\draw [arrow] (see.east) -- (voir.west);
\draw [arrow] (see.east) -- (vois.west);
\draw [arrow] (see.east) -- (voyons.west);
\draw [arrow] (see.east) -- (voyez.west);
\draw [arrow] (see.east) -- (voient.west);
\draw [arrow] (see.east) -- (voie.west);
\draw [arrow] (see.east) -- (voies.west);
\draw [arrow] (see.east) -- (voyions.west);
\draw [arrow] (see.east) -- (voyiez.west);
\end{tikzpicture}
\caption{\textit{One-to-many} relation between English verb `see' and its infinitive translation (`voir') and conjugations (`vois' (1$^{st}$, 2$^{nd}$ person singular present tense), `voyons' (1$^{st}$ person plural present tense), `voyez' (2$^{nd}$ person plural present tense), `voient' (3$^{rd}$ person plural present tense), `voie' (1$^{st}$ person singular subjunctive mood), `voies' (2$^{nd}$ person singular subjunctive mood), `voyions' (1$^{st}$ person plural subjunctive mood) and `voyiez' (2$^{nd}$ person plural subjunctive mood)) in French.} \label{fig:voir}
\end{figure}
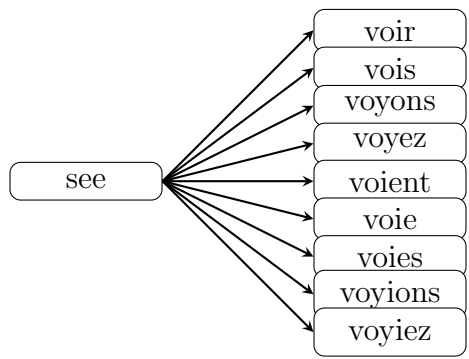

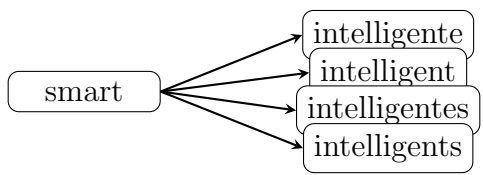
\begin{figure}[h!]
\centering
\begin{tikzpicture}[node distance=2cm]
\node (smart) [startstop, yshift=-0.5cm] {smart};
\node (intelligente) [startstop, right of=smart, xshift=2cm, yshift=0.75cm] {intelligente };
\node (intelligent) [startstop, right of=smart, xshift=2cm, yshift=0.25cm] {intelligent  };
\node (intelligentes) [startstop, right of=smart, xshift=2cm, yshift=-0.25cm] {intelligentes};
\node (intelligents) [startstop, right of=smart, xshift=2cm, yshift=-0.75cm] {intelligents };
\draw [arrow] (smart.east) -- (intelligente.west);
\draw [arrow] (smart.east) -- (intelligent.west);
\draw [arrow] (smart.east) -- (intelligentes.west);
\draw [arrow] (smart.east) -- (intelligents.west);

\end{tikzpicture}
\caption{\textit{One-to-many} relation between English adjective `smart' and its male counterparts `intelligent' (singular) and `intelligents' (plural) and female counterparts `intelligente' (singular) and `intelligentes' (plural) in French.} \label{fig:genderb}
\end{figure}

As generating accurate translations has been the main objective of current MT systems, maintaining lexical richness and creating diverse outputs has understandably not been a priority. Nevertheless, the issue of lexical loss in MT might at the same time be a symptom and a cause of a more serious issue underlying current systems. The difference between a \emph{one-to-many} relationship such as the one illustrated in Figure~\ref{fig:otm}, is very different from the one illustrated in Figure~\ref{fig:voir} or Figure~\ref{fig:genderb} from a (human) translator point of view. Even when a person does not speak the language (here French) used in these examples, just by looking at them it is relatively easy to see that the words in Figure~\ref{fig:voir} and~\ref{fig:genderb} are somehow related as they have the same root while the words in Figure~\ref{fig:otm} do not. However, from a statistical point of view and for the MT systems, they are not always clearly distinguishable. When presented with an ambiguous sentence, like `I am intelligent' or `See?' where there is little context to decide on a particular target variant of the same source word, it essentially boils down to the same thing: picking the translation that maximizes the probability over the entire sentence. As such, the loss of richness and diversity and the exacerbation of already frequent patterns might not simply be limited to the loss of (near) synonyms and/or rare words, but could also be the underlying cause of, for example, the inability of PB-SMT systems to handle morphologically richer language correctly~\citep{vanmassenhove2016improving}, the already observed issues with gender bias~\citep{Prates2018,vanmassenhove2019getting} in MT output or the difficulties of dealing with agglutinative languages~\citep{unanue2018}.

The inability of neural models to generate diverse output has already been observed for tasks involving language generation, where creating intrinsically diverse outputs is more of a necessity~\citep{Li2015,Cao2017,Serban2017,Wen2017,Yang2017}. However, from a translation point of view, the ability of MT systems to be (1) consistent and (2) learn and generalize well are -- compared to previous MT systems -- the biggest asset of NMT. However, we hypothesize that this type of generalization might also have serious drawbacks and that diversity, although not deemed a priority, is of importance for the field of MT as well. Overgeneralization over a seen input and the exacerbation of dominant forms might not only lead to a loss of lexical choice, but could also be the underlying cause of gender bias exacerbation. Although, in the context of gender, some researchers have already alluded to the existence of so-called `algorithmic bias' \citep{zhao2017men,Prates2018}, no empirical evidence has been provided so far.  

With our empirical approach, comparing the lexical diversity of different MT systems and further analyzing the frequencies of words, we aim to shed some light on the relation between the loss of diversity and the exacerbation or loss of certain words. Thus, the first objective of our work is to verify how NMT compares to PB-SMT (and the original data it was trained) on in terms of lexical richness or the loss thereof. The second objective is to quantify to what extent the different MT architectures favour translations that are more frequently observed in the training data.

The structure of the chapter is the following: related work in the field of linguistics, PB-SMT and NMT is described and discussed in Section~\ref{sec:relworkLL}; our hypotheses are defined in detail in Section~\ref{sec:hypothesisLL}; information on the data and the MT systems used in our experiments is provided in Section~\ref{sec:expLL}; Section~\ref{sec:analysisLL} discusses the results of our experiments and finally, we conclude in Section~\ref{sec:conclusionsLL}.

\section{Related Work}\label{sec:relworkLL}
We discuss some of the related work in the field of linguistics, PB-SMT and NMT. We cover both related work on lexical diversity and more recent work mentioning algorithmic bias or bias augmentation in neural models as we believe the two topics are closely related. 

\subsection{Linguistics}
In the field of linguistics and translation theory,~\cite{berman2000translation} researched the so-called deforming tendencies that are inherent to the act of translation. Although these tendencies can be mitigated by the (human) translator, they are to a large extent inevitable. Quantitative impoverishment (or lexical loss), is one of the tendencies mentioned. Berman discusses the abundance of prose and how there is a proliferation of signifiers and signifying chains in the work of great novelists. As an example, he explains how the novelist Robert Arlt uses for the same signified `visage', multiple signifiers such as `semblante', `rostro' and `cara' without justifying a particular choice in a particular sentence. By using multiple signifiers, Arlt simply marks that `visage' is an important reality in his work. When a translation does not respect this multiplicity, there is a quantitative loss. Similarly, \cite{Kruger2012} compared human-translated to comparable non-translated English texts and found the translations to be more simplified in terms of language use than the original writings.

Multiple studies on translation focus on identifying features concerning the nature of translation~\citep{baker1993,berman2000translation} by comparing translations with their originals. Such studies largely concentrate on the analysis of human translations, ignoring those produced by MT systems or translations resulting from the interaction of both~\citep{Lapshinova2015}. Some of the more recent studies (like ~\cite{Lapshinova2015}) do apply corpus-based methods to identify translation features using both human- and machine- (PB-SMT and RBMT) translated texts. One of the features she examined is `simplification'. On the lexical level, she opted to measure simplification by comparing content and grammatical words (which they refer to as lexical density) and by computing the type-token-ratio. On the syntax level, she compared the average length of the sentences. One shortcoming of her approach is that she relied on taggers to extract nouns and other certain patterns in the machine-translated text. Unexpectedly, the average lexical density, computed by comparing content and grammatical words, is higher in PB-SMT than in the human-translated texts. She hypothesizes this is the case because of the fact that all the `untranslated' words were kept in their original form by the PB-SMT system and consequently tagged as proper nouns by the tagger. In terms of standardized type-token-ratio (STTR)~\citep{Scott2006}, she observed a difference between the various human- and machine-translated texts: on average, translations showed lower STTR than the source texts, and the mean value of the human translations was higher than that of the machine translations~\citep{Lapshinova2015}.


\subsection{Statistical Machine Translation}
In the field of MT, the concept of lexical loss/diversity and its importance is indirectly related to the research of~\cite{Wong2012} on cohesion. They illustrated the relevance of the under-use of linguistic devices (superordinates, meronyms, synonyms and near-synonyms) for PB-SMT in terms of cohesion. More directly related to our work is the work of \cite{klebanov2013} who presented findings regarding the loss of associative texture by comparing original and back-translated texts, references and system translations and a set of different MT systems. Although the destruction of the underlying networks of signification might be, to some extent, unavoidable in any translation process, the work of~\cite{klebanov2013} shows that PB-SMT specifically suffers from lexical loss, more than OT.

\subsection{Neural Machine Translation}
In the field of NMT, lexical diversity or the loss thereof has been used as a feature to estimate the quality of NMT systems. \cite{Bentivogli2016} used lexical diversity, measured by using the type-token ratio (TTR), as an indicator of the size of vocabulary as well as the variety of subject matter in a text. Their experiments compared PB-SMT to NMT and the results suggested that NMT is better able to cope with lexical diversity than PB-SMT. 

Related to our work is research on mixture models that aim to increase diversity. Despite the popularity of mixture models in machine learning, there are only a handful of works exploring them for text generation applications~\citep{Li2015,Cao2017,Serban2017,Wen2017,Yang2017} and MT~\citep{He2018,Shen2019}. \cite{He2018} developed a soft mixture model in order to improve the diversity and the quality of neural sequence-to-sequence MT models. They adopt a committee of specialized translation models instead of one single model. Each specialized model selects its own training data, leading to a soft clustering of the parallel data. Most of the parameters of the specialized models are shared and the method only requires a negligible amount of additional parameters. A more recent work, still under revision, by~\cite{Shen2019} argues that \cite{He2018} did not evaluate on multiple references nor analyze the full spectrum of their soft mixture model. \cite{Shen2019} provide a more comprehensive study to shed light on how different settings affect the performance of mixture models. There is a similarity between the problem we are trying to shed light on and the aim of mixture models, i.e. the issues with diversity in MT combined with the question of how to accurately model multi-modal\footnote{Male/female morphological variants, formal/informal, alternative syntactic constructions, usage of the tenses, etc.} output. However, we believe mixture models are more closely related to domain adaptation\footnote{In a broad sense as it would also include formal vs informal language, for example.} and specializing models to be adapted towards more specific tasks while still maintaining the benefits of a bigger, more general model in terms of language usage.  


As mentioned in the introduction, there have been a handful of studies mentioning the possibility of bias in algorithms on top of the existing bias in word embeddings~\citep{bolukbasi2016man,Caliskan2017,garg2018}. The algorithmic bias should be interpreted as an exacerbation of an already biased input, i.e. the inequalities observed in the input are not simply maintained in the output but increased by the models trained on such data. For MT, a study by \cite{Prates2018} provides experimental evidence showing that gender bias found in the output of sentences produced by Google Translate's GNMT is exacerbated when compared to actual demographic data. Compared to other tasks, bias in MT has so far received relatively little attention. This can be explained by the fact that, in comparison to other tasks, MT has little freedom to produce biased output, as the system is heavily constrained by the original source input.\footnote{This is less so for many other NLP tasks such as language generation.} In the field of visual semantic-role labeling and multi-label object classification, a study by \cite{zhao2017men} described a way of handling a phenomenon they call bias amplification. They observe 45\% of verbs and 37\% of objects are gender biased in a proportion greater or equal to a 2:1 distribution. For example, in the training set an activity such as `cooking' is 33\% times more associated with women than with men. After training a model on this training set, the disparity is amplified to 68\% times more associations with women compared to men for the `cooking' activity. They implement constraints into their model so that the disparity is not amplified compared to the training set and is thus kept equal to the observations. Although this is feasible when dealing with a limited set of verbs and nouns, doing so for translations would require too many constraints, some of which might contradict each other. Furthermore, we argue that keeping the bias equal to what has been observed during training might not always be the best solution nor does it offer control over the output. Finally, the work of \cite{Lu2018} focuses on mitigating gender bias by adding counterfactual data targeting NLP tasks such as language modeling and coreference resolution. However, while performing their experiments, they observe that as the training of their language model and coreference resolution model proceeds (i.e. the loss reduces), the observed gender bias grows. This might indicate that the optimization of the model encourages bias. Adding counterfactual data can limit the bias growth but does not offer any control over the output.

\section{Experiments}\label{sec:experimentsLL}
First, we state our hypothesis in Section~\ref{sec:hypothesisLL}. Then, the experiments conducted are described in detail in Section~\ref{sec:expLL}.

\subsection{Hypothesis}\label{sec:hypothesisLL} 

Data-driven PB-SMT paradigms 
are concerned with (i) identifying the most probable target words, phrases, or sub-word units given a source-language input sentence and the preceding decoded information, via the translation model, and (ii) chaining those words, phrases or sub-word units in a way that maximizes the likelihood of the generated sentence with respect to the grammatical and stylistic properties of the target language, via the language model. In NMT, where translation and language modeling are co-occurring in the decoder, it boils down to finding the most likely word at each time step.

Our hypothesis is that the inherent nature of data-driven MT systems to generalise over the training data has a quantitatively distinguishable negative impact on the word choice, expressed by favouring more frequent words and disregarding less frequent ones. We hypothesize that the most visible effect of such bias is to be found in the word frequencies and the disappearance (or `non-appearance') of scarce words. Apart from a general effect on lexical diversity, such behaviour might also lead to the disappearance or amplified use of certain morphological variants of the same word, accounting, for example, for the already observed over-use of male forms in ambiguous sentences, the preference for certain verb forms over other less frequent ones ($3^{rd}$ person $>$ $1^{st}$ person), or the difficulties of MT systems to appropriately handle morphologically richer target languages in general. 

Furthermore, because NMT handles translation and language modelling (or alignment) jointly~\citep{Bahdanau2014,Vaswani2017}, which makes it harder to optimize compared to PB-SMT, we further hypothesise that NMT is more susceptible to problems related to overgeneralisation. Given that the quality of PB-SMT and NMT has been widely explored~\citep{Bentivogli2016,Shterionov2018} we do not question that NMT performs better in terms of adequacy and fluency, but instead investigate how quality evaluation metrics reflect language richness.

\subsection{Experimental Setup}\label{sec:expLL}
To test our hypothesis we built three types of MT systems and analysed their output for two language pairs on Europarl data. The language pairs are English $\rightarrow$ French (EN-FR) and English $\rightarrow$ Spanish (EN-ES). We trained attentional RNN, Transformer and Moses MT systems. To draw more general conclusions on the effects of bias propagation and loss of lexical richness, we assessed output from seen (during training) and unseen data.

\paragraph{Data}
We used +/- 2M sentence pairs from the Europarl corpora for each of the language pairs. We randomised the order of the sentence pairs and split the data into train, test and development sets, filtering out empty lines. Details on the different datasets can be found in Table~\ref{tbl:data}. We chose to include large quantities of data in our test sets -- the unseen data -- in order to maximise the language variability and explore general tendencies.

\begin{table}[]
    \centering
    {\begin{tabular}{|c|c|c|c|}
    \hline
Language pair & Train & Test & Dev\\\hline
EN--FR & 1,467,489 & 499,487 & 7,723\\\hline
EN--ES & 1,472,203 & 459,633 & 5,734\\\hline
    \end{tabular}}
    \caption{Number of parallel sentences in the train, test and development splits for the language pairs (EN--FR and EN--ES) used.}
    \label{tbl:data}
\end{table}

\paragraph{MT systems}\label{sec:systems}
For each of the three MT architectures we first trained a standard MT system (the forward or FF system) on the original data. For the RNN and Transformer systems we used OpenNMT-py. The systems were trained for 150K steps, saving an intermediate model every 5000 steps. We scored the perplexity of each model on the development set and chose the one with the lowest perplexity as our best model, which was used later for translation. The options we used for the neural systems are as follows: 
\begin{itemize}
    \item RNN: size: 512, RNN type: bidirectional LSTM, number of layers of the encoder and of the decoder: 4, attention type: mlp, dropout: 0.2, batch size: 128, learning optimizer: adam~\citep{Kingma2014} and learning rate: 0.0001.
    \item Transformer: number of layers: 6, size 512, transformer\_ff: 2048, number of heads: 8, dropout: 0.1, batch size: 4096, batch type: tokens, learning optimizer adam with beta$_2 = 0.998$, learning rate: 2.
\end{itemize}
All neural systems have the learning rate decay enabled and their training is distributed over 4 nVidia 1080Ti GPUs. The selected settings for the RNN systems are optimal according to \cite{Britz2017}; for the Transformer we use the settings suggested by the OpenNMT community\footnote{\url{http://opennmt.net/OpenNMT-py/FAQ.html}} as the optimal ones that lead to quality on par with the original Transformer work~\citep{Vaswani2017}.

For the PB-SMT engines we use Moses with default settings~\citep{Koehn2007} and a 5-gram language model with pruning of bigrams. Each PB-SMT engine is further tuned with MERT~\citep{Och2003} until convergence or for a maximum of 25 iterations.

For the neural systems, we opted not to use sub-word units as is typically done for NMT. This is because we focus on the word frequencies in the translations and do not want any algorithm for splitting into sub-word units to add extra variability in our data. To construct the dictionaries we use all words in our training data. Table~\ref{tbl:vocabularies} shows the training vocabularies for the source and target sides.

\begin{table}[]
    \centering
    {
    \begin{tabular}{|c|c|c|}\hline
         Language pair & SRC & TRG \\\hline
EN-FR & 113 132 & 131 104 \\\hline
EN-ES & 113 692 & 168 195 \\\hline
    \end{tabular}}
    \caption{Training vocabularies for the English, French and Spanish data used for our models.}
    \label{tbl:vocabularies}
\end{table}

To assess how MT amplifies bias and loss of lexical richness, along with the original-data systems, we trained MT with back-translated (BT) data, which is typically used to complement original data for MT training when the quantity of the original data is not sufficient for reaching high translation quality~\citep{Sennrich2015b,Poncelas2018}. 

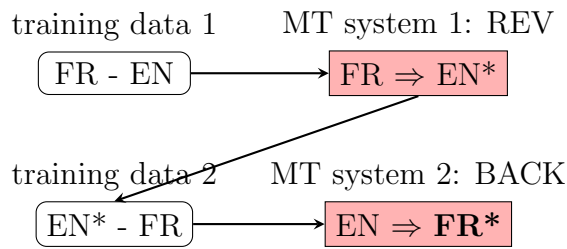
\begin{figure}[h!]
\centering
\begin{tikzpicture}[node distance=4cm]

\node (in1) [startstop, label=north:{training data 1}] {FR - EN};
\node (tr1) [transl, label=north:{MT system 1: REV}, right of =in1] {FR $\Rightarrow$ EN*};
\node (in2) [startstop, label=north:{training data 2}, below of= in1, node distance=2cm] {EN* - FR};
\node (tr2) [transl,  label=north:{MT system 2: BACK}, right of=in2] {EN $\Rightarrow$ \textbf{FR*}};
\draw [arrow] (in1.east) -- (tr1.west);
\draw [arrow] (in2.east) -- (tr2.west);
\draw [arrow] (tr1.south) -- (in2.north);

\end{tikzpicture}
\caption{Back-translated data pipeline example for EN--FR. The same pipeline was used for EN--ES.} \label{fig:revback}
\end{figure}

We first trained MT systems for the reverse language directions, i.e. for FR--EN and ES--EN. We used the same data sets, but reversed the associations of the source and the target with FR/ES $\rightarrow$ EN instead of EN $\rightarrow$ FR/ES. We then used these \textit{reversed} (\textit{REV} or \textit{rev}) systems to translate the training set: the same set used for training the FF systems and the REV systems. That is, we use a system trained on (say) FR-EN data to translate the same FR set into English (EN*). The aim is to see what the impact of the underlying algorithms is on the data in the most-favourable scenario; when the data has already been seen. With the translated English target data, we trained new systems for the EN*$\rightarrow$FR and EN*$\rightarrow$ES directions, where the source data was the back-translated set. We refer to these systems with \textit{BACK} and use the suffix \textit{back} to denote them. We end up with what can be seen as a combination of back-translation and round-trip-translation. See Figure~\ref{fig:revback} for a visualization of the pipeline of systems.

For the REV and BACK engines we used the same settings as for the FF ones. However, at this stage, the source side of the training data is different and thus impacts the learnable vocabulary. Table~\ref{tbl:vocabularies_rev} presents the source-side vocabulary sizes for the RNN, PB-SMT and Transformer systems. These are in practice the number of distinct words of the translations produced by the REV systems. Compared to Table~\ref{tbl:vocabularies}, Table~\ref{tbl:vocabularies_rev} clearly shows how source and target vocabularies are comparable in the original datasets, but translating the same original English dataset with the Neural REV systems (RNN and Trans) results in a huge drop in vocabulary size; with the PB-SMT REV systems the decrease is still significant, but not as profound as in the former case. We would like to note that the comparison between the PB-SMT and the two neural systems in terms of the training vocabularies presented in Table~\ref{tbl:vocabularies_rev} might be distorted because of the fact that PB-SMT deals with `unseen'\footnote{As these are the training vocabularies resulting from the REV systems there should, in theory, not be any unseen tokens. However, in practice we saw that there were Spanish/French tokens copied from the source into the output produced by the PB-SMT systems.} tokens by simply copying the source word into the output. NMT systems instead produce a token indicating that the word is unknown. Usually this token is \verb|<unk>|. With this in mind, we computed the overlap between the translated English (with the REV systems) and the original English. By doing so, we were able to compute all the words that did not appear in the original data but did appear in the output. Even when assuming that all these words were copied by the SMT system, the words that overlap for the EN--ES and EN--FR systems, 78,647 and 76,412 respectively, are still significantly higher than the English vocabularies of the RNN (28,742 and 27,349) and Transformer (40,321 and 40,629) systems. 


\begin{table}[]
    \centering
    {\small 
     \setlength\tabcolsep{2.2pt} 
    \begin{tabular}{|c|c|c|c|c|c|c|}\hline
        Lang. & \multicolumn{3}{c|}{EN*} & \multicolumn{3}{c|}{FR*/ES*}\\\cline{2-7}
         pair & RNN & SMT & Trans. & RNN & SMT & Trans. \\\hline
EN--FR & 28,742 & 106,441 & 40,321 & 36,991 & 123,770 & 42,309\\\hline
EN--ES & 27,349 & 118,362 & 40,629 & 39,805 & 138,193 & 44,545\\\hline
    \end{tabular}}
    \caption{Vocabularies of the English translation from the REV systems, used as source for the BACK systems and the French/Spanish output from the BACK systems.}
    \label{tbl:vocabularies_rev}
\end{table}

\section{Results}\label{sec:resultsLL}

In Table~\ref{tbl:bleu} we present automatic evaluation scores (BLEU and TER) for the 12 analysed systems. For completeness we present BLEU and TER for the REV systems in Table~\ref{tbl:bleu_rev}, although we do not consider them in our analysis. 

In what follows we use the following denotations to indicate the system we refer to: \verb|{src}-{trg}-{system}-{dir}|, where \verb|{src}| indicates the source language `en', that is English, \verb|{trg}| indicates the target language -- `fr' for French and `es' for Spanish -- and the system is one of `OT' for the original training data,\footnote{We are referring to the data as `original training data' and not to `human translations' as Europarl is a mixture of original data combined with human translations.} `PB-SMT' for PB-SMT, `rnn' for the RNN models and `trans' for the Transformer models; \verb|{dir}| is one of `ff' to indicate that the system is the forward, trained on the original data, `back' to indicate that the system is trained with back-translated data or `rev' to denote that it is the reverse system, trained after swapping source and target (the OT has no \verb|dir| index).

\begin{table}[]
    \centering
    {
    \begin{tabular}{|l|c|c|c|c|}\hline
        \multicolumn{1}{|c|}{System} & \multicolumn{2}{c|}{Dev set} & \multicolumn{2}{c|}{Test set}\\\cline{2-5}
        \multicolumn{1}{|c|}{reference}  & BLEU$\uparrow$ & TER$\downarrow$ & BLEU$\uparrow$ & TER$\downarrow$\\\hline
en-fr-rnn-ff & 33.7 & 50.7 & 33.8 & 51.0\\\hline
en-fr-smt-ff & 35.9 & 50.4 & 35.7 & 50.7\\\hline
en-fr-trans-ff & 35.9 & \textbf{49.5} & 36.0 & \textbf{49.4}\\\hline
en-fr-rnn-back & 32.8 & 52.1 & 33.0 & 52.1\\\hline
en-fr-smt-back & 35.2 & 51.0 & 35.0 & 51.3\\\hline
en-fr-trans-back & \textbf{36.3} & 49.8 & \textbf{36.3} & 49.9\\\hline\hline
en-es-rnn-ff & 37.4 & 45.3 & 37.9 & 45.3\\\hline
en-es-smt-ff & 38.5 & 45.8 & 38.6 & 45.9\\\hline
en-es-trans-ff & \textbf{39.4} & \textbf{44.5} & \textbf{39.5} & \textbf{44.5}\\\hline
en-es-rnn-back & 36.0 & 47.0 & 36.3 & 47.0\\\hline
en-es-smt-back & 38.0 & 46.5 & 38.0 & 46.5\\\hline
en-es-trans-back & \textbf{39.4} & 45.2 & 39.3 & 45.5\\\hline
    \end{tabular}}
    \caption{Automatic evaluation scores (BLEU and TER) for all MT systems.}
    \label{tbl:bleu}
\end{table}

\begin{table}[h]
    \centering
     {
     \begin{tabular}{|l|c|c|}\hline
        \multicolumn{1}{|c|}{System reference} & BLEU$\uparrow$ & TER$\downarrow$ \\\hline
        en-fr-rnn-rev & 33.3 & 50.2\\\hline
en-fr-smt-rev & 36.5 & 47.1\\\hline
en-fr-trans-rev & \textbf{36.8} & \textbf{46.8}\\\hline
en-es-rnn-rev & 37.8 & 45.0\\\hline
en-es-smt-rev & 39.2 & 44.0\\\hline
en-es-trans-rev & \textbf{40.4} & \textbf{42.7}\\\hline
    \end{tabular}}
    \caption{Automatic evaluation scores (BLEU and TER) for the REV systems.}
    \label{tbl:bleu_rev}
\end{table}
\paragraph{Evaluated output}
In total we trained 18 MT systems. To assess the validity of our hypothesis and provide a quantitative analysis of the investigated phenomena, we use the outputs from the FF and the BACK systems; the REV systems are used just to generate the back-translated data.

\subsection{Analysis}\label{sec:analysisLL}
In the analysis we compare word frequencies of the original target data to the translation output of the forward (FF) and backward (BACK) MT systems. We investigate two scenarios: (i) \textit{seen} and (ii) \textit{unseen} data. For (i) we translate the original source side of the training set (i.e. the English sentences) with the FF and with the BACK systems. The reason behind performing this kind of test is that since the MT system has seen this data during training, any loss of lexical richness and/or bias exacerbation are due to the inherent workings of the systems. That is, the observed differences between lexical diversity on seen data can only be attributed to the algorithm itself. For (ii) we are evaluating the lexical diversity on the (unseen) test set. This evaluation scenario is the one that gives us an indication of the overall lexical diversity of the translations produced by MT systems as compared to the data they were trained on. 

\paragraph{Lexical diversity score}
Lexical diversity (LD) refers to the amount or range of different words that are used in a text. The greater that range, the higher the diversity. Although LD has many applications (neuropathology, data mining, language acquisition), coming up with a robust index to quantify it has proven to be a difficult task. A comparison between different measures of LD~\citep{mccarthy2010} concluded by saying that, although there is no consensus yet, LD can be assessed in different ways, with each  measurement having its own assets and drawbacks. Therefore, we evaluated LD by using four different widely used metrics: type/token ratio (TTR)~\citep{Templin1975certain}, Yule's K (in practice, we use the reverse Yule's I)~\citep{Yule1944}, and the measure of textual lexical diversity (MTLD)~\citep{Mccarthy2005assessment}. 

The easiest lexical richness metric is TTR. TTR is the ratio of the types, i.e. total number of \textit{different} words in a text to its tokens, i.e. the total number of words. A high/low TTR indicates a high/low degree of lexical diversity. While TTR is one of the most widely used metrics, it has some drawbacks linked to the assumption of a linear relation between the types and the tokens. Because of that, TTR is only valid when comparing texts of a similar size, as it decreases when texts become longer due to repetitions of words \citep{brezina2018}. 

Yule's characteristic constant, or Yule's K, is a probability model of the changes that take place in the lexical frequency spectrum of a text as the text becomes longer. Yule's K and its reverse Yule's I are considered to be more immune to fluctuations related to text length than TTR~\citep{oakes2012}. 

Another metric used to study lexical richness and diversity is MTLD. The difference with the two previous methods is that MTLD is evaluated sequentially as the mean length of sequential word strings in a text that maintain a given TTR value \citep{McCarthyJarvis2007}. A more recent study by \cite{mccarthy2010} shows that MTLD is the most robust with respect to text length.

Our metrics are presented in Table~\ref{tbl:results_train} and Table~\ref{tbl:results_test}. Higher scores indicate higher lexical richness, and lower scores indicate lower lexical richness. Table~\ref{tbl:results_train} shows the metrics for the original human training data and the machine translations of the training set, i.e. the seen data, and Table~\ref{tbl:results_test} shows the scores for original training data and the machine translations of the test sets, i.e. the unseen data. Due to the large number of output words (e.g. the rnn-ff translation of the EN-FR test set contains 14,561,653 words), and the low vocabulary size relative to the total number of words our TTR scores are quite low. For readability and for ease of comparison we present these scores multiplied by a factor of 1000. We further discuss the results presented in Table~\ref{tbl:results_train} and Table~\ref{tbl:results_test} after having presented our additional experiments. 

\begin{table}[h]
     \centering
    {\begin{tabular}{|l|c|c|c|}\hline
         Translation & Yule’s I & TTR & MTLD\\
          &  & * 1000 & \\\hline\hline
en-fr-ot & \textit{9.2793} & \textit{2.9277} & \textit{127.1766} \\\hline\hline
en-fr-rnn-ff & 0.7107 & 0.8656 & 109.4506 \\\hline
en-fr-smt-ff & 6.7492 & 2.6442 & 118.1239 \\\hline
en-fr-trans-ff & 1.1768 & 1.0925 & 120.5179\\\hline
en-fr-rnn-back & 0.7587 & 0.8776 & 116.8942\\\hline
en-fr-smt-back & \textbf{7.8738} & \textbf{2.7496} & 120.9909\\\hline
en-fr-trans-back & 1.0325 & 1.0172 & \textbf{121.5801}\\\hline\hline
en-es-ot & \textit{12.3065} & \textit{3.7037} & \textit{99.0850}\\\hline\hline
en-es-rnn-ff & 0.6298 & 0.9394 & 89.3562\\\hline
en-es-smt-ff & 7.3249 & 3.1170 & 95.1146\\\hline
en-es-trans-ff & 1.0022 & 1.1581 & \textbf{96.2113}\\\hline
en-es-rnn-back & 0.7355 & 0.9829 & 95.7198\\\hline
en-es-smt-back & \textbf{8.1325} & \textbf{3.2166} & 95.1479\\\hline
en-es-trans-back & 0.9162 & 1.1014 & 95.0886\\\hline
    \end{tabular}}
    \caption{Lexical richness metrics (Train set).}
    \label{tbl:results_train}
\end{table}

\begin{table}[h]
     \centering
    {\begin{tabular}{|l|c|c|c|}\hline
         Translation & Yule’s I & TTR & MTLD\\
          &  & * 1000 & \\\hline\hline
en-fr-ot & 33.6709 & 5.7022 & 124.1889\\\hline\hline
en-fr-rnn-ff & 4.4766 & 2.1969 & 106.1370\\\hline
en-fr-smt-ff & 21.1230 & 4.8034 & 113.9262\\\hline
en-fr-trans-ff & 6.5352 & 2.5957 & 118.9642\\\hline
en-fr-rnn-back & 5.1490 & 2.3092 & 112.9991\\\hline
en-fr-smt-back & \textbf{25.7705} & \textbf{5.1254} & 117.6979\\\hline
en-fr-trans-back & 6.7921 & 2.6287 & \textbf{119.1729}\\\hline\hline
en-es-ot & 48.2366 & 7.6151 & 97.0591\\\hline\hline
en-es-rnn-ff & 4.7988 & 2.6250 & 85.4589\\\hline
en-es-smt-ff & 24.6771 & 5.9171 & 92.6397\\\hline
en-es-trans-ff & 6.7967 & 3.0432 & \textbf{94.4709}\\\hline
en-es-rnn-back & 6.0098 & 2.8357 & 92.4704\\\hline
en-es-smt-back & \textbf{28.0153} & \textbf{6.1887} & 92.3310\\\hline
en-es-trans-back & 7.3824 & 3.1483 & 92.8928\\\hline
    \end{tabular}}
    \caption{Lexical richness metrics (Test set).}
    \label{tbl:results_test}
\end{table}
We ought to note that also for the lexical diversity metrics, the comparison between PB-SMT systems and NMT systems is not entirely fair because of the different way in which both systems deal with unseen\footnote{Note that for some of our systems there was no `unseen' data so theoretically none of these words should be `unseen'.} words. We conducted additional experiments that compare the propagation of bias by looking at actual words and their frequencies. These experiments overcome the previously mentioned issue with unseen words when comparing PB-SMT to NMT output. We describe our approach in more detail in the next paragraph.

\paragraph{Word frequencies and bias} 
In order to prove/disprove our hypothesis, along with investigating lexical richness, we aim to investigate to what extent MT systems propagate bias in the output. This we assess by whether more/less frequent words in the original training translations have higher/lower frequency in the MT output (see Section~\ref{sec:hypothesisLL}). As soon as we started training the BACK systems, the first thing we observed was the reduced vocabularies from the FF systems. The loss of certain words (in the case of unknown words, the RNN and Transformer systems would generate the \verb|<unk>| token) already suggests biased MT. Comparing Table~\ref{tbl:vocabularies} and Table~\ref{tbl:vocabularies_rev}, we see that a lot of words are not accounted for in all systems, but that the RNN and Transformer models suffer the most. We believe this is due to the fact that NMT systems' advantage over more traditional systems, namely its ability to generalize and learn over the entire sentence, has a negative affect on lexical diversity, particularly for the least frequent words. 

Due to the differences in vocabularies and sentence lengths of the generated translations, in order to conduct a realistic comparison of the frequencies we applied 3 post-processing steps on the collected data: (i) we accounted for sentence variability by normalizing the frequency of each word (in the OT and the MT output) by the length of sentences in which it appears, (ii) we normalized the frequency of each word (in the OT and the MT output) by the accumulated frequency, reducing each frequency to a probability, and (iii) to account for the missing words in the MT output we counted words with zero frequencies separately. In addition, we need to make a distinction between frequent and non-frequent words. While this is a hard task in itself, here we commit to the average normalized word frequency of the original training data.

Once we applied the aforementioned post-processing, we compactly represent our data in six classes: 
\begin{itemize}
    \item \textit{Frequency increase of frequent words}: for a frequent word in the OT, its frequency in the MT is higher.  We denote this class using `\verb|+ +|' symbol combination. This class also indicates positive bias exacerbation.
    \item \textit{Frequency decrease of frequent words}: for a frequent word in the OT, its frequency in the MT is lower (but not zero).  We denote this class using `\verb|+ -|' symbol combination. 
    \item \textit{Frequency increase of non-frequent words}: for a non-frequent word in the OT, its frequency in the MT is higher.  We denote this class using `\verb|- +|' symbol combination. 
    \item \textit{Frequency decrease of non-frequent words}: for a non-frequent word in the OT, its frequency in the MT is lower (but not zero).  We denote this class using `\verb|- -|' symbol combination. This class indicates negative bias exacerbation.
    \item \textit{Zero frequency of frequent words}: a frequent word in the OT, does not appear in the MT.  We denote this class using `\verb|+ 0|' symbol combination. 
    \item \textit{Zero frequency of non-frequent words}: a non-frequent word in the OT, does not appear in the MT.  We denote this class using `\verb|- 0|' symbol combination. This class indicates negative bias exacerbation.
\end{itemize}

For each of these classes we count the (normalized) number of words, and we accumulate the absolute value of the differences for each of these cases. We present our results for the training data in Table~\ref{tbl:ex_freq_train}, Table~\ref{tbl:acc_freq_train} and for the test data -- in Table~\ref{tbl:ex_freq_test}, Table~\ref{tbl:acc_freq_test}. The numbers in Table~\ref{tbl:ex_freq_train} and Table~\ref{tbl:ex_freq_test} can be interpreted as the amount of translated words with higher, lower or zero frequency compared to the OT.\footnote{Note that these numbers are normalized for fair comparison.} The numbers in Table~\ref{tbl:acc_freq_train} and Table~\ref{tbl:acc_freq_test} quantify the differences between frequencies; they indicate the amount of increase or decrease in the frequencies presented by an MT system as compared to the OT. To derive information from these numbers, one should (mainly) compare the '\verb|+ +|' to `\verb|+ -|' and `\verb|- +|' to `\verb|- -|' and '\verb|+ 0|' to `\verb|- 0|'. The conclusions we draw from the tables presented are discussed and summarized in the next paragraphs.

\begin{table}[h]
    \centering
    {
    \begin{tabular}{|l|c|c|c|c|c|c|}\hline
         System & + + & + - & - + & - - & + 0 & - 0\\\hline
en-fr-rnn-ff & 3710 & 3023 & 10157 & 18683 & 10 & 95519\\\hline
en-fr-smt-ff & 3362 & 3381 & 32577 & 46714 & 0 & 45068\\\hline
en-fr-trans-ff & 3839 & 2901 & 12398 & 24403 & 3 & 87558\\\hline
en-fr-rnn-back & 3356 & 3372 & 13009 & 17253 & 15 & 94097\\\hline
en-fr-smt-back & 3246 & 3496 & 34111 & 43472 & 1 & 46776\\\hline
en-fr-trans-back & 3482 & 3254 & 14610 & 20962 & 7 & 88787\\\hline\hline
en-es-rnn-ff & 4667 & 3532 & 9929 & 19149 & 41 & 130875\\\hline
en-es-smt-ff & 4276 & 3963 & 39817 & 56169 & 1 & 63967\\\hline
en-es-trans-ff & 4626 & 3601 & 11379 & 25698 & 13 & 122876\\\hline
en-es-rnn-back & 4265 & 3951 & 13716 & 17872 & 24 & 128365\\\hline
en-es-smt-back & 4006 & 4233 & 39636 & 51831 & 1 & 68486\\\hline
en-es-trans-back & 4288 & 3929 & 14295 & 22032 & 23 & 123626\\\hline
    \end{tabular}}
    \caption{Frequency exacerbation and decay count for the Train or \emph{seen} data.}
    \label{tbl:ex_freq_train}
\end{table}
\begin{table}[h]
    \centering
    { 
     \begin{tabular}{|l|c|c|c|c|c|c|}\hline
         System & + + & + - & - + & - - & + 0 & - 0\\\hline
en-fr-rnn-ff & 2917 & 2335 & 10653 & 15400 & 11 & 57623\\\hline
en-fr-smt-ff & 2652 & 2610 & 20587 & 26949 & 1 & 36140\\\hline
en-fr-trans-ff & 2997 & 2264 & 12537 & 17430 & 2 & 53709\\\hline
en-fr-rnn-back & 2642 & 2610 & 13513 & 14963 & 11 & 55200\\\hline
en-fr-smt-back & 2577 & 2684 & 22604 & 26608 & 2 & 34464\\\hline
en-fr-trans-back & 2701 & 2554 & 14932 & 17101 & 8 & 51643\\\hline\hline
en-es-rnn-ff & 3541 & 2669 & 10636 & 16425 & 27 & 75113\\\hline
en-es-smt-ff & 3252 & 2982 & 23389 & 29057 & 3 & 49728\\\hline
en-es-trans-ff & 3508 & 2716 & 12069 & 19046 & 13 & 71059\\\hline
en-es-rnn-back & 3241 & 2971 & 14394 & 15847 & 25 & 71933\\\hline
en-es-smt-back & 3163 & 3072 & 24547 & 28389 & 2 & 49238\\\hline
en-es-trans-back & 3256 & 2967 & 15160 & 18606 & 14 & 68408\\\hline
    \end{tabular}}
    \caption{Frequency exacerbation and decay count for the test or \emph{unseen} data.}
    \label{tbl:ex_freq_test}
\end{table}
\begin{table}[h]
    \centering
    { 
     \begin{tabular}{|l|c|c|c|c|c|c|}\hline
         System & + + & + - & - + & - - & + 0 & - 0\\\hline
en-fr-rnn-ff & 840.76 & 687.16 & 46.36 & 115.27 & 1.47 & 83.22\\\hline
en-fr-smt-ff & 664.86 & 555.60 & 31.17 & 119.64 & 0.00 & 20.79\\\hline
en-fr-trans-ff & 663.00 & 552.74 & 49.98 & 108.63 & 0.40 & 51.20\\\hline
en-fr-rnn-back & 770.72 & 680.73 & 83.68 & 96.68 & 2.19 & 74.81\\\hline
en-fr-smt-back & 620.67 & 525.26 & 40.36 & 112.35 & 0.13 & 23.29\\\hline
en-fr-trans-back & 639.69 & 568.68 & 75.88 & 90.25 & 1.05 & 55.58\\\hline\hline
en-es-rnn-ff & 733.44 & 535.15 & 42.54 & 117.47 & 4.93 & 118.43\\\hline
en-es-smt-ff & 547.86 & 423.87 & 33.22 & 129.73 & 0.12 & 27.35\\\hline
en-es-trans-ff & 587.22 & 436.02 & 47.61 & 119.98 & 1.37 & 77.46\\\hline
en-es-rnn-back & 677.23 & 564.31 & 94.47 & 101.57 & 2.92 & 102.90\\\hline
en-es-smt-back & 561.03 & 438.09 & 44.31 & 133.35 & 0.12 & 33.78\\\hline
en-es-trans-back & 548.37 & 438.33 & 72.27 & 98.11 & 2.33 & 81.87\\\hline
    \end{tabular}}
    \caption{Accumulated frequency differences for the Train or \emph{seen} data.}
    \label{tbl:acc_freq_train}
\end{table}
\begin{table}[h]
    \centering
    { 
     \begin{tabular}{|l|c|c|c|c|c|c|}\hline
         System & + + & + - & - + & - - & + 0 & - 0\\\hline
en-fr-rnn-ff & 827.07 & 655.81 & 68.84 & 133.21 & 2.48 & 104.41\\\hline
en-fr-smt-ff & 790.41 & 640.60 & 60.98 & 156.94 & 0.13 & 53.71\\\hline
en-fr-trans-ff & 662.76 & 533.83 & 73.15 & 123.07 & 0.31 & 78.70\\\hline
en-fr-rnn-back & 751.49 & 655.35 & 112.32 & 114.16 & 2.28 & 92.01\\\hline
en-fr-smt-back & 679.17 & 551.88 & 64.50 & 142.50 & 0.34 & 48.96\\\hline
en-fr-trans-back & 625.59 & 548.18 & 104.26 & 107.39 & 1.41 & 72.88\\\hline\hline
en-es-rnn-ff & 726.16 & 509.28 & 67.76 & 134.45 & 4.16 & 146.04\\\hline
en-es-smt-ff & 679.08 & 503.57 & 70.86 & 169.33 & 0.38 & 76.67\\\hline
en-es-trans-ff & 592.32 & 414.37 & 73.00 & 134.59 & 1.84 & 114.52\\\hline
en-es-rnn-back & 653.89 & 533.03 & 128.86 & 119.04 & 4.22 & 126.46\\\hline
en-es-smt-back & 630.86 & 462.82 & 74.19 & 165.11 & 0.31 & 76.81\\\hline
en-es-trans-back & 538.03 & 415.49 & 103.32 & 118.89 & 2.40 & 104.57\\\hline
    \end{tabular}}
    \caption{Accumulated frequency differences for the Test or \emph{seen} data.}
    \label{tbl:acc_freq_test}
\end{table}

\paragraph{Remarks on automatic evaluation}
The summary of our results allows us to draw the following conclusions:
\begin{enumerate}
    \item \textit{Lexical richness}: All metrics and results presented in Table~\ref{tbl:results_train} and Table~\ref{tbl:results_test} and for both language pairs indicate that neither of the MT systems reaches the lexical richness of the OT. While PB-SMT systems (for both language pairs) retain more language richness according to two out of the three metrics (Yule's I and TTR) than the neural methods, the MTLD scores indicate that the Transformer systems lead to translations of higher lexical richness. This we may account to the different numbers of distinct words produced by PB-SMT and neural systems, which may be favoured by Yule's I and TTR (see Table~\ref{tbl:vocabularies_rev}). However, consistently, the worst systems are the RNN ones. 
    \item \textit{Automatic quality evaluation vs. lexical richness}: The results in Table~\ref{tbl:bleu} show that the Transformer systems perform best in terms of BLEU. The only lexical richness metric that corroborates the BLEU and TER scores is MTLD. This observation can act as a future research direction for integrating or improving quality evaluation metrics of MT to accommodate for lexical richness by possibly adopting features from MTLD.
    \item \textit{Bias}: To understand how the inherent probabilistic nature of PB-SMT and NMT systems exacerbates (or not) the bias, we rely on the result in Table~\ref{tbl:ex_freq_train}, Table~\ref{tbl:ex_freq_test}, Table~\ref{tbl:acc_freq_train} and Table~\ref{tbl:acc_freq_test}. More precisely, we focus on the comparison between `\verb|+ +|' and `\verb|+ -|', and the `\verb|- +|' and `\verb|- -|' classes as well as the values in the `\verb|+ 0|' and `\verb|- 0|' classes. One could simplify the analysis by joining the latter two classes together with `\verb|+ -|' and `\verb|- -|'. However, their independent analysis carries more important information. Precisely, we see that all of the systems lose less frequent words, indicated by the low numbers for the `\verb|+ 0|' class for both the training and the test set translations. Second, not all MT systems produce more words with higher frequencies (for the Train set: en-fr-smt-ff with 3362 vs 3381, en-fr-smt-back with 3246 vs 3496 and en-es-smt-back with 4006 vs 4233; for the test set: en-fr-smt-back with 2577 vs 2684), but the accumulative normalized frequency for such words is higher than that of the OT. The accumulated frequency differences indicate that MT systems are indeed biased towards these more frequent words. This observation, together with the fact that all MT systems suffer from loss of less frequent words, further supports our hypothesis that MT systems target learning the more frequent words and disregard the less frequent ones. 
    \item \textit{Lost words}:
    The `\verb|- 0|' columns indicate the words that were not frequent in the training data and completely lost in the automatic translation process. In Table~\ref{tbl:ex_freq_train} and Table~\ref{tbl:ex_freq_test}, we can see that all MT systems lose many of those infrequent words. However, the forward and backward Transformer and RNN systems lose approximately double the amount of infrequent words compared to the PB-SMT systems in both the EN--FR an EN--ES MT systems (Table~\ref{tbl:ex_freq_train}, the \emph{seen} data). The difference between the PB-SMT systems and the neural models is less on the \emph{unseen} data (Table~\ref{tbl:ex_freq_test}) but still very obvious. 
    \item \textit{Seen and unseen data}:
    We divided our experiments over seen and unseen data. From the perspective of lexical richness we see the same trends in both cases, although a slight decrease can be observed for the unseen test set (measured by the MTLD metric). With regards to the word frequencies comparing `\verb|+ +|' and `\verb|+ -|' classes in Table~\ref{tbl:acc_freq_train} and Table~\ref{tbl:acc_freq_test} we see similar trends. Furthermore, more words are lost altogether when translating the unseen test set. 
\end{enumerate}

It should be stressed that in this work we looked at the frequency of words, and as such the RNN and Transformer models we trained are not optimized according to state-of-the-art settings. In particular, no BPE is used to account for out-of-vocabulary problems, and the vocabularies have not been restricted prior to training (typically the vocabulary of an NMT system consists of the K, e.g. $50$k most frequent words/tokens).

Another observation that we ought to note is the fact that the BACK systems score quite high not only based on word frequencies and lexical richness metrics, but also based on the evaluation metrics presented in Table~\ref{tbl:bleu}. We assume this is due to the fact that the simplified source (translated by the REV systems) changes the complexity of the learned association. We plan to further investigate these systems. 

\paragraph{Semi-Manual Evaluation}
To obtain a more concrete image of the observed bias exacerbation by MT, we looked into the translations of 15 random English words: `picture', `create', `states', `happen', `genuine', `successful', `also', `reasons', `membership', `encourage', `selling', `site', `vibrant', `still' and `event'. This evaluation does not have the intention to be exhaustive, as the general tendencies of the systems have already been discussed in the previous sections. However, looking into some actual translations produced by the systems does further clarify the exacerbation effect of the learning algorithm. 

\begin{figure}[H]
    \centering
        \includegraphics[scale=0.90]{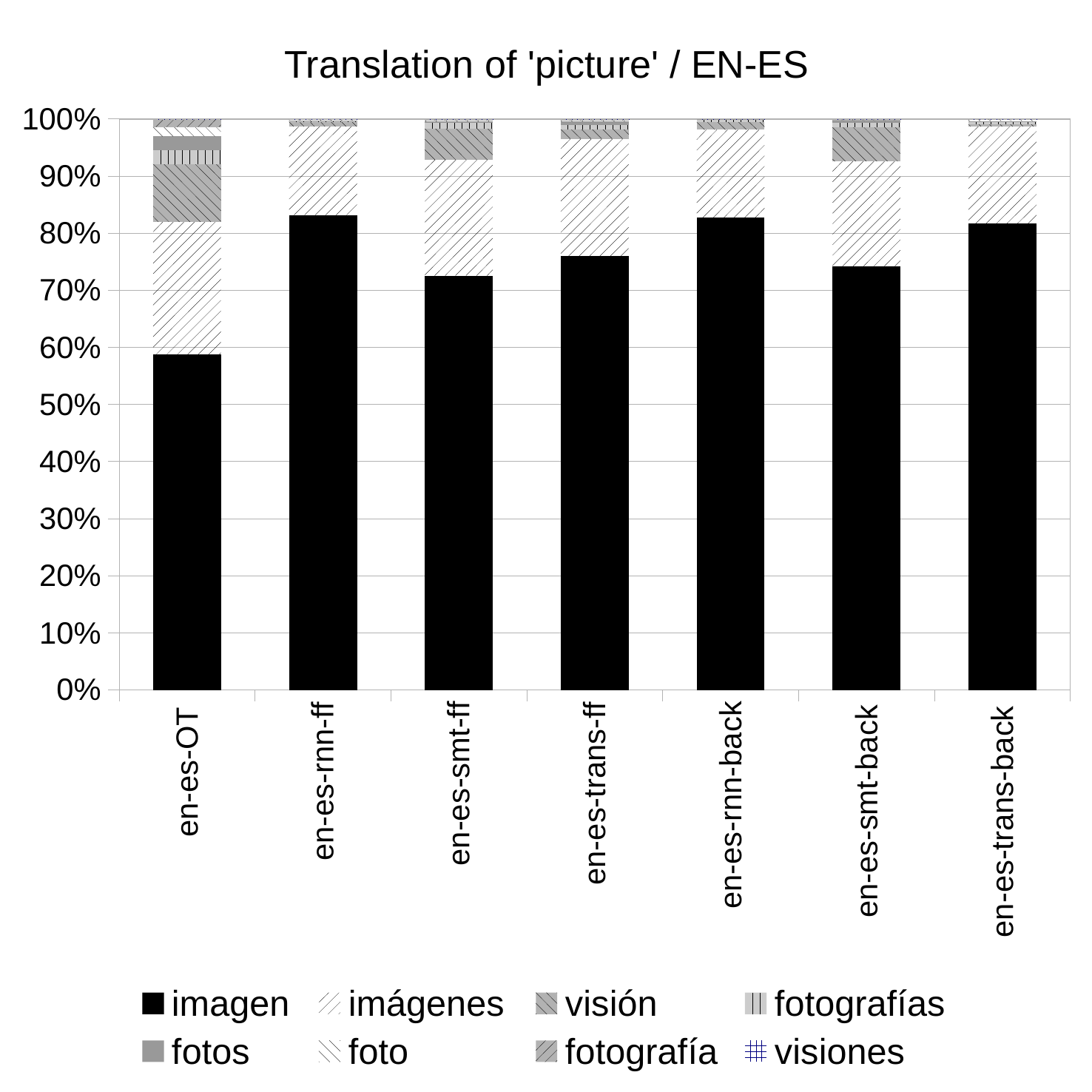}
    \caption{Relative frequencies of the Spanish translations of the English words `picture'.}
    \label{fig:translations_esPIC}
\end{figure}

Let us first look at the Spanish translations of the English word `picture', presented in Figure~\ref{fig:translations_esPIC}. The original training data shows quite a lot of diversity as `picture' can be translated into among others `imagen', `im\'{a}genes', `visi\'{o}n', `foto',`fotograf\'{i}as' and `fotos'. However, when we look at the output of the EN-ES MT systems, we see that all of them use the most frequent translation -- `imagen' -- even more frequently than in the original data. This comes at the expense of the other translation variants. Although the second most frequent translation (`im\'{a}genes') is still frequent, all others show a decrease and the least frequent ones disappear entirely.

\begin{figure}[h]
    \centering
        \includegraphics{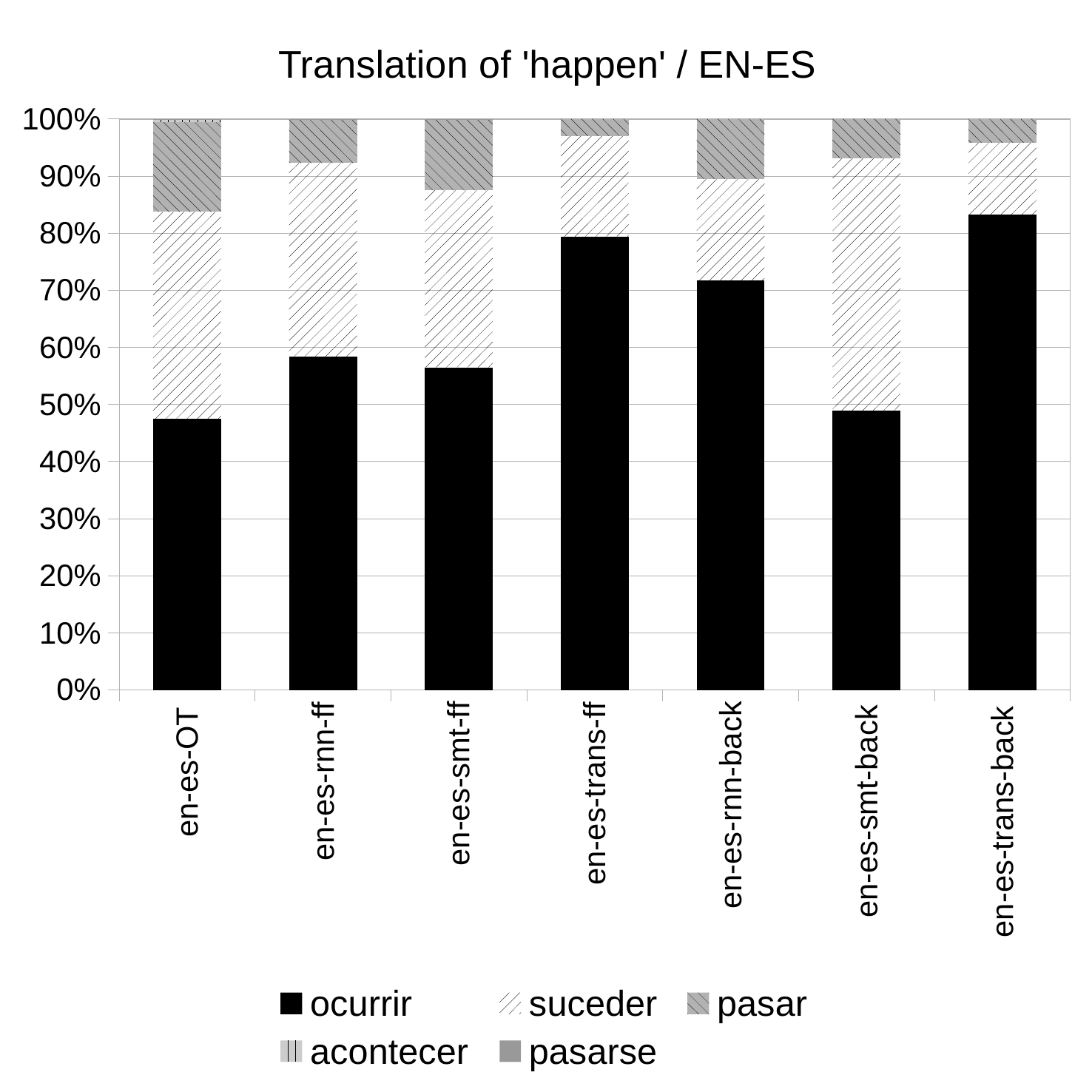}
    \caption{Relative frequencies of the Spanish translations of the English words `happen'.}
    \label{fig:translations_esHAP}
\end{figure}

Similar, though slightly different patterns are observed for the translations of the other words we examine. Presented in Figure~\ref{fig:translations_esHAP} are the translations of the English verb `happen' into the Spanish verbs `ocurrir', `suceder', `pasar', `acontecer' and `pasarse'. Again, the graphs show how the most frequent translation(s) gain in relative frequency at the cost of less frequent options.

\begin{figure}[h]
    \centering
        \includegraphics{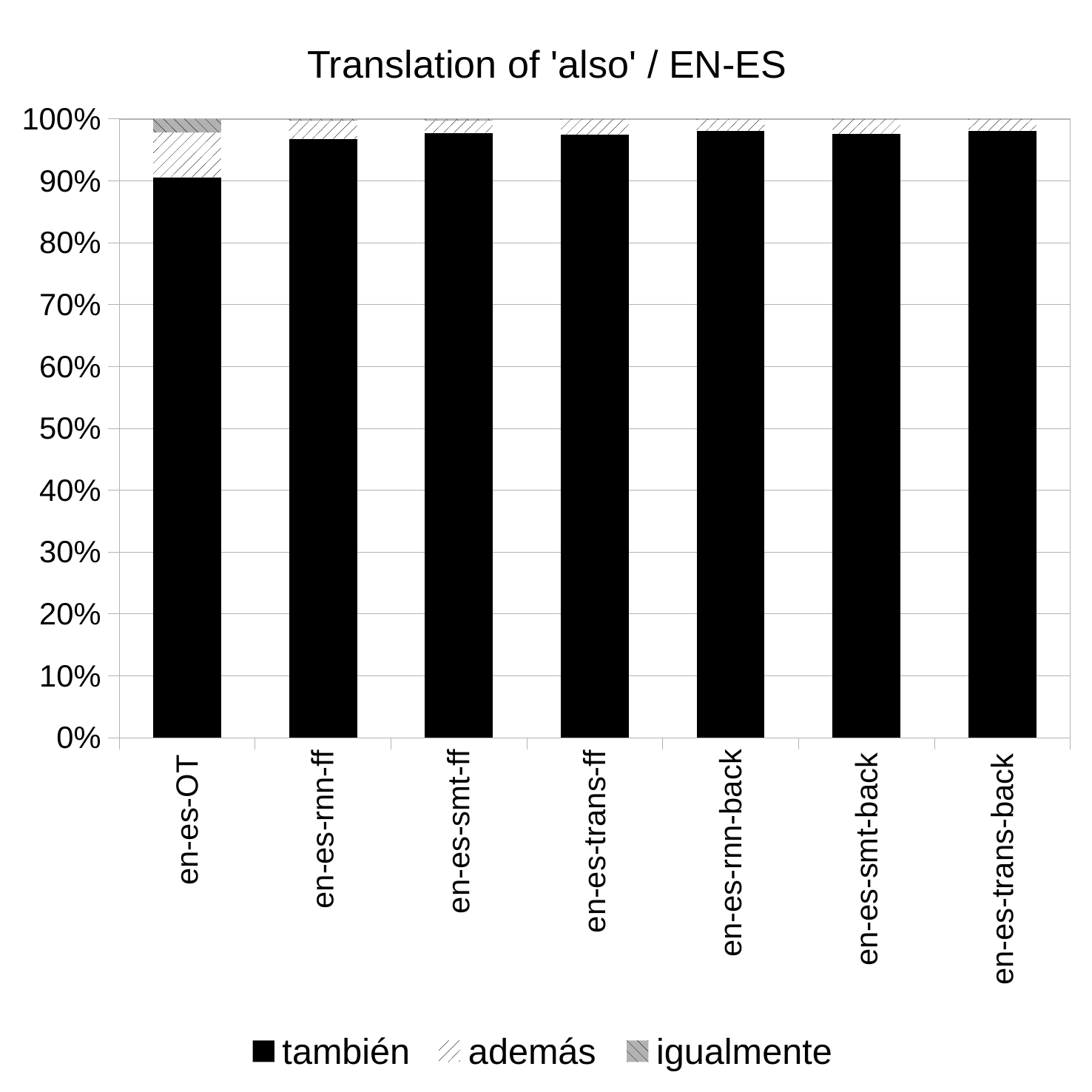}
    \caption{Relative frequencies of the Spanish translations of the English words `also'.}
    \label{fig:translations_esALSO}
\end{figure}

A third example can be found in Figure~\ref{fig:translations_esALSO} where the English adverbial connector `also' is translated into `tambi\'{e}n', `adem\'{a}s' and `igualmente'. The word `also' is translated in 90.6\% of the cases into the Spanish connector `tambi\'{en}' in the human data. It also has a 7.2\% chance of being translated into `adem\'{a}s', and a 2.2\% change of being translated into `igualmente'. Figure~\ref{fig:translations_esALSO} shows how all the MT systems' translations increase this disparity.
~\newline

Since Figure~\ref{fig:translations_esALSO} is slightly harder to interpret visually given the big disparity between the initial numbers in the human translation, we decided to provide the actual numbers in Table~\ref{tbl:translations_esALSO}.

\begin{table}[H]
    \centering
    { 
     \begin{tabular}{|l|c|c|c|}\hline
         ALSO       & tambi\'{en}   &adem\'{a}s     &igualmente     \\\hline
en-es-ot            & 90.6          & 7.2           & 2.1         \\\hline\hline
en-es-smt-ff     & 96.8          & 2.9           & 0.3         \\\hline
en-es-rnn-ff        & 97.5          & 2.4           & 0.1         \\\hline
en-es-trans-ff      & 97.6          & 2.3           & 0.1         \\\hline\hline
en-es-smt-back   & 97.8          & 2.0           & 0.3         \\\hline
en-es-rnn-back      & 98.1          & 1.8           & 0.2         \\\hline
en-es-trans-back    & 98.1          & 1.8           & 0.1         \\\hline
    \end{tabular}}
    \caption{Translation percentages of the English word `also' into the Spanish `tambi\'{e}n', `adem\'{a}s and `igualmente' for the forward and backward PB-SMT, RNN and Transformer systems.}
    \label{tbl:translations_esALSO}
\end{table}
~\newline
From Table~\ref{tbl:translations_esALSO} we can see more clearly that the MT systems increase the disparity compared to the original training data, as the percentage of the most frequent translation `tambi\'{e}n' rose even further while the two less frequent options `adem\'{a}s' and `igualmente' decreased. The back-translation systems exacerbate this phenomenon even more.

\section{Conclusions}\label{sec:conclusionsLL}
This chapter investigates bias exacerbation and loss of lexical richness through the process of MT. We analyse these problems using a number of LD metrics on the output of 12 different MT systems: PB-SMT, RNN and Transformer models for EN-FR and EN-ES with original\footnote{Original human data and original human translations.} and back-translated data. 

Via our experiments and their subsequent analysis, we observe that the process of MT causes a general loss in terms of lexical diversity and richness when compared to human-generated text. This confirms our hypothesis. Furthermore, we investigate how this loss comes about and whether it is indeed the case that the more frequent words observed in the input occur even more in the output, negatively affecting the frequency of less seen events or words by causing them to become even rarer events or causing them to disappear altogether. Our analysis shows that MT systems indeed increase the frequencies of more frequent words while the frequencies of less frequent words drops to such an extent that a very large amount of words are completely `lost in translation'. We believe this demonstrates that current systems overgeneralize and thus, we deem it appropriate to speak of a form of quantifiable algorithmic bias. The analysis and the results obtained address and answer RQ3.

Overall, the RNN systems are among the worst performing in terms of LD, although we do need to take into account that, for the sake of comparison, we did not use BPE, which might have given the neural models a disadvantage compared to the PB-SMT systems. PB-SMT systems performed better in terms of LD compared to the neural systems. Still, Transformer was the best-performing system in terms of automatic evaluation.


\chapterbib

%% file: Conclusion.tex

\newpage
\epigraph{That's what she said.}{\textit{Ancient Biblical Epitaph}}

\chapter{Conclusions and Future Work}\label{ch:Conclusion}

In this thesis, we address specific linguistic issues in the field of data-driven MT. We analyzed MT output and observed issues related to the translation of tenses, number and gender agreement and loss of lexical richness in general. By integrating features at the word- or sentence-level, we were able to overcome some of these problems providing the MT systems with the necessary additional information it is lacking. Moreover, we quantified the general loss of lexical richness in MT systems compared to original training data and linked this to the aforementioned linguistic issues. In this final chapter, we summarize and highlight the major contributions and findings of our work. 

First, in Section~\ref{subsec:conc:rq}, we briefly revisit our initial research questions defined in Chapter~\ref{ch:Introduction} and explain how they were addressed throughout our thesis. Next, we summarize the contributions of our work to the field of MT in Section~\ref{subsec:cont}. We provide ideas for future research avenues in Section~\ref{subsec:end:future_work}. Finally, we provide some closing remarks in Section~\ref{subsec:final}.

\section{Research Questions}\label{subsec:conc:rq}
In Chapter~\ref{ch:Introduction}, we formulated the following three research questions:
\begin{itemize}
\item \textbf{RQ1}: Is linguistic theory reflected in practice in the knowledge sources of data-driven Machine Translation systems?
\item \textbf{RQ2}: What type of (necessary) linguistic knowledge is lacking, and how can this be integrated in data-driven MT systems?
\item \textbf{RQ3}: Can we identify and quantify the underlying cause of many of the linguistic issues remaining in current MT systems?
\end{itemize}

We will briefly discuss how they have been addressed throughout our work and discuss the main findings and observations below.

\paragraph{RQ1}
Our first goal is to verify whether linguistic theory is reflected in practice in the knowledge sources of data-driven MT. The question we ask is a general question that we approached by looking at a specific linguistic phenomenon: the translation of tense and aspect. The correct handling of such a complex contrastive linguistic issue is sometimes linked to the meaning of the verb itself but at other times  requires information from a broader network of aspectual information (to be found at the sentence level or even across sentences). Subtle interactions between words and their (broader) surroundings are in no way exceptions when dealing with language. We believe that the approach and conclusions drawn are generalizable to other phenomena with similar characteristics. 

We looked at aspectual information provided in the knowledge sources of current MT systems: phrase-tables (PB-SMT) and encoding vectors (NMT). For PB-SMT, we observed that phrase-tables reflect the basic lexical aspect of verbs in a `static' way, while in reality, these aspects can change depending on the aspectual triggers in the broader context in which they appear. As PB-SMT has no means of looking at that broader context, it cannot effectively handle the complexity of tense translation. PB-SMT encodes basic linguistic information, but lacks contextual clues in order to deal with more complex linguistic patterns.

NMT encodes the entire sentence at once and thus has more contextual information available. We observed that a classifier trained on the encoding vectors can indeed more accurately\footnote{We used the human translations for comparison.} predict the appropriate aspectual value of a target sentence (French and Spanish) than a baseline PB-SMT system. However, NMT's advantage is lost during the actual decoding process, resulting eventually in similar performance when comparing PB-SMT and NMT output. Statistically speaking leveraging only local context will allow the automatic translation system to get it right \textit{most of the time}. By singling out more complex cases, where contextual information plays a more crucial role, we observe a decrease in SMT and NMT's performance with respect to their ability to deal with aspect and tense translations. Our work related to RQ1 is mainly covered in Chapter~\ref{ch:Aspect}.

\paragraph{RQ2} The second question we addressed is related to identifying features that may aid the translation process and integrating them into the MT pipeline. In order to answer this question, we started off by looking at PB-SMT and NMT translations, identifying the observed issues. We then experimented with the integration of specific linguistic features on the word level (PB-SMT and NMT) and on the sentence level (NMT). We worked on the integration of features for subject-verb agreement (Chapter~\ref{ch:Agreement}), gender agreement (Chapter~\ref{ch:Gender}) and more general semantic and syntactic features (Chapter~\ref{ch:Supertag}). 

For PB-SMT, we experimented with changing the original word forms on the source-side. We adapted the English source verb forms in such a way that disambiguation is facilitated, enabling the MT system to select the correct French target translation.

For NMT, we experimented with factored NMT systems. We integrated both higher-level semantic features (supersenses) and more fine-grained syntactic features (POS-tags and CCG-tags). The idea behind this approach was to provide the NMT systems simultaneously with very fine-grained syntactic information while also providing a means for better semantic abstraction. We experimented with various feature combinations and two language pairs, English--French and English--German. The combination of CCG-tags and supersense tags led to significant improvements in terms of automatic evaluation. Furthermore, the linguistic features allowed the NMT system to learn and converge more quickly than the baseline systems.

Additionally, we experimented with sentence-level features in NMT by providing a tag at the beginning of every sentence indicating the gender of the speaker. This approach is to some extent similar to domain adaptation. As such, we observed not only changes in terms of gender agreement in the output but also different word preferences. It is unclear whether these `side-effects' can be attributed to more general differences in male and female language usage. The (socio)linguistic literature provides us with contradictory evidence with respect to the existence of characteristic differences between male and female speech. As such, although improving the automatic evaluation scores, the side-effects we observed are in our opinion not desirable and should be further investigated.

Agreement issues and tenses are just a few topics we delved into and they are in no way an exhaustive list of the remaining problems in MT. Aside from the techniques we employed, there are other ways in which linguistic information can be integrated into the MT pipelines, as discussed in the related work sections of the relevant chapters. Still, our experiments show that there is indeed a need for more in-depth analyses of MT output in order to reveal systematic problems. Furthermore, we indicate how some of these issues cannot be resolved without employing additional (meta)contextual information (such as the gender of the speaker). 

\paragraph{RQ3} With our final question, we aim to identify the underlying cause of some of the aforementioned shortcomings of MT. 
Neural and statistical models learn by generalizing over the seen events. It is thus rather intuitive to think that such models could be prone to overgeneralization. Our objective was to demonstrate and quantify this overgeneralization. In order to do so, we looked at the problem from a more global perspective in Chapter~\ref{ch:Loss}. By showing that MT systems overgenerate the most frequently seen words and `undergenerate' or even completely ignore less frequent words, we demonstrate that it is justified to speak of an algorithmic bias. One of the most obvious consequences of algorithmic bias can be found when looking at the frequency of synonyms as it implies that the most frequent synonyms appear even more often, at the cost of less commonly used variants. However, such a bias does not limit itself to simple word choices as it can have an effect on morphological variants (e.g. tenses, gender and number) as well.     

We experimented with a variety of MT systems (PB-SMT, RNN and Transformer) for two language pairs, English--French and English--Spanish. We observed a significant loss for all MT systems in terms of lexical diversity when compared to human translations. On average, the PB-SMT models performed better than the neural models (RNN and Transformer) in terms of lexical richness metrics. The Transformer model obtained the highest BLEU score.

\subsection{Contributions}\label{subsec:cont}
The aforementioned research questions guided our research; answering them led to multiaspectual contributions that bridge, to some extent, the gap between linguistics and data-driven approaches to MT. The main contributions can be summarized as follows:
\begin{itemize}
    \item We observed that NMT's ability to encode more context than PB-SMT does not necessarily imply it can handle sentence-level phenomena better. In Chapter~\ref{ch:Aspect}, the correct aspectual value could be predicted from the encoding vectors with accuracies up to 91\% on a general test set. However, in the final MT output, this dropped to 79\%, only slightly better than the results obtained for PB-SMT with 78\%. As such, we identify that there is a discrepancy between NMT's ability to encode linguistic information and its ability to successfully decode it.
    \item We were the first to combine semantic and syntactic features into the NMT pipeline and explored various sets and combinations of features: CCG-tags, supersensetags and POS-tags. Our results are promising, especially when combining syntactic and semantic features.
    \item We published a corpus for 20 language pairs enriched with gender information in order to facilitate research on gender-feature integration in MT.
    \item We were the first to integrate speaker gender features into the NMT pipeline aiming to resolve gender agreement issues. We observed significant improvements according to BLEU. Interestingly, these improvements were not only due to better gender agreement (syntax) but also because of different word choices (semantics). The controllability that our gender-aware MT system offers is still limited. Nonetheless, we believe our work is important in highlighting issues with MT that have received very little attention by the community so far.
    \item We traced back the underlying cause of the linguistic issues discussed to a more general problem by quantifying the loss and decay of linguistic richness in MT systems. 
\end{itemize}

From a more general perspective, we would like to point out that what initially started off as research aiming to improve morphological agreement (Chapter~\ref{ch:Agreement} and Chapter~\ref{ch:Gender}), addressed purely from a  linguistic point of view, turned into something broader the more we delved into some of the problems observed. We identified that one of the main datasets used for MT -- Europarl -- has a 2:1 male/female speakers ratio and that using such biased datasets to train our MT algorithms is problematic in terms of generating diverse outputs. The lack of diversity in the data highlights the importance for research to integrate some factor of control over the output. Being able to control the specification of the gender of a speaker/writer when generating translations could furthermore be used to help with many practical applications of MT (e.g. translations on social media platforms, translations of speeches, talks, dialogues, subtitles, etc.). As we discussed in an article for \textit{Slator},\footnote{\url{https://slator.com/technology/he-said-she-said-addressing-gender-in-neural} \url{-machine-translation/}} an online language industry  magazine, there are situation in which failing to generate diverse translations could have consequences in real life. A search engine or selection algorithm that uses an MT system internally could be eliminating many perfectly good candidates simply because it failed to translate a gender-neutral term from one language into a male and female variant in another language.

Linking the issue of gender agreement to our next research topic, dealt with in Chapter~\ref{ch:Loss} on the loss of lexical richness and algorithmic bias, depicts an even more problematic picture for minority word forms in a broader sense. We believe that the work and experiments we conducted on gender and algorithmic bias are important not only because of the improvements in terms of automatic evaluation and agreement, but also for raising awareness with respect to the  broader ethical issues involved. As a research community, we have a responsibility to not only improve our systems but also raise awareness with respect to their shortcomings and drawbacks.

\section{Future Work}\label{subsec:end:future_work}
Our work provides many directions for future research opportunities. 

In Chapter~\ref{ch:Supertag} we experimented with various sets of syntactic and semantic features. However, our work lacks a consistent analysis of the benefits of such features. We showed that combining more diverse features can lead to an improvement in terms of automatic evaluation metrics, a more robust learning curve and faster convergence. Aside from the theoretical linguistic motivation behind our experimental design, we could not empirically pinpoint where the improvement came from. A systematic analysis of the semantics and syntax of output sentences generated by such systems would provide further insights into the benefits of such features. Additionally, we only explored the integration of linguistic features on the source side (English). Target-side features have proven to be useful in some cases as well and including them would provide a more complete picture in our search for useful feature combinations. We did not carry out such experiments as the tools we employed were not all available for the languages we experimented with.

In Chapter~\ref{ch:Gender}, we worked on the integration of gender features. We limited our work to gender features for the $1^{st}$ person (in this case, the speaker). Gender features for the $2^{nd}$ or $3^{rd}$ person singular or plural and the $1^{st}$ person plural are for many languages often equally necessary and we hope that in future work, this can be accounted for. Furthermore, our gender-aware NMT system showed promising results but had some (arguably undesirable) side-effects in terms of differences in word choices. Current research directions on gender seem to focus on gender bias in MT. However, we believe future work should prioritize controllability over the removal of gender biases. Stripping word-embedding vectors from `gender' would still allow for random variations between male and female endings, while for many real-life applications, a user or a system would want to (or have to) generate the appropriate male or female version of a translation and not leave it to chance (as with even a perfectly gender-balanced corpus without gender biases, we would still need to pick a gender). Furthermore, without additional context, accounting for sentences involving several agents such as the English sentence in Example~(\ref{ex:multi}) can lead to a multitude of possible translations with respect to gender (masculine (M) or feminine (F)) and/or number (singular (Sg) or plural (Pl)) in French.

\begin{li}
\item\begin{tabular}{ll}
(ENG) & \textbf{I} am happy \textbf{you} got married.\\
(Sg-M $+$ Pl-M) & \textbf{Je} suis heureu\textbf{x} que \textbf{vous} vous soyez mari\textbf{\'es}.\\
(SG-M $+$ Pl-F) & \textbf{Je} suis heureu\textbf{x} que \textbf{vous} vous soyez mari\textbf{\'ees}.\\
(Sg-F $+$ Pl-M) & \textbf{Je} suis heureu\textbf{se} que \textbf{vous} vous soyez mari\textbf{\'es}.\\
(SG-F $+$ Pl-F) & \textbf{Je} suis heureu\textbf{se} que \textbf{vous} vous soyez mari\textbf{\'ees}.\\
(Sg-M $+$ Pl-M) & \textbf{Je} suis heureu\textbf{x} que \textbf{tu} te sois  mari\textbf{\'e}.\\
(SG-M $+$ Pl-F) & \textbf{Je} suis heureu\textbf{x} que \textbf{tu} te sois  mari\textbf{\'ee}.\\
(Sg-F $+$ Pl-M) & \textbf{Je} suis heureu\textbf{se} que \textbf{tu} te sois  mari\textbf{\'e}.\\
(SG-F $+$ Pl-F) & \textbf{Je} suis heureu\textbf{se} que \textbf{tu} te sois  mari\textbf{\'ee}.\\
\end{tabular}\label{ex:multi}
\end{li}

Providing translations for sentences such as (\ref{ex:multi}) is not only challenging from a linguistic point of view (in terms of generating all the correct alternatives), it would also require a careful integration into existing MT tools in order for it to remain accessible and interpretable in practice.

In Chapter~\ref{ch:Loss} we identified the loss of diversity and linguistic richness in data-driven MT systems. The experiments conducted aimed at identifying, analysing and quantifying lexical loss. We observed that the current PB-SMT and NMT systems overgeneralize, leading to a significant loss in terms of lexical richness when compared to the human reference translations. In the future, potential solutions should be explored to overcome such overgeneralizations. In order not to lose the richness of language, one would need a model that allows a degree of randomness while simultaneously maintaining a strong learning (and thus generalizing) ability. This is a very complex (and potentially contradictory) task. Because of that, we believe current directions should focus especially on those cases where overgeneralization leads to grammatical mistakes. One possible direction could be inspired by the diversity-encouraging models that have been explored already in the field of natural language generation.

A more general direction for future research would be the experimentation with and the creation of more diverse datasets. We often limited our experiments to a specific dataset (e.g. Europarl, OpenSubtitles) depending on the availability of the necessary amount of data in combination with the extra-linguistic features needed for our experiments (e.g. the possibility to retrieve information about the speakers). We hypothesize that gender features could be even more beneficial when dealing with less formal texts that contain more interaction between speakers. Furthermore, the loss of lexical diversity could be even more pronounced when looking at a more general domain, as the legal domain of Europarl  already limits the lexical ambiguity of the published texts.


\section{Final Remarks}\label{subsec:final}
The arrival of NMT challenged once more the usefulness of linguistic features. NMT's great learning ability is undeniable and, by now, we believe it is fair to say it beats PB-SMT on almost every level. Nevertheless, some more complex problems remain and they reveal something about NMT's underlying competence. As such, we do not believe NMT is able to effectively learn grammar rules at this point, as even with tremendous amounts of parallel training data, occasionally a short simple sentence can reveal a simple number-agreement issue (see Example~(\ref{ex:belle}) in Chapter~\ref{ch:Background}). Furthermore, we highlighted other linguistic problems that remain and require further in-depth linguistic analysis and understanding in order for a solution to be engineered. We only scratched the surface and put the first cornerstone towards potential solutions. We believe more effective solutions will require multidisciplinary research and collaborations between experts from different disciplines such as (socio)linguistics, computer scientists and ethicists.




\chapterbib
